\documentclass[a4paper,fleqn]{cas-sc}

\usepackage[authoryear]{natbib}
\usepackage{amsmath,amsfonts}
\usepackage{tikz}
\usepackage{tikz-network}
\usepackage{caption}
\usepackage{subcaption}
\usepackage{upgreek}
\usepackage[dvipsnames]{xcolor}
\newcommand{\rev}[1]{\textcolor{black}{#1}}

\newcommand{\realrev}[1]{\textcolor{black}{#1}}

\usepackage{amssymb} %
\usepackage{placeins}
\definecolor{okgreen}{RGB}{0,120,60}
\definecolor{oomred}{RGB}{190,30,30}
\definecolor{errorange}{RGB}{200,110,0}
\newcommand{\rOK}{\textcolor{okgreen}{\ensuremath{\checkmark}}}
\newcommand{\rOOM}{\textcolor{oomred}{\ensuremath{\times}}}

\newcommand{\rSKIP}{\textcolor{black!35}{\textendash}}

\def\tsc#1{\csdef{#1}{\textsc{\lowercase{#1}}\xspace}}
\tsc{WGM}
\tsc{QE}
\tsc{EP}
\tsc{PMS}
\tsc{BEC}
\tsc{DE}

\begin{document}
\let\WriteBookmarks\relax
\def\floatpagepagefraction{1}
\def\textpagefraction{.001}

\shorttitle{GREAT}

\shortauthors{Lischka et~al.}

\title [mode = title]{GREAT: A Graph Edge Attention Network for the Electric Vehicle Routing Problem with Asymmetric Costs}

\author[1]{Attila Lischka}[
                        orcid=0009-0003-1371-5996]

\cormark[1]

\ead{lischka@chalmers.se}

\affiliation[1]{organization={Electrical Engineering, Chalmers University of Technology},
    city={Gothenburg},
    country={Sweden}}

\affiliation[2]{organization={Architecture and Civil Engineering, Chalmers University of Technology},
    city={Gothenburg},
    country={Sweden}}

\affiliation[3]{organization={Computer Science and Engineering, Chalmers University of Technology and University of Gothenburg},
    city={Gothenburg},
    country={Sweden}}

\author[1]{Filip Rydin}[orcid=0009-0002-4311-0771]
\author[2]{Jiaming Wu}[orcid=0000-0002-0235-4246]
\author[3]{Morteza Haghir Chehreghani}[orcid=0000-0002-2912-7422]

\author[1]{Balázs Kulcsár}[orcid=0000-0002-3688-1108]

\cortext[cor1]{Corresponding author}

  \nonumnote{Code, trained models, and datasets are available at \url{https://github.com/attila-lischka/GREAT}.}

\begin{abstract}
Due to the ongoing electrification of transportation systems, energy-based routing problems such as the Electric Vehicle Routing Problem (EVRP) are becoming increasingly important. Compared to traditional routing problems based on Euclidean distances, the EVRP involves inherently asymmetric energy costs caused by road topology, including elevation changes and one-way streets. Due to regenerative braking, energy costs may even become negative on downhill segments. Existing learning-based routing approaches are typically based on graph neural networks (GNNs) or transformer architectures operating on Euclidean node coordinates, making them poorly suited for asymmetric, non-Euclidean edge-based metrics such as energy consumption.
To address this limitation, we propose the Graph Edge Attention Network (GREAT), a novel GNN-related architecture that operates purely in edge space rather than on node coordinates, making it naturally compatible with asymmetric routing costs. Integrated into an end-to-end reinforcement learning framework, GREAT jointly optimizes routing and charging decisions for the EVRP. We train the model using real-world energy consumption data from San Francisco which is derived from physical road networks enriched with topological features, and evaluate generalization capabilities on datasets from San Francisco, Vienna, and Gothenburg.
Across diverse battery sizes ($15-40$ kWh) and customer demand distributions, GREAT outperforms a strong HGS-based EVRP heuristic in most tested settings, achieving energy reductions of up to 8.78\% while reducing runtime by a factor of $60$. 
Moreover, we benchmark against a custom adaptive large neighborhood search (ALNS) that jointly optimizes routing and charging. While GREAT achieves competitive performance and a much faster runtime, we show that the ALNS can profit from a warm initialization by the GREAT solution and additionally improve performance.
We further evaluate GREAT on standard (asymmetric) routing problems, including Traveling Salesman Problem, Capacitated Vehicle Routing Problem, and Orienteering Problem, where it achieves state-of-the-art performance among neural methods for non-Euclidean settings.
\end{abstract}


\maketitle

\section{Introduction}
\label{sec:introduction}

With the transport sector accounting for approximately 23\% of global energy-related $\text{CO}_2$ emissions \citep{jaramillo2023transport}, its electrification represents one of the most critical levers for achieving climate neutrality  \citep{iea_netzero_2021}.
Consequently, with the number of electric vehicle (EV) fleets growing, a need for computationally efficient methods to plan routes that minimize energy consumption while respecting battery constraints and integrating charging decisions arises.
This problem is known as the Electric Vehicle Routing Problem (EVRP) and it is a complex, NP-hard generalization \citep{ERDOGAN2012100, schneider2014electric} of the classical Capacitated Vehicle Routing Problem (CVRP).

A further, key characteristic of real-world EVRP instances is that energy consumption between locations is inherently \emph{asymmetric}. Traveling from node $i$ to node $j$ typically consumes a different amount of energy than traveling from $j$ to $i$ even though the Euclidean distance between the two points on a map is symmetric.
This asymmetry arises from the topological characteristics of the underlying road network such as elevation changes and one-way street constraints.
Moreover, by regenerative braking, modern electric vehicles can recover energy on downhill segments. This can result in \emph{negative} energy costs along certain directed edges.
As a consequence, the energy consumption matrix of a real-world EVRP instance is neither symmetric nor does it generally satisfy the triangle inequality (cf. Section~\ref{sec:data_gen}).

These characteristics result in considerable challenges for machine learning-based methods to solve vehicle routing problems. 
Such machine learning-based methods have gained traction in recent years as effective heuristics for solving vehicle routing and other combinatorial optimization problems quickly.
By interpreting routing problems as graph problems, graph neural network (GNN)-based approaches \citep{joshi2019efficient, hudson2021graph, xin2021neurolkh} and transformer-based approaches \citep{kool2018attention, kwon2020pomo, jin2023pointerformer} have been used to compute node-level feature representations of underlying routing problems, starting from Euclidean coordinates as node inputs.
This design choice is not merely a convenience: the message passing operations and attention mechanism of these architectures are designed to operate on node-level data and are ill-suited for purely edge-based settings such as asymmetric energy matrices that characterize real-world EVRP.

To overcome this limitation, we propose the Graph Edge Attention Network (GREAT), a novel, GNN-related neural architecture that is tailored to operate directly on \emph{edge features} (e.g., a pairwise energy consumption matrix) by propagating information along edges sharing endpoints.
Since GREAT processes inputs that are purely edge-based, it applies natively to any routing problem regardless of whether the cost matrix is Euclidean or asymmetric.

By embedding GREAT into an end-to-end reinforcement learning (RL) framework, we train it to construct EVRP solutions that jointly minimize energy consumption while respecting battery constraints and planning charging stops.
To train the framework, we use real-world road network data from San Francisco and evaluate it, without retraining, on data from Vienna and Gothenburg, demonstrating convincing generalization capabilities.
In addition to the EVRP, we validate the generality of GREAT by applying it to established combinatorial routing benchmarks for learning-based methods, including the TSP, CVRP, and the Orienteering Problem (OP), across Euclidean and two asymmetric distributions.

The contributions of this paper are as follows:
\begin{itemize}
    \item We propose GREAT, a novel \rev{edge-input} GNN-related architecture that propagates information along edges sharing endpoints via an attention-like mechanism. Unlike existing GNN and transformer architectures for routing, \rev{GREAT takes only an edge feature matrix as input (i.e., no node coordinates are required), making it natively applicable to routing problems with any cost structure, including asymmetric and non-metric ones.} \rev{The key insight is that, while node-based architectures can optionally incorporate edge features, they still require node coordinates as primary inputs and are fundamentally ill-suited when only a pairwise cost matrix is available.}

    \item We develop an RL framework for the EVRP based on GREAT. Our framework jointly optimizes routing and charging decisions using real-world energy consumption data. The framework can handle asymmetric energy costs, negative energy consumption (resulting from regenerative braking), limited battery capacity, customer demands, and charging stops, with battery sizes from 15 to 40 kWh and varying customer demand distributions.

    \item We train the model on data from San Francisco and demonstrate strong generalization capabilities by applying it to data from Vienna and Gothenburg without any retraining. GREAT outperforms an adapted HGS baseline in the majority of tested configurations while, at the same time, being approximately $60\times$ faster. Moreover, we design a baseline based on an ALNS metaheuristic that jointly optimizes routing and charging. We show that GREAT has a considerable runtime advantage compared to ALNS and that GREAT can serve as a powerful warm initialization for this search.

    \item We validate the generality of GREAT on standard routing benchmarks for learning-based approaches (TSP, CVRP, OP) across Euclidean and asymmetric distributions. Notably, the asymmetric OP and CVRP settings have not been explored before in the learning-for-routing literature, establishing GREAT as a first neural model for this setting. The ability to handle asymmetric routing problems is a key prerequisite for approaching more sophisticated, real-world formulations like the EVRP. On most asymmetric distributions, GREAT outperforms MatNet \citep{kwon2021matrix}, the leading neural baseline for non-Euclidean routing.

    \item We demonstrate the ability of GREAT to scale to large routing problem instances via a few-shot curriculum learning pipeline, achieving competitive performance on TSP instances with up to 1,000 nodes. Further, we show strong zero-shot generalization on instances significantly larger than those encountered during training.
\end{itemize}

The remainder of this article is structured as follows.
Section~\ref{sec:related_work} reviews related work on learning-based routing and non-Euclidean routing problems.
Section~\ref{sec:problem_formulation} formally defines the EVRP and the auxiliary routing problems considered.
Section~\ref{sec:great} introduces the GREAT architecture.
Section~\ref{sec:framework} describes the GREAT-based RL framework, training setup, and curriculum learning.
Section~\ref{sec:experiments} presents experimental results, and Section~\ref{sec:conclusion} concludes.

\begin{table}[h]
\centering
\caption{List of Abbreviations}
\label{tab:abbreviations}
\begin{tabular}{@{}ll@{}}
\toprule
\textbf{Abbreviation} & \textbf{Definition} \\
\midrule
EVRP   & Electric Vehicle Routing Problem \\
CVRP   & Capacitated Vehicle Routing Problem \\
TSP   & Traveling Salesman Problem \\
OP   & Orienteering Problem \\
EV & Electric Vehicle \\
SoC & State of Charge \\
GNN & Graph Neural Network \\
GAT & Graph Attention Network \\
GREAT & Graph Edge Attention Network \\
GREAT NF & node-free GREAT version \\
GREAT NB & node-based GREAT version \\
ALNS & Adaptive Large Neighborhood Search \\
HGS & Hybrid Genetic Search \\
MatNet & Matrix Encoding Network  \\
RL & Reinforcement Learning \\
POMO & Policy Optimization with Multiple Optima \\
EUC & Euclidean distance distribution \\
TMAT & Asymmetric distance distribution with triangle inequality \\
XASY & Extremely asymmetric distance distribution without triangle inequality \\
\bottomrule
\end{tabular}
\end{table}

\section{Related Work}
\label{sec:related_work}

\subsection{Learning-Based Vehicle Routing}
Due to their NP-hard nature, vehicle routing problems are challenging to solve, resulting in a need for strong heuristic methods that can quickly compute  good solutions.
In recent years, many studies have tried to solve routing problems such as the TSP or CVRP with learning-based methods.

Popular approaches for solving routing problems with the help of machine learning include autoregressive RL frameworks. These frameworks contain an initial phase where encodings for the problem instances are computed. In a second phase, the encodings are then used to incrementally build solutions by selecting one node in the problem instance at a time. Successful works in this category include \cite{deudon2018learning, nazari2018reinforcement, kool2018attention, kwon2020pomo, jin2023pointerformer}.

Another possibility to use machine learning for solving routing problems is to predict edge probabilities or scores which are later used in search algorithms such as beam search or guided local search. Examples for such works are \cite{joshi2019efficient, kool2018attention, fu2021generalize, xin2021neurolkh, hudson2021graph, sun2023difusco, min2024unsupervised}.

A further possibility involves iterative methods where a solution to a routing problem is repeatedly refined until a stopping criterion (e.g., convergence) is met. Possibilities for such improvements are optimizing subproblems or applying improvement operators such as $k$-opt. Examples for such works are \cite{da2021learning, wu2021learning, cheng2023select, lu2019learning, chen2019learning, li2021learning}.

\subsection{Non-Euclidean and Asymmetric Routing}
Many of the works mentioned so far use GNNs or transformer models (which are related to GNNs via GATs  \citep{joshi2020transformers}) to capture the structure of the routing problem in their neural architecture. This is done by interpreting the coordinates of Euclidean routing problem instances as node features. These node features are then processed in the GNN or transformer architectures to produce encodings of the overall problem.
However, this limits the applicability of such works to Euclidean routing problems. This is unfortunate, as non-Euclidean routing problems are highly relevant in reality \citep{boyaci2021vehicle}.

So far, only a few studies have also investigated non-Euclidean versions of routing problems, such as the asymmetric TSP (ATSP). 
One such study is \cite{gaile2022unsupervised} where they solve synthetic ATSP instances with unsupervised learning, RL, and supervised learning approaches. Another study is \cite{wang2023reinforcement} that uses online RL to solve ATSP instances of TSPLIB \citep{reinhelt2014tsplib}.
Another successful work tackling ATSP is \cite{kwon2021matrix}. There, the Matrix Encoding Network (MatNet) is proposed, a neural model suitable to operate on matrix encodings representing combinatorial optimization problems such as the distance matrices of (A)TSP. Their model is trained using RL.
\cite{ye2024glop} proposes a modular framework to solve larger routing problem instances and uses MatNet as a subsolver to tackle bigger (A)TSP instances.
\cite{drakulic2023bq, drakulic2025goalgeneralistcombinatorialoptimization} proposes a learning formulation that allows training on relatively small routing problem instances while still performing well on larger instances at inference. Further, they propose a model that is trained for several routing problems at once. While doing so, they also tackle ATSP, using a standard GNN as their encoder that uses random one-hot vectors as node-feature inputs together with edge features representing the actual distances.

In contrast to all of the above, our model is the first to process the asymmetric cost matrix as edge features natively, without any artificial node-level representation.

\rev{We further note that more recently, several concurrent and subsequent works have extended neural routing to asymmetric settings. RADAR \citep{yi2026radar} uses Singular Value Decomposition on asymmetric distance matrices to initialize compact node embeddings in node-based architectures. RRNCO \citep{son2026rrnco} introduces adaptive node embeddings that fuse spatial coordinates with real-world distance features to bridge the gap between models trained on synthetic data and real-world data. EFormer \citep{ijcai2025p954} picks up a similar idea to the GREAT preprint \citep{lischka2024great_arxiv} and develops a transformer model operating on edge inputs. UniteFormer \citep{meng2026uniteformer} proposes a unified solver supporting node-only, edge-only, and hybrid inputs. Several works have directly built upon GREAT's edge-input paradigm and use GREAT as a component in their processing pipelines: \citet{rydin2026beyond} and \cite{rydin2026twostagelearneddecompositionscalable} extend edge-based routing to multigraphs with multiple edges between node pairs.
\cite{jacobsen2026learningdialaridedeepgraph} tackle the electric dial-a-ride problem which is defined by asymmetric energy cost matrices. Further \cite{lischka2026greateergraphedgeattention} tackle an emergency evacuation setting where asymmetric travel time matrices define the underlying problem instances.
These subsequent works demonstrate that the edge-input paradigm introduced by GREAT is a productive research direction.}

\subsection{Traditional and Learning-Based Methods for the EVRP}
Traditionally, variants of the EVRP have been tackled using metaheuristics. \citet{schneider2014electric} use a hybrid variable neighborhood search with tabu search. Other works often use ALNS to solve versions of the EVRP that contain mixed electric and conventional vehicle fleets \citep{GOEKE201581}, ECVRP with partial recharging \citep{KESKIN2016111} or mixed electric fleets \cite{HIERMANN2016995}.

More recently, the EVRP has been approached with tabular RL \citep{BASSO2022102496} and with online RL \citep{dorokhova2021routing}, the latter therefore requiring retraining for each individual instance.
\cite{lin2021deep} tackles EVRP with time windows using deep RL, however, they assume Euclidean distances in an attention-based encoder which is unsuitable for asymmetric distances. Similarly, also using deep RL, Euclidean EVRP is also solved in \cite{TANG2023121711}.
In \cite{BASSO202124}, machine learning was used to predict \realrev{a vehicle's} energy consumption in the EVRP, but the routing problem was solved using traditional methods.

To the best of our knowledge, no deep learning-based approaches for EVRP exist that can simultaneously handle asymmetric, real-world energy costs, limited battery capacity and charging decisions within a single end-to-end trainable model that generalizes across cities and customer demand distributions without retraining.
Our work is the first to address all of these aspects together.

\subsection{Graph Neural Networks and Edge-Based Architectures}
Graph neural networks (GNNs) are a class of neural architectures that operate on graph-structured data. 
In contrast to classical neural network architectures, where the neuron connections are fixed and grid-shaped, the connections in a GNN reflect the structure of the input data.
GNNs have emerged as a powerful tool for learning on graph-structured data such as molecules, social networks, or citation graphs \citep{wu2020comprehensive}.

Standard GNNs, like, e.g., the Graph Attention Networks (GAT) \citep{velickovic2017graph}, compute node feature representations via node-level message-passing operations.
While several extensions additionally incorporate edge features into node-level message passing \citep{chen2021edge, shi2020masked, jin2023edgeformers}, none of these architectures are purely edge-focused.

\rev{Ours is the first GNN-based architecture whose sole input is an edge feature matrix, requiring no node coordinates whatsoever. This makes GREAT natively applicable to any routing problem regardless of whether the cost structure is Euclidean, asymmetric, or non-metric. Internally, information propagates exclusively between edges that share an endpoint, without maintaining persistent node features across layers.}
This design is ideally suited for edge-level tasks such as routing, and it avoids the $\mathcal{O}(n^3)$ complexity that would result from converting edges to nodes via a line graph transformation (cf. \cite{chen2017supervised}).
\rev{Crucially, such a line-graph transformation is not a viable alternative in practice: vehicle routing graphs are complete, which means a graph with $n$ nodes has $n^2$ edges. Transforming it into a line graph yields $\mathcal{O}(n^2)$ nodes and $\mathcal{O}(n^3)$ edges. We note that training neural solvers requires processing \emph{batches} of many such graphs simultaneously. For $n=100$ a single line-graph instance would already have $10{,}000$ nodes and up to $10^6$ edges, making batch training on current hardware infeasible.}

\section{Problem Formulation}
\label{sec:problem_formulation}

\subsection{Electric Vehicle Routing Problem}
\label{sec:evrp_def}
We define the EVRP on a complete, directed graph $G = (V, E)$.
The node set $V = \{0\} \cup V_c \cup V_s$ consists of a depot (node 0), a set of customer nodes $V_c$, and a set of charging station nodes $V_s$.
Each customer node $v \in V_c$ has an associated demand $d_v$.
A delivery vehicle departs from the depot with full battery, serves all customers, and in the end returns to the depot (intermediate depot visits are possible).
Each directed edge $(i, j) \in E$ has an associated energy consumption $\omega_{ij} \in \mathbb{R}$ in kWh. Notably, these values can be negative representing downhill edges that allow for energy recovery via regenerative braking.

The vehicle is subject to two types of capacity constraints:
\begin{enumerate}
    \item Load capacity: The total demand of all customers on a subtour between two depot visits must not exceed the vehicle's load capacity $Q$.
    \item Battery capacity: The state of charge (SoC) $s \in [0, 1]$ must remain in the interval $[0.2, 0.8]$ at all times (i.e., we charge up to 80\% only and the vehicle does not discharge below 20\%). This effectively reduces the usable battery to 60\% of its nominal capacity but reflects realistic operating conditions to reduce battery degradation and prevent the risk of full depletion.
\end{enumerate}

\begin{figure}
    \centering
    \includegraphics[width=0.99\linewidth]{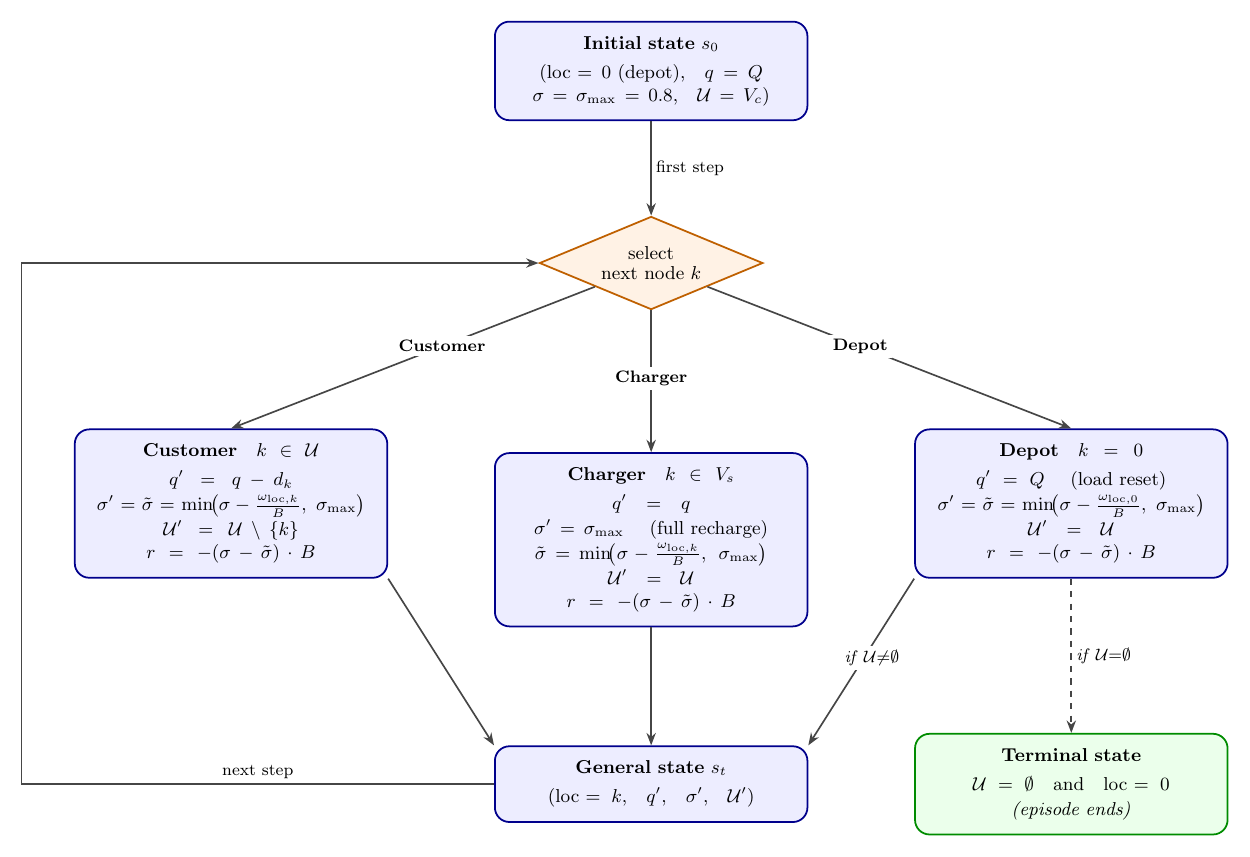}
    \caption{\rev{MDP transition diagram for the EVRP.  At each step we select a feasible next node by masking out invalid options. The three possible node types (customer, charger, depot) result in distinct transitions for load $q$, SoC $\sigma$, and unvisited set $\mathcal{U}$.  The MDP terminates when all customers are served and the vehicle has returned to the depot.  Reward equals the negative energy cost of each traversed edge adjusted for possible battery limits.}}
\label{fig:mdp}
\end{figure}

Traversing an edge $(i,j)$ changes the SoC by $-\omega_{ij}/B$, where $B$ is the battery capacity in kWh. By visiting a charging station, the SoC is reset to its maximum allowed level. The objective of the problem is to minimize the total energy consumption of the solution tour. 
\realrev{We point out that the edge costs $\omega_{ij}$ are derived under the premise of fastest path routing in the underlying road network that connects $i$ and $j$ (Section~\ref{sec:data_gen}). This reflects the reality of drivers generally choosing the fastest routes between deliveries. Consequently, the objective of our problem can be interpreted as the ``minimization of energy consumption over the \emph{sequence of stops} under fastest-path routing.''}
\rev{A formal Markov Decision Process (MDP) specification of these dynamics, including the precise state space, transition function, and reward signal used by our RL framework, is given in Section~\ref{sec:mdp} and illustrated in Figure~\ref{fig:mdp}.}

\subsubsection{Markov Decision Process Formulation}
\label{sec:mdp}

We define a Markov Decision Process (MDP) with $(\mathcal{S},\, \mathcal{A},\, \mathcal{T},\, r,\, s_0)$ for the EVRP as follows:

\paragraph{State space $\mathcal{S}$.}
A state $s_t \in \mathcal{S}$ captures all the information that is needed to decide which node to visit next:
\[
  s_t \;=\; \bigl(\,\text{loc}_t,\; q_t,\; \sigma_t,\; \mathcal{U}_t\,\bigr),
\]
where
\begin{itemize}
  \item $\text{loc}_t \in V$ is the current location (depot, customer, or charger) of the vehicle;
  \item $q_t \in [0, Q]$ is the current remaining load capacity of the vehicle;
  \item $\sigma_t \in [\sigma_{\text{min}}, \sigma_{\text{max}}] = [0.2, 0.8]$ is the current state of charge (SoC) of the vehicle's battery;
  \item $\mathcal{U}_t \subseteq V_c$ is the set of unvisited customers.
\end{itemize}

\paragraph{Initial state $s_0$.}
The vehicle starts at the depot with full usable battery and the total vehicle load capacity available:
\[
  s_0 \;=\; \bigl(\,0,\; Q,\; \sigma_{\max},\; V_c\,\bigr).
\]

\paragraph{Action space $\mathcal{A}$.}
At state $s_t$ we select the next node to visit as an action $a_t$. The \emph{feasible} action set
$\mathcal{A}(s_t) \subseteq V$ excludes:
\begin{itemize}
  \item customers who have already been visited ($V_c \setminus \mathcal{U}_t$);
  \item customers $j \in \mathcal{U}_t$ whose demand $d_j$ exceeds the remaining load $q_t$ of the vehicle;
  \item any node $j$ that cannot be reached without violating the SoC lower bound \realrev{ while going to the node $j$ itself or, subsequently, while going to the charger $j' \in V_s$ nearest to $j$, i.e.}\ $\sigma_t - \omega_{\text{loc}_t, j}/B - \min_{j' \in V_s}\omega_{j, j'}/B  < \sigma_{\min}$.
\end{itemize}
The MDP ends when all customers have been served and the vehicle has returned to the depot.

\paragraph{Transition function $\mathcal{T}$.}
Given action $a_t = j$, the next state $s_{t+1}$ is determined deterministically:
\begin{align*}
  \text{loc}_{t+1} &= j, \\[4pt]
  q_{t+1} &= \begin{cases}
    q_t - d_j & \text{if } j \in \mathcal{U}_t \text{ (customer visit)},\\
    Q          & \text{if } j = 0 \text{ (depot return)},\\
    q_t        & \text{if } j \in V_s \text{ (charger visit)},
  \end{cases}\\[4pt]
  \sigma_{t+1} &= \begin{cases}
    \min\!\bigl(\sigma_t - \omega_{\text{loc}_t,j}/B,\;\sigma_{\max}\bigr) & \text{if } j \notin V_s,\\
    \sigma_{\max} & \text{if } j \in V_s \text{ (full recharge)},
  \end{cases}\\[4pt]
  \mathcal{U}_{t+1} &= \begin{cases}
    \mathcal{U}_t \setminus \{j\} & \text{if } j \in \mathcal{U}_t,\\
    \mathcal{U}_t & \text{otherwise.}
  \end{cases}
\end{align*}
The $\min(\cdot,\sigma_{\max})$ clamp in the SoC update captures regenerative braking on downhill edges:
the energy recovered on such an edge ($\omega_{ij}<0$) increases the SoC but it cannot exceed the maximum allowed level.

\paragraph{Reward $r$.}
The reward at each step is the \emph{negative} energy consumed on the traversed edge. However, since the SoC cannot exceed $\sigma_{\max}$, we need to account for cases where a vehicle travels downhill with an already (almost) full battery. In such cases, only parts of the available energy can be recovered. Therefore, we define the temporary SoC at arrival at a new node after choosing an action $a_t$ as $\tilde{\sigma} = \min\!\bigl(\sigma_t - \omega_{\text{loc}_t,a_t}/B,\;\sigma_{\max}\bigr)$. This allows us to compute the difference in SoC and, by rescaling with the battery size, obtain the actual consumed amount of energy as the reward:
\[
  r(s_t, a_t) \;=\; -\,(\sigma_t - \tilde{\sigma}) \cdot B.
\]
Note that also charging station visits contribute their actual edge cost to the objective.
The overall cumulative reward $\sum_t r(s_t,a_t)$ equals the negative total energy consumption, so \emph{maximizing} the reward is equivalent to \emph{minimizing} total energy consumption which is exactly the EVRP objective.

\paragraph{Terminal condition.}
The MDP terminates when $\mathcal{U}_t = \emptyset$ and $\text{loc}_t = 0$ (i.e., all customers are served and the vehicle is back at the depot).
Feasibility is guaranteed throughout by the action mask on $\mathcal{A}(s_t)$.

\subsection{Auxiliary Routing Problems for Benchmarking}
\label{sec:aux_problems}

\begin{figure*}
\centering
    \begin{subfigure}[b]{0.30\textwidth}
        \centering
        \includegraphics[width=\textwidth]{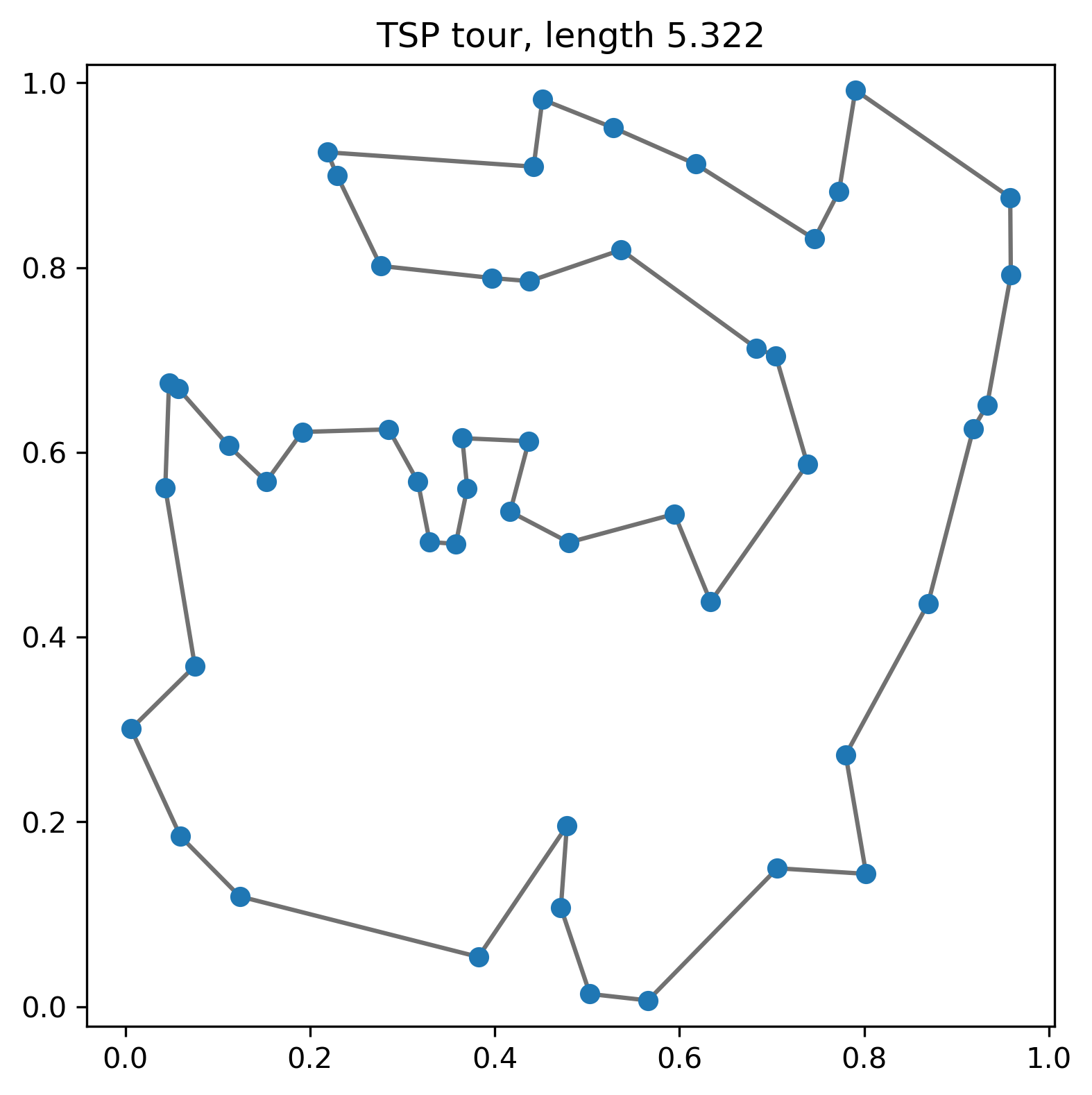}
        \caption{TSP (50 nodes)}
        \label{tsp_example_figure}
    \end{subfigure}
    \hfill
    \begin{subfigure}[b]{0.30\textwidth}
        \centering
        \includegraphics[width=\textwidth]{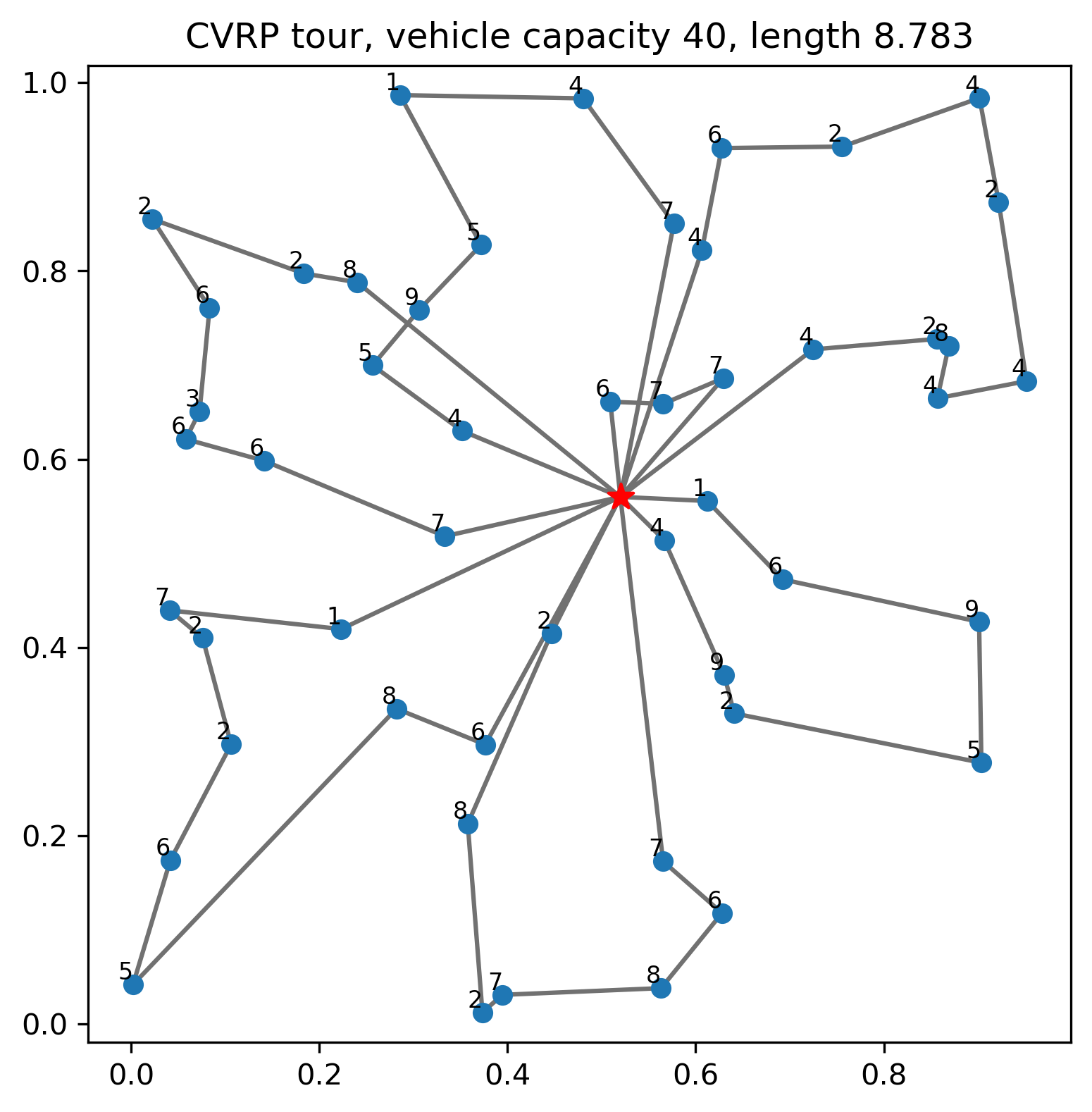}
        \caption{CVRP (50 customers)}
        \label{cvrp_example_figure}
    \end{subfigure}
    \hfill
    \begin{subfigure}[b]{0.30\textwidth}
        \centering
        \includegraphics[width=\textwidth]{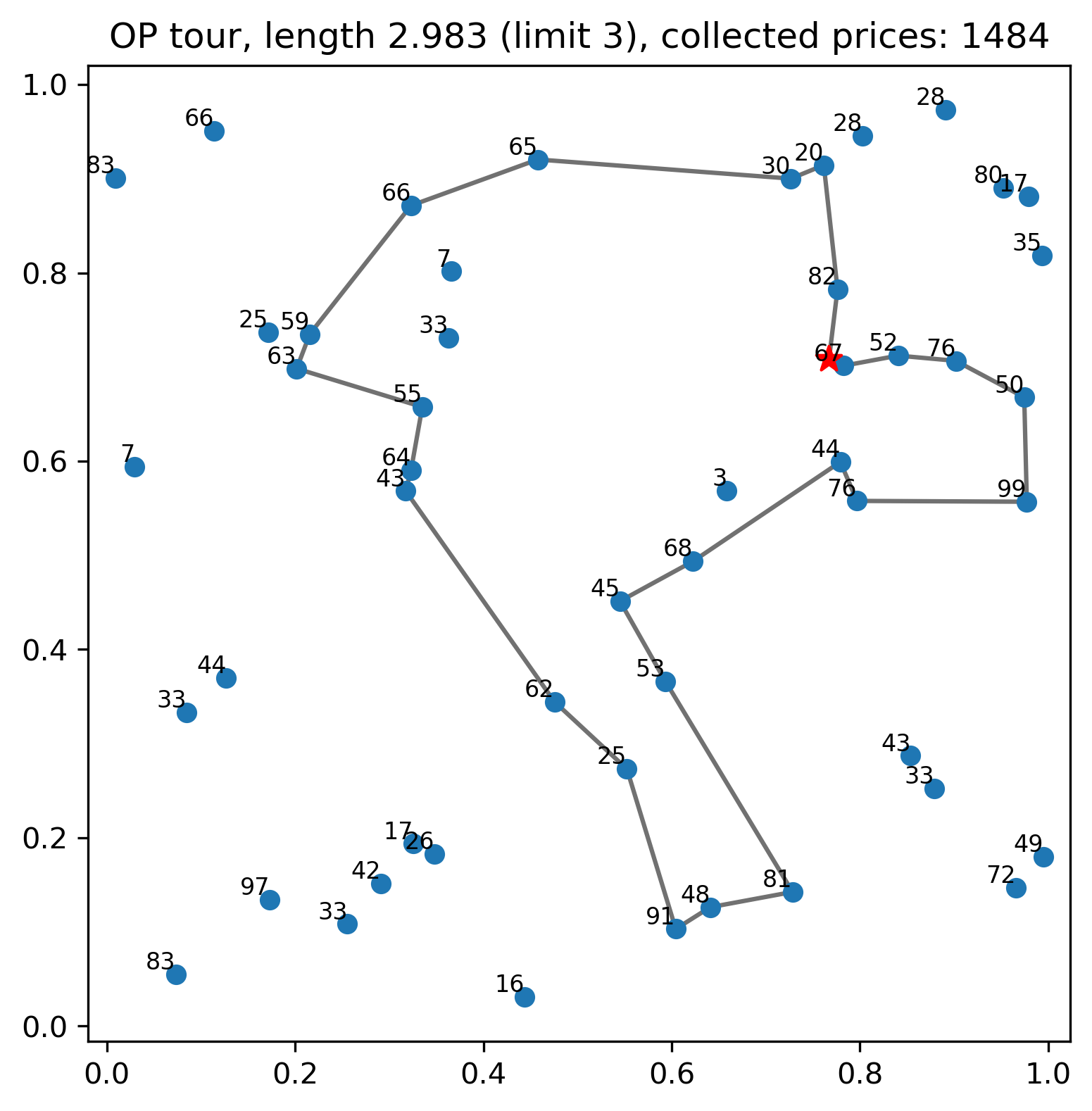}
        \caption{OP (50 customers)}
        \label{op_example_figure}
    \end{subfigure}
    \caption{Example instances of the three auxiliary routing benchmarks in the Euclidean setting.}
    \label{routing_problems}
\end{figure*}

To compare our GREAT framework to other learning-based routing approaches, we additionally consider three standard vehicle routing problems as a benchmark across Euclidean and two asymmetric distance distributions. Examples of the three auxiliary problems are shown in Figure~\ref{routing_problems}. 
The capability to solve asymmetric versions of these standard routing problems lays the foundation for solving more advanced, real-world settings such as the aforementioned EVRP.

\subsubsection{Traveling Salesman Problem (TSP)}
Given a set of nodes and pairwise distances, the goal of the TSP is to find the shortest tour visiting every node exactly once.
We consider three distributions: Euclidean (EUC; with Euclidean coordinates in the unit square), TMAT (asymmetric with triangle inequality, following \cite{kwon2021matrix}), and XASY (extremely asymmetric, where all pairwise distances are sampled uniformly at random in $(0,1)$).

\subsubsection{Capacitated Vehicle Routing Problem (CVRP)}
The CVRP extends the TSP by requiring a vehicle with limited capacity $Q$ to serve all customers (with associated demands) from a depot while respecting load constraints and allowing multiple depot visits.
Our demand and capacity parameters are inspired by \cite{kool2018attention} and we use the same EUC, TMAT, and XASY distributions as for TSP as distances.

\subsubsection{Orienteering Problem (OP)}
In the OP, each customer has an associated prize. The objective is to maximize total collected prizes while respecting a maximum travel distance (note that not all customers have to or even can be visited). For more information on the OP, we refer to \cite{vansteenwegen2019orienteering}.
We use the same distance distributions as for TSP and CVRP.

\section{The GREAT Architecture}
\label{sec:great}

\subsection{Motivation: Edge-Focused Representation Learning}

\begin{figure*}
    \centering
    \begin{subfigure}{0.48\textwidth}
        \centering
        \resizebox{0.9\columnwidth}{!}{
        \begin{tikzpicture}[
    node/.style={circle, draw=black, fill=cyan!80, text=white, minimum size=1cm},
    arrow/.style={->, thick, red},
    d_arrow/.style={
    ->,
    line width=1.5pt,
    dashed,
    Dandelion!100!black,
    scale=1,
    every arrow/.append style={
        scale=0.7
    }
},
    every label/.style={black}
]

\node[node] (1) at (0, 0) {1};
\node[node] (2) at (-2.5, 1.5) {2};
\node[node] (3) at (-3, -1.5) {3};
\node[node] (4) at (3.5, 1.5) {4};
\node[node] (5) at (5, 0) {5};
\node[node] (6) at (3.5, -1.5) {6};

\draw[-, thick] (1) -- node[pos=0.5, fill=white, inner sep=1pt] (l12) {$e_{12}$} (2);
\draw[-, thick] (1) -- node[pos=0.5, fill=white, inner sep=1pt] (l13) {$e_{13}$} (3);
\draw[-, thick] (1) -- node[pos=0.5, fill=white, inner sep=1pt] (l14) {$e_{14}$} (4);
\draw[-, thick] (1) -- node[pos=0.5, fill=white, inner sep=1pt] (l15) {$e_{15}$} (5);
\draw[-, thick] (1) -- node[pos=0.5, fill=white, inner sep=1pt] (l16) {$e_{16}$} (6);

\draw[arrow, bend right=30, shorten >=4pt, shorten <=4pt] (1) to (2);
\draw[arrow, bend left=30, shorten >=4pt, shorten <=4pt] (1) to (3);
\draw[arrow, bend left=30, shorten >=4pt, shorten <=4pt] (1) to (4);
\draw[arrow, bend right=15, shorten >=4pt, shorten <=4pt] (1) to (5);
\draw[arrow, bend right=40, shorten >=4pt, shorten <=4pt] (1) to (6);

\draw[d_arrow, bend left=30, shorten >=2pt, shorten <=4pt] (1) to (l12);
\draw[d_arrow, bend right=30, shorten >=2pt, shorten <=4pt] (1) to (l13);
\draw[d_arrow, bend left=25, shorten >=2pt, shorten <=4pt] (1) to (l14);
\draw[d_arrow, bend left=20, shorten >=2pt, shorten <=4pt] (1) to (l15);
\draw[d_arrow, bend right=30, shorten >=2pt, shorten <=4pt] (1) to (l16);

\end{tikzpicture}
        }
        \caption{Classical GNN (e.g., GAT): node feature is updated by aggregating over neighboring nodes' feature vectors.}
        \label{fig:gnn}
    \end{subfigure}
    \hfill
    \begin{subfigure}{0.48\textwidth}
        \centering
        \resizebox{0.9\columnwidth}{!}{
        \includegraphics{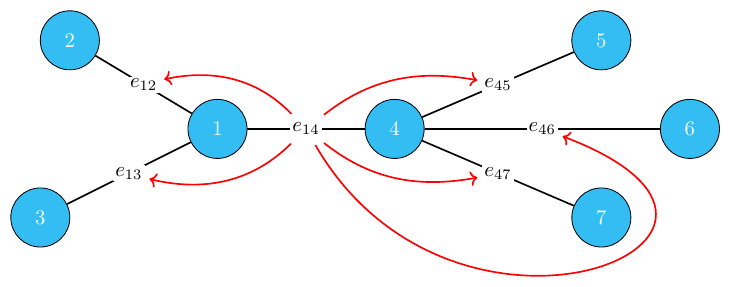}
        }
        \caption{GREAT (node-free variant): edge $e_{14}$ attends to all adjacent edges sharing a node endpoint.}
        \label{fig:great}
    \end{subfigure}
\caption{Comparison of aggregation mechanisms.}
\label{fig:comparison1}
\end{figure*}

Existing GNNs and transformers for routing problems produce node-level embeddings via message passing \citep{wu2020comprehensive} starting from Euclidean node coordinates as input features.
A standard GNN node update (shown in Figure~\ref{fig:gnn} for the Graph Attention Network (GAT) \citep{velickovic2017graph}) takes the form:
\begin{equation}
x^{\ell+1}_{i} = \sum_{j \in N(i) \cup \{i\}} \alpha_{i,j} \Uptheta_{t} x^{\ell}_{j}  \nonumber
\end{equation}
where $N(i)$ denotes the set of neighboring nodes of node $i$, $\alpha_{i,j}$ is an attention score between nodes $i$ and $j$, $x^\ell_i$ is the feature vector of node $i$ at layer $\ell$, and $\Uptheta_t$ is a trainable weight matrix.
While the computation of $\alpha_{i,j}$ can optionally include edge information, no real edge-features are computed.
This hinders the applicability of such GNNs to non-Euclidean settings where no node features are given and the problem is defined by a distance matrix corresponding to edge features. 

To overcome this limitation, we introduce a new architecture that, similar to the Graph Attention Network (GAT), we call Graph Edge Attention Network (GREAT).
GREAT operates natively on edge features and works by propagating information along edges that share an endpoint, as illustrated in Figure~\ref{fig:great}.
This makes GREAT an edge-level counterpart to node-focused GNNs and a natural fit for routing problems specified by pairwise cost matrices.

In theory it would also be possible to construct a line graph to transform edges into nodes and by this make ``normal'' GNNs applicable to the problem. However, this would transform an original dense routing graph with $n$ cities and $n^2$ edges into a representation with $\mathcal{O}(n^2)$ nodes and $\mathcal{O}(n^3)$ edges (cf. \cite{hudson2021graph}), increasing the order of magnitude of the representation by one. As a result, the representations would become prohibitively large for instances with only few dozens of nodes.
\realrev{Similarly, it is not possible to deploy a transformer globally on the whole edge space of a complete graph with $n$ nodes and $n^2$ edges. Since the complexity of transformers scales quadratically, applying it to an input size that is already quadratic results in $\mathcal{O}(n^4)$. GREAT, however, as we will describe in Section~\ref{GREAT_math} scales in $\mathcal{O}(n^2)$ which is the same order of magnitude as its input size (from an edge-based point of view) and the same complexity as a regular transformer operating on the node space. We investigate this empirically in Appendix~\ref{appendix_gpu_limits} where we show the limits of GREAT and a regular Transformer operating on the node space during training and inference.}

\subsection{Architecture}
In the following, we provide the mathematical formulations defining the different layers of a GREAT model.
In particular, we propose two variants of GREAT. 

The first variant is purely edge-focused and does not have any node features. Here, each edge compares information with all other edges it shares at least one endpoint with. The idea essentially corresponds to the visualization in Figure \ref{fig:great}. In the following, we refer to this variant as ``node-free'' GREAT or GREAT ``NF''.

The second variant is also edge-focused but has intermediate, temporary node features. This essentially means that nodes are used to save all information on adjacent edges. Afterwards, the features of an edge are computed by combining the temporary node features of their respective endpoints. 
These node features are then deleted and \textit{not} passed on to the next layer and only the edge features are passed on, making this GREAT variant fundamentally different from regular GNNs. 
The idea of this GREAT variant is visualized in Figures \ref{fig:great_n} and \ref{fig:great_n2}.
In the remainder of this study, we refer to this GREAT version as ``node-based'' or GREAT ``NB''. 

\begin{figure*}
    \centering
    \begin{subfigure}{0.48\textwidth}
        \centering
        \resizebox{0.9\columnwidth}{!}{
        \begin{tikzpicture}[
    node/.style={circle, draw=black, fill=cyan!80, text=white, minimum size=1cm},
    arrow/.style={->, thick, red},
    d_arrow/.style={
    ->,
    line width=1.5pt,
    dashed,
    Dandelion!100!black,
    scale=1,              %
    every arrow/.append style={
        scale=0.7          %
    }
},
    every label/.style={black}
]

\node[node] (1) at (0, 0) {1};
\node[node] (2) at (-2.5, 1.5) {2};
\node[node] (3) at (-3, -1.5) {3};
\node[node] (4) at (3, 0) {4};
\node[node] (5) at (6.5, 1.5) {5};
\node[node] (6) at (8, 0) {6};
\node[node] (7) at (6.5, -1.5) {7};

\draw[-, thick] (1) -- node[pos=0.5, fill=white, inner sep=1pt] (l12) {$e_{12}$} (2);
\draw[-, thick] (1) -- node[pos=0.5, fill=white, inner sep=1pt] (l13) {$e_{13}$} (3);
\draw[-, thick] (1) -- node[pos=0.5, fill=white, inner sep=1pt] (l14) {$e_{14}$} (4);
\draw[-, thick] (4) -- node[pos=0.5, fill=white, inner sep=1pt] (l45) {$e_{45}$} (5);
\draw[-, thick] (4) -- node[pos=0.5, fill=white, inner sep=1pt] (l46) {$e_{46}$} (6);
\draw[-, thick] (4) -- node[pos=0.5, fill=white, inner sep=1pt] (l47) {$e_{47}$} (7);

\draw[arrow, bend right=50, shorten >=2pt, shorten <=4pt] (1) to (l12);
\draw[arrow, bend left=50, shorten >=2pt, shorten <=4pt] (1) to (l13);
\draw[arrow, bend right=50, shorten >=2pt, shorten <=4pt] (1) to (l14);
\draw[arrow, bend left=50, shorten >=2pt, shorten <=4pt] (4) to (l45);
\draw[arrow, bend right=50, shorten >=2pt, shorten <=4pt] (4) to (l47);
\draw[arrow, out=-10, in=-120, looseness=0.7, shorten >=2pt, shorten <=4pt] (4) to (l46);
\draw[arrow, bend right=50, shorten >=2pt, shorten <=4pt] (4) to (l14);

\end{tikzpicture}
        }
        \caption{First step: compute temporary node features}
        \label{fig:great_n}
    \end{subfigure}
    \hfill
    \begin{subfigure}{0.48\textwidth}
        \centering
        \resizebox{0.9\columnwidth}{!}{
        \begin{tikzpicture}[
    node/.style={circle, draw=black, fill=cyan!80, text=white, minimum size=1cm},
    arrow/.style={->, thick, red},
    d_arrow/.style={
    ->,
    line width=1.5pt,
    dashed,
    Dandelion!100!black,
    scale=1,              %
    every arrow/.append style={
        scale=0.7          %
    }
},
    every label/.style={black}
]

\node[node] (1) at (0, 0) {1};
\node[node] (2) at (-2.5, 1.5) {2};
\node[node] (3) at (-3, -1.5) {3};
\node[node] (4) at (3, 0) {4};
\node[node] (5) at (6.5, 1.5) {5};
\node[node] (6) at (8, 0) {6};
\node[node] (7) at (6.5, -1.5) {7};

\draw[-, thick] (1) -- node[pos=0.5, fill=white, inner sep=1pt] (l12) {$e_{12}$} (2);
\draw[-, thick] (1) -- node[pos=0.5, fill=white, inner sep=1pt] (l13) {$e_{13}$} (3);
\draw[-, thick] (1) -- node[pos=0.5, fill=white, inner sep=1pt] (l14) {$e_{14}$} (4);
\draw[-, thick] (4) -- node[pos=0.5, fill=white, inner sep=1pt] (l45) {$e_{45}$} (5);
\draw[-, thick] (4) -- node[pos=0.5, fill=white, inner sep=1pt] (l46) {$e_{46}$} (6);
\draw[-, thick] (4) -- node[pos=0.5, fill=white, inner sep=1pt] (l47) {$e_{47}$} (7);

\draw[arrow, bend right=50, shorten >=2pt, shorten <=4pt] (l14) to (1);
\draw[arrow, bend left=50, shorten >=2pt, shorten <=4pt] (l14) to (4);

\end{tikzpicture}
        }
        \caption{Second step: combine temporary node features}
        \label{fig:great_n2}
    \end{subfigure}
    \caption{Overview of the two steps of GREAT (node-based)}
    \label{fig:comparison2}
\end{figure*}

\subsubsection*{GREAT NB vs. GREAT NF}
The advantages and disadvantages of both GREAT variants focus mostly on technical implementation aspects. While both GREAT variants were implemented in Python using Pytorch \citep{NEURIPS2019_9015}, the GREAT NB implementation requires additional message-passing operations by Pytorch Geometric \citep{Fey/Lenssen/2019} which makes the architecture slightly slower. However, the GREAT NB implementation generalizes more easily to non-complete, sparse graphs. 
Moreover, we hypothesize that the NB variant ensures a more considerable information exchange between adjacent edges compared to the purely edge-based variant. \rev{We note that NB outperforming NF in our experiments does not undermine the core contribution of GREAT: both variants share the same fundamental property of accepting only an edge feature matrix as input. The NF variant is the purer realisation of edge-input processing; NB introduces intermediate node aggregations purely for implementation convenience and better empirical performance, but these temporary node features are discarded after each layer and never passed forward, keeping the architecture fundamentally different from standard node-based GNNs.}
The code (as well as the trained models and datasets of the experiments in Section \ref{sec:experiments}) \rev{is publicly available at \url{https://github.com/attila-lischka/GREAT}.}

\subsection{Mathematical Formulations} \label{GREAT_math}
We now describe the mathematical equations defining the internal operations of GREAT.
Similar to the original transformer architecture of \cite{vaswani2017attention}, GREAT consists of two types of sublayers: attention sublayers and feedforward sublayers.
We always alternate between attention and feedforward sublayers.
The attention sublayers can be node-based (with temporary nodes features) or completely node-free. 
Using the respective sublayers leads to overall node-based or node-free GREAT. A visualization of the block stacking can be found in Figure \ref{fig:great_layer_block}.

\textbf{GREAT NB, Attention Sublayers}:
For each node $i$ in the graph, we utilize the edges features $e^{\ell -1}_{.,.} \in \mathbb{R}^d$ to compute a temporary node feature $x_i^\ell \in \mathbb{R}^d$ in layer $\ell$ as
\begin{equation*}
x_i^{\ell} = W_4 \sum_{j \in N(i)} (\alpha'_{i,j} W'_1 e_{i,j}^{\ell-1} || \alpha''_{i,j} W''_1 e_{j,i}^{\ell-1} || \alpha'_{j,i} W'_1 e_{j,i}^{\ell-1} || \alpha''_{j,i} W''_1 e_{i,j}^{\ell-1})
\end{equation*} 
where the attention scores are calculated using
\begin{align*}
\alpha'_{i,j} &= \text{softmax}\big(\frac{(W'_2 e_{i,j}^{\ell-1})^\top W'_3 e_{i,j}^{\ell-1}}{\sqrt{d}}\big)
=\frac{\exp\big(\frac{(W'_2 e_{i,j}^{\ell-1})^\top W'_3 e_{i,j}^{\ell-1}}{\sqrt{d}}\big)}{\sum_{k \in N(i)}\exp\big(\frac{(W'_2 e_{i,k}^{\ell-1})^\top W'_3 e_{i,k}^{\ell-1}}{\sqrt{d}}\big)} 
\end{align*}
\begin{align*}
\alpha''_{i,j} &= \text{softmax}\big(\frac{(W''_2 e_{j,i}^{\ell-1})^\top W''_3 e_{j,i}^{\ell-1}}{\sqrt{d}}\big) =\frac{\exp\big(\frac{(W''_2 e_{j,i}^{\ell-1})^\top W''_3 e_{j,i}^{\ell-1}}{\sqrt{d}}\big)}{\sum_{k \in N(i)}\exp\big(\frac{(W''_2 e_{k,i}^{\ell-1})^\top W''_3 e_{k,i}^{\ell-1}}{\sqrt{d}}\big)} 
\end{align*}
Note that we compute two types of attention scores (for outgoing and ingoing edges) and concatenate the resulting four values (each edge adjacent to node $i$ appears once weighted by its score normalized at node $i$ and once weighted by its score normalized at node $j$), which are then projected back to dimension $d$ by $W_4$ to form the temporary node feature.
This allows GREAT to differentiate between incoming and outgoing edges which, e.g. in the case of asymmetric TSP, can have different values.
If symmetric graphs are processed (where $e^\ell_{i,j} = e^\ell_{j,i}$ for all nodes $i,j$) we can simplify the expression to only one attention score.

The temporary node features are concatenated and mapped to the hidden dimension again to compute the actual edge features of the layer.
\begin{equation}
e^\ell_{i,j} = W_5 (x^\ell_i || x^\ell_j ) \nonumber
\end{equation}
where $W'_1, W''_1, W'_2, W''_2, W'_3, W''_3 \in \mathbb{R}^{d \times d}, W_4 \in \mathbb{R}^{d \times 4d}, W_5 \in \mathbb{R}^{d \times 2d}$ are trainable weight matrices of suitable dimension (all linear maps additionally include a bias term), $d$ is the hidden dimension and $||$ denotes concatenation. $W'_1 e_{i,j}^{\ell-1}, W'_2 e_{i,j}^{\ell-1}$ and $W'_3 e_{i,j}^{\ell-1}$ are similar to the ``values'', ``keys'' and ``queries'' of the original transformer architecture.
For readability, we state all equations for a single attention head. In practice, we use $H$ heads of dimension $d/H$ each, scale the attention scores by $\sqrt{d/H}$ instead of $\sqrt{d}$, and concatenate the outputs of all heads before applying the output projection.

\textbf{GREAT NF, Attention Sublayers}:
Here, edge features of network layer $\ell$ are directly computed as
\begin{equation}
e_{i,j}^\ell = W_4(\alpha'_{i,j}W'_{1}e_{i,j}^{\scriptscriptstyle \ell - 1} || \alpha'_{j,i}W'_{1}e_{j,i}^{\scriptscriptstyle \ell - 1} || \alpha''_{j,i}W''_{1}e_{i,j}^{\scriptscriptstyle \ell - 1} || \alpha''_{i,j}W''_{1}e_{j,i}^{\scriptscriptstyle \ell - 1}  ). \nonumber
\end{equation}
Note that the edge feature consists of four individual terms that are concatenated and projected back to dimension $d$ by $W_4 \in \mathbb{R}^{d \times 4d}$. Due to the attention mechanism, these terms summarize information on all edges outgoing from node $i$, outgoing from node $j$, ingoing to node $j$, and ingoing to node $i$. The distinction between in- and outgoing edges is again necessary due to asymmetric graphs. The $\alpha'$ and $\alpha''$ scores are computed as for the node-based GREAT variant above.

Remark that the attention used in GREAT (NB and NF) could also be interpreted as a form of gating and it is more similar to the type of ``attention'' used in architectures like the attention free transformer proposed by \cite{zhai2021attentionfreetransformer} that do not use self-attention. 
Computing real self-attention scores (as in the original transformer paper \citep{vaswani2017attention}) would require us to compare each of the $\mathcal{O}(n^2)$ edges in a routing problem graph to its $\mathcal{O}(n)$ adjacent edges. 
This would result in calculating $\mathcal{O}(n^3)$ many attention scores (compared to $\mathcal{O}(n^2)$ in the current formulation) and lead to a considerable computational increase.

\textbf{Feedforward Sublayer}:
Like in the original transformer architecture, the feedforward layer has the form
\begin{equation} e^\ell_{i,j} = W_2 \text{ReLU}(W_1 e_{i,j}^{\ell-1} + b_1) + b_2 \nonumber
\end{equation}
where $W_1, b_1, W_2, b_2$ are trainable weight matrices and biases of suitable sizes.
Moreover, again like in \cite{vaswani2017attention}, the feedforward sublayers have internal up-projections, which temporarily double the hidden dimension before scaling it down to the original size.

We further note that we add residual layers and normalizations to each sublayer (attention and feedforward). Therefore, the output of each sublayer is (like in the original transformer architecture):

\begin{equation}
e^\ell_{i,j} = \text{LayerNorm}(e_{i,j}^{\ell-1} + \text{Sublayer}(e_{i,j}^{\ell-1})) \nonumber
\end{equation}

\begin{figure}
\centering
\begin{tikzpicture}[
    block/.style={rectangle, rounded corners, draw=black, fill=blue!20, 
                  text width=10em, text centered, minimum height=1.5em},
    smallblock/.style={rectangle, rounded corners, draw=black, fill=green!20, 
                       text width=10em, text centered, minimum height=1.5em},
    input/.style={rectangle, draw=black, fill=red!20, text width=10em, 
                  text centered, minimum height=1.5em},
    line/.style={-Stealth, thick},
    framed/.style={draw=black, inner sep=1em},
    vertical text/.style={rotate=90, text centered, font=\bfseries},
    ]

    \node [framed] (frame) {
        \begin{tikzpicture}[
            block/.style={rectangle, rounded corners, draw=black, fill=red!20, 
                          text width=10em, text centered, minimum height=1em},
            ff/.style={rectangle, rounded corners, draw=black, fill=blue!20, 
                          text width=10em, text centered, minimum height=1em},
            smallblock/.style={rectangle, rounded corners, draw=black, fill=yellow!20, 
                               text width=10em, text centered, minimum height=1em},
            input/.style={rectangle, draw=black, fill=blue!20, text width=10em, 
                          text centered, minimum height=1em},
            line/.style={-Stealth, thick},
            ]
    \node [] (input) {Edge Feature Inputs};

    \node [block, above=1.5em of input] (mha) {GREAT Sublayer};
    \node [above=0.6em of input] (dummy) {};
    \node [smallblock, above=1.5em of mha] (addnorm1) {Add \& Norm};
    
    \node [ff, above=1.5em of addnorm1] (ffn) {Feed Forward};
    \node [above=0.6em of addnorm1] (dummy2) {};
    \node [smallblock, above=1.5em of ffn] (addnorm2) {Add \& Norm};

    \draw [line] (input) -- (mha);
    \draw [line] (mha) -- (addnorm1);
    \draw [line] (addnorm1) -- (ffn);
    \draw [line] (ffn) -- (addnorm2);
    
    \draw [line] (dummy.south) -- ++(2.5,0) |- (addnorm1.east);
    \draw [line] (dummy2.south) -- ++(2.5,0) |- (addnorm2.east);
    
    \node [above=1em of addnorm2] (annotation) {Edge Embeddings};
    \draw [line] (addnorm2.north) -- (annotation);

           \end{tikzpicture}
    };

    \node [vertical text, left=2.5em of mha.west] (verticaltext) {GREAT Layer $\times n$};

\end{tikzpicture}
\caption{A GREAT layer with sublayers and normalizations}
\label{fig:great_layer_block}
\end{figure}

\section{GREAT-Based Reinforcement Learning Framework}
\label{sec:framework}

\subsection{Framework Overview}
To optimize the parameters of our GREAT model, we build an end-to-end RL framework visualized in Figure \ref{fig:great_framework}.
Our framework follows the encoder-decoder approach (where GREAT serves as the encoder and the multi-pointer network adapted from \cite{jin2023pointerformer} as the decoder).
GREAT processes the input edge features (e.g., the $106 \times 106$ energy matrix for EVRP) and produces edge encodings that reflect how ``promising'' an edge is to be part of a high-quality solution tour.
The encodings are passed to the decoder which constructs a complete tour autoregressively.

\subsubsection{Intuition of the Framework}
\begin{figure*}
\centering
\begin{tikzpicture}

\node[anchor=south west] (img1) at (0,0) {\includegraphics[width=3.5cm]{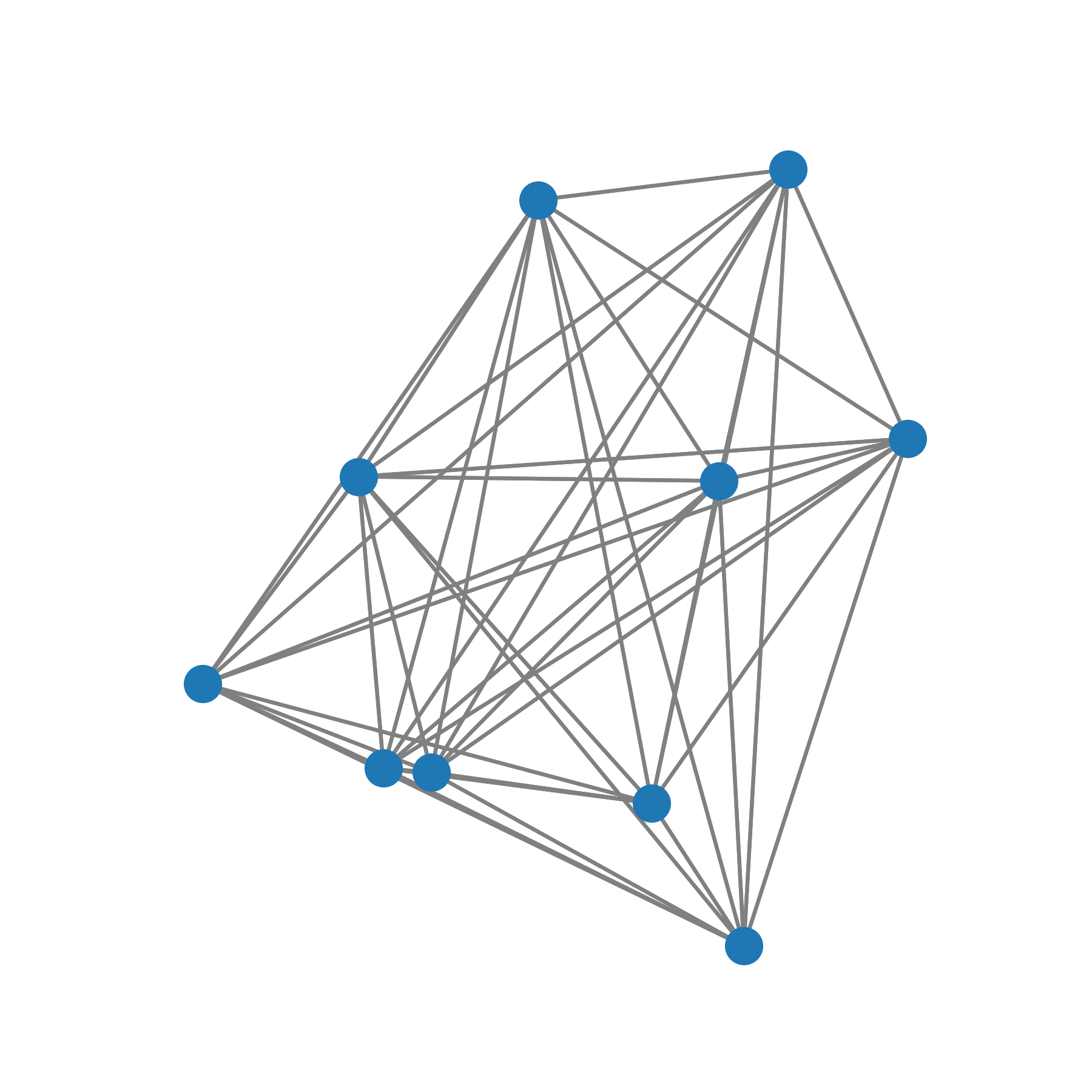}};
\node[above] at (img1.north) {\makebox[3cm]{\parbox{3cm}{\centering GREAT \\ Input}}};
\node[anchor=south west] (img2) at (3.5,0) {\includegraphics[width=3.5cm]{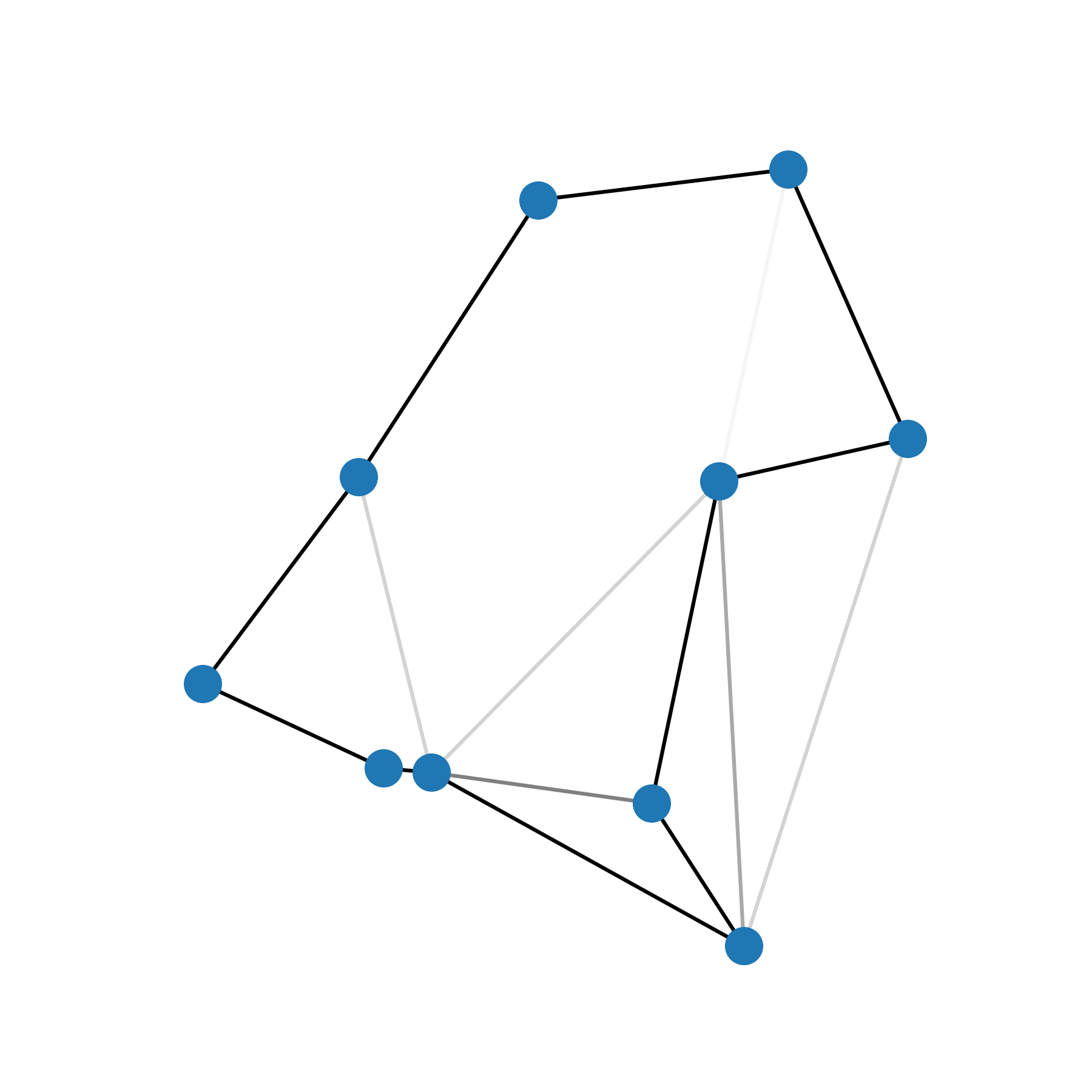}};
\node[above] at (img2.north)  {\makebox[3cm]{\parbox{3cm}{\centering GREAT \\ Edge Encodings}}};
\node[anchor=south west] (img3) at (7,0) {\includegraphics[width=3.5cm]{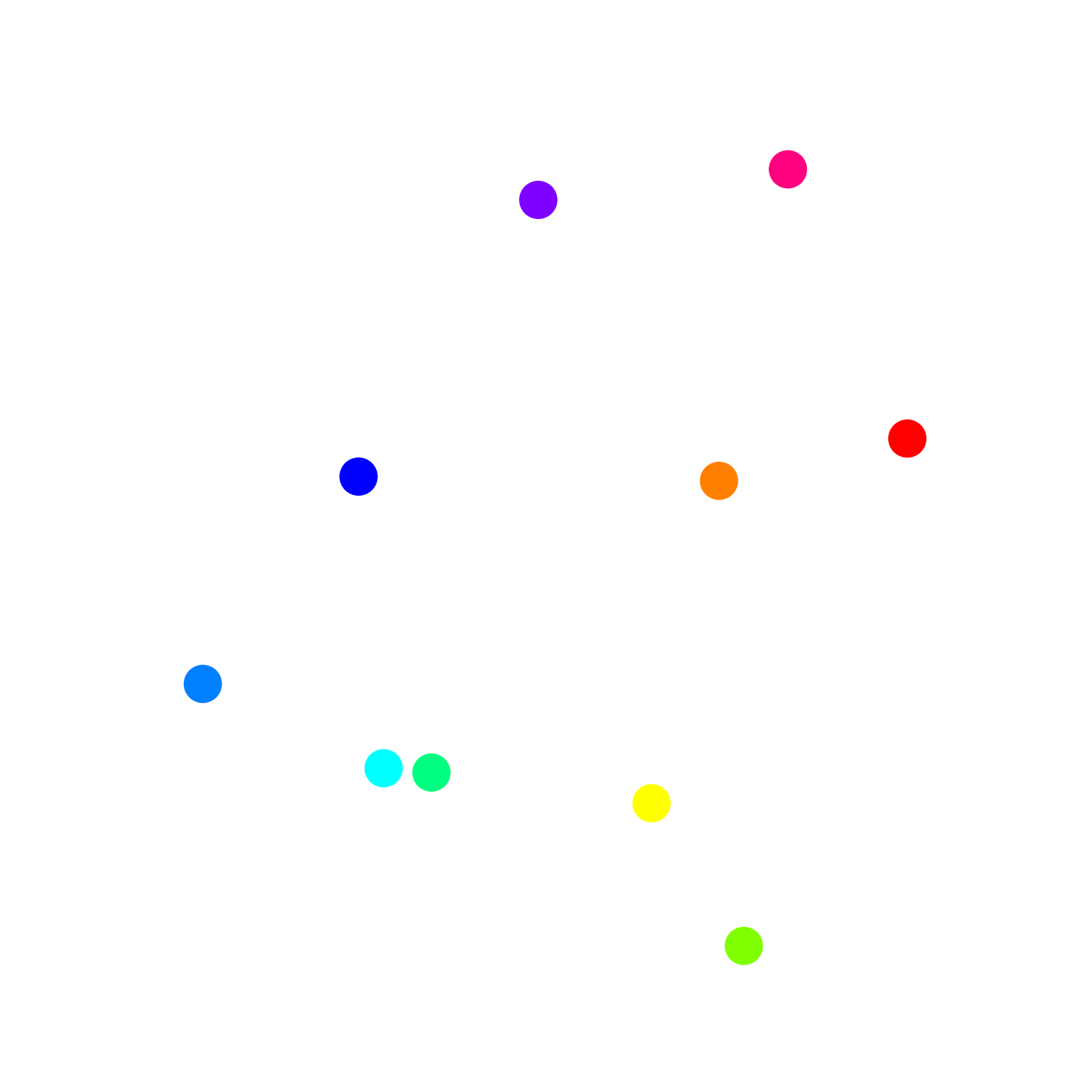}};
\node[above] at (img3.north) {\makebox[3cm]{\parbox{3cm}{\centering Transformed \\ Node Encodings}}};
\node[anchor=south west] (img4) at (10.5,0) {\includegraphics[width=3.5cm]{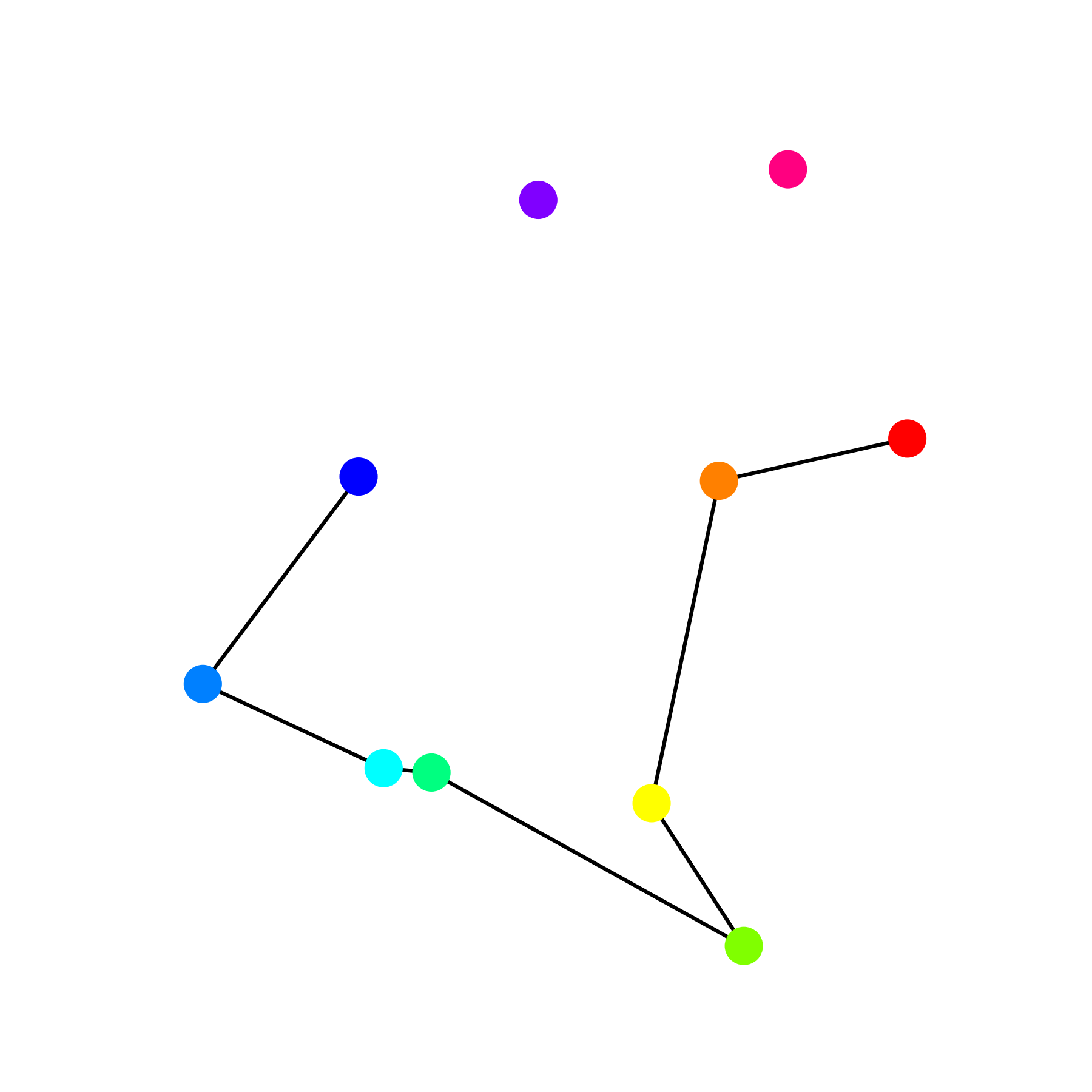}};
\node[above] at (img4.north) [yshift=0.06cm] {\makebox[3cm]{\parbox{3cm}{\centering Decoder \\ Solution Generation}}};

\draw[<-, thick] (img1.east) -- (img2.west) node[midway, above] {};
\draw[<-, thick] (img2.east) -- (img3.west) node[midway, above] {};
\draw[<-, thick] (img3.east) -- (img4.west) node[midway, above] {};

\end{tikzpicture}
\caption{The GREAT-based RL framework. The input edge feature matrix is encoded by GREAT into edge embeddings, transformed into node embeddings, and decoded into a solution tour.}
\label{fig:great_framework}
\end{figure*}

\begin{figure*}
\centering
    \begin{subfigure}[b]{0.30\textwidth}
        \centering
        \includegraphics[width=\textwidth]{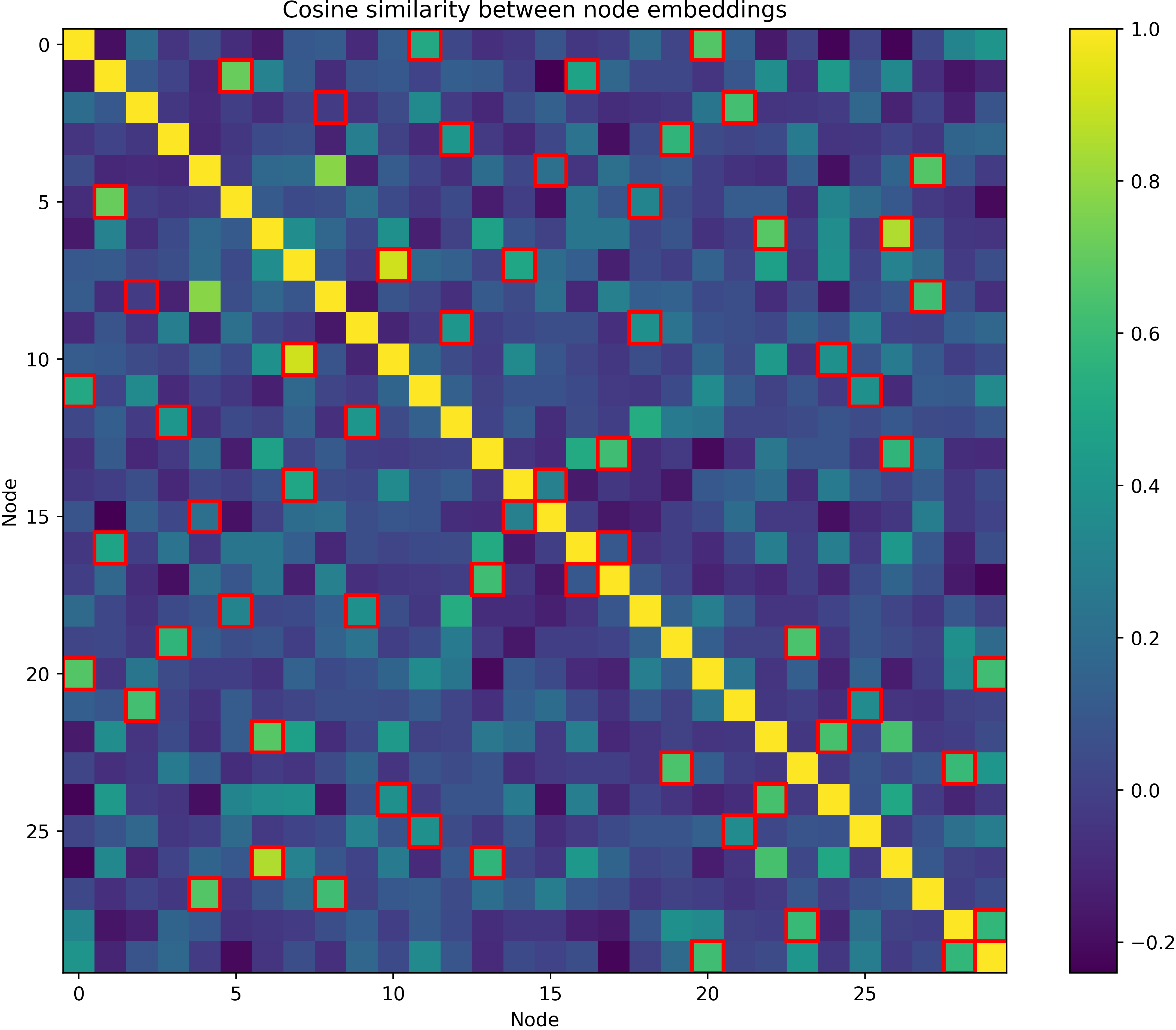}
        \caption{Cosine similarity, TSP30}
    \end{subfigure}
    \hfill
    \begin{subfigure}[b]{0.30\textwidth}
        \centering
        \includegraphics[width=\textwidth]{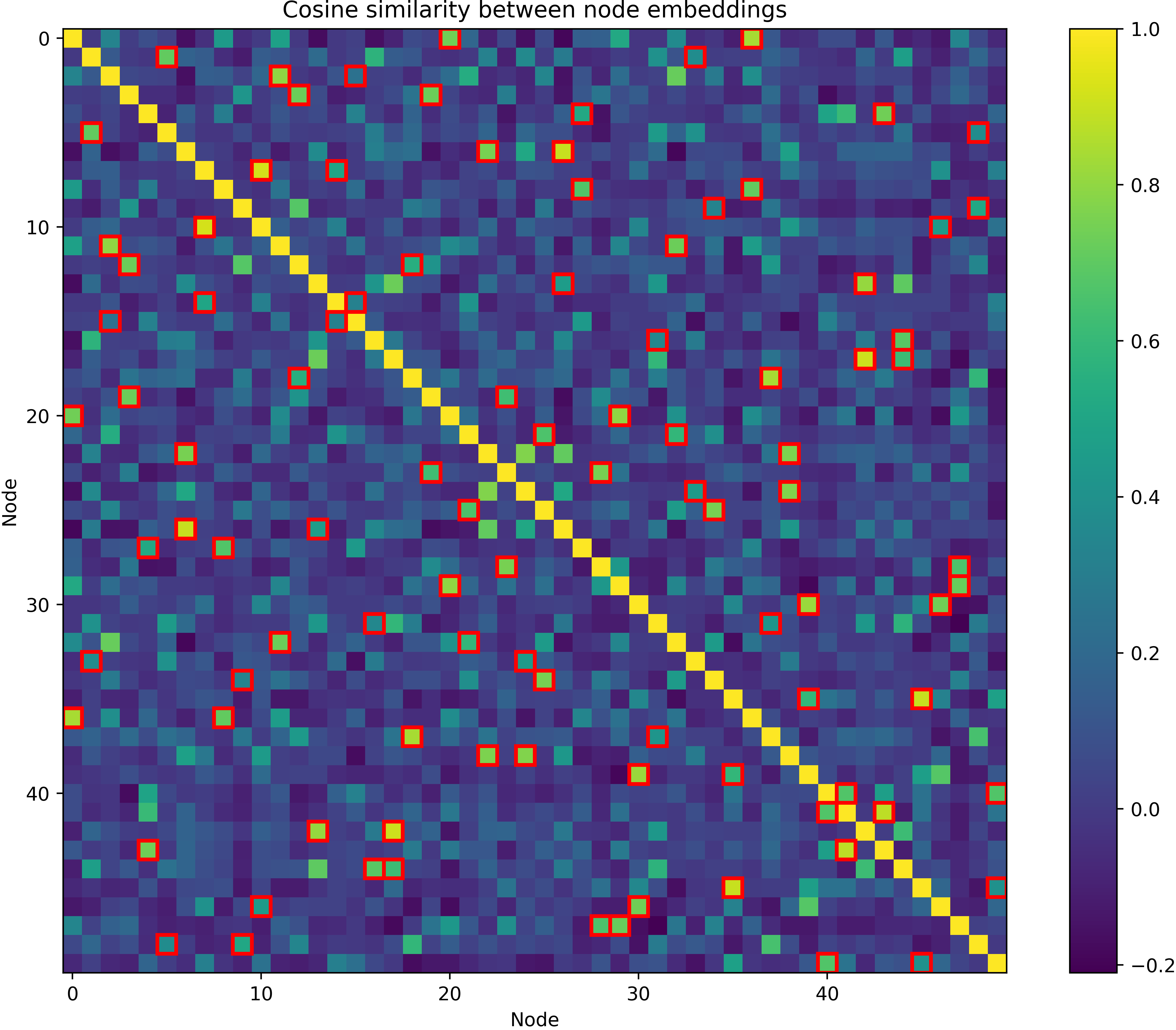}
        \caption{Cosine similarity, TSP50}
    \end{subfigure}
    \hfill
    \begin{subfigure}[b]{0.30\textwidth}
        \centering
        \includegraphics[width=\textwidth]{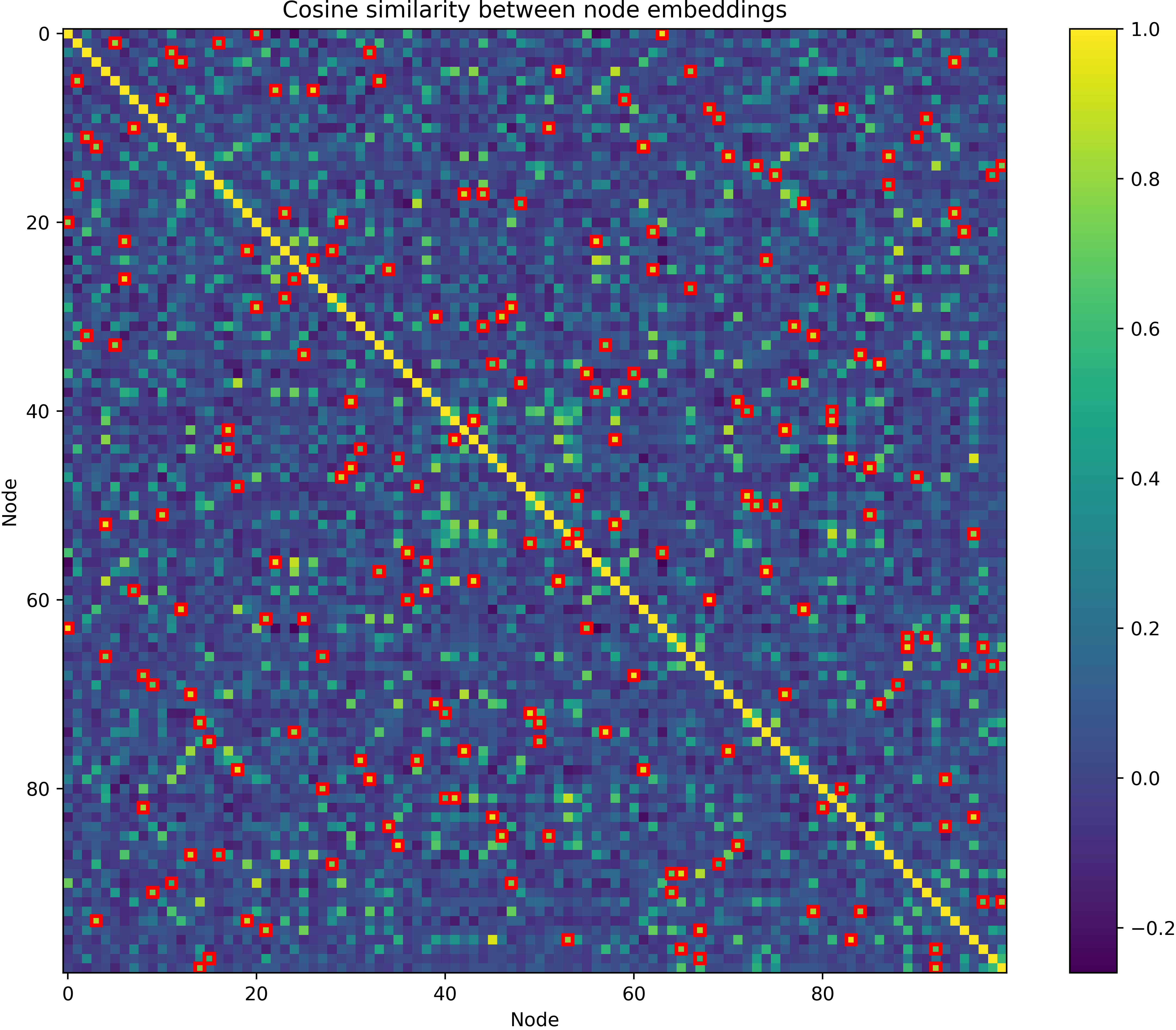}
        \caption{Cosine similarity, TSP100}
    \end{subfigure}

    \vspace{0.5cm}

    \begin{subfigure}[b]{0.30\textwidth}
        \centering
        \includegraphics[width=\textwidth]{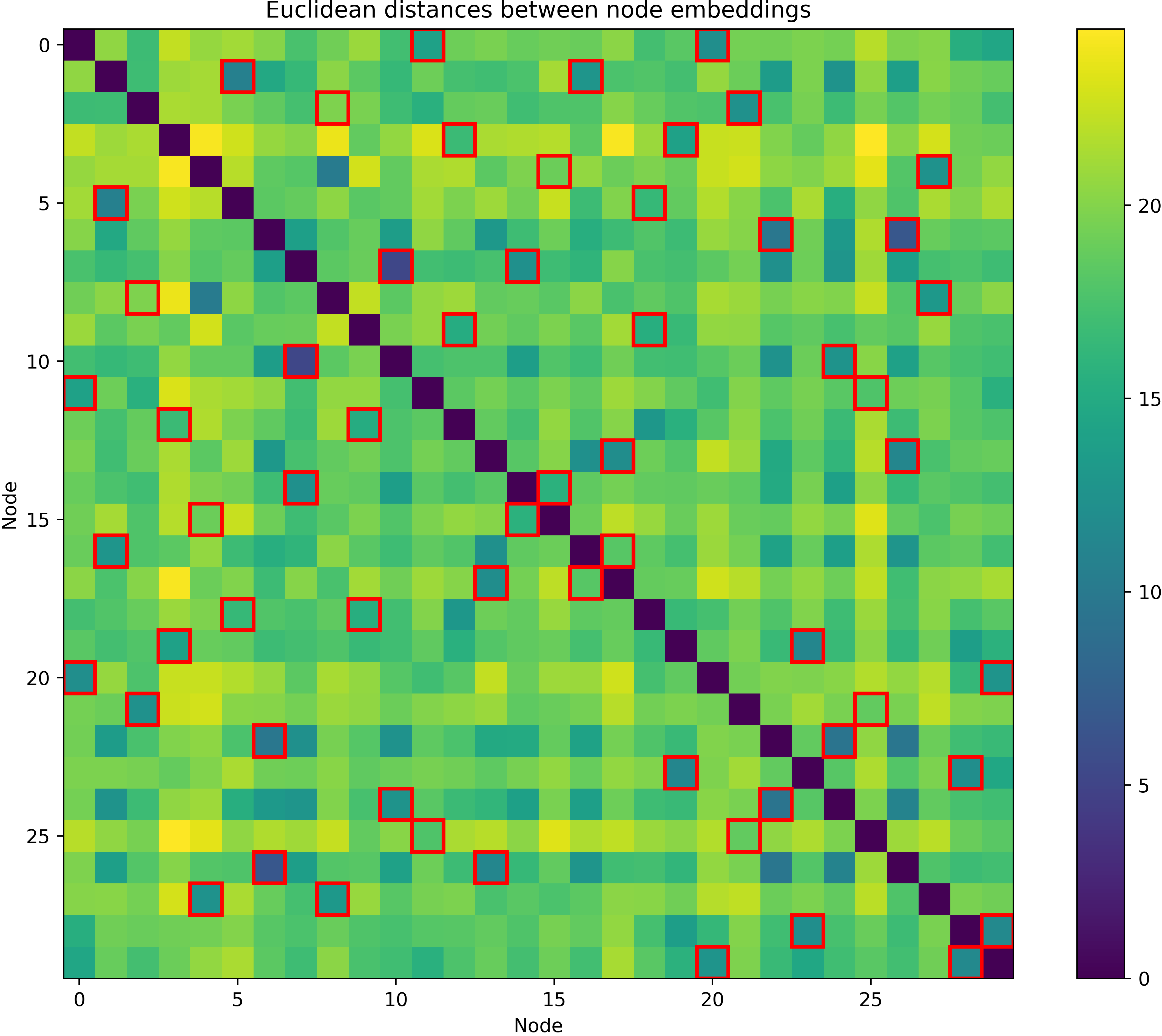}
        \caption{Euclidean distance, TSP30}
    \end{subfigure}
    \hfill
    \begin{subfigure}[b]{0.30\textwidth}
        \centering
        \includegraphics[width=\textwidth]{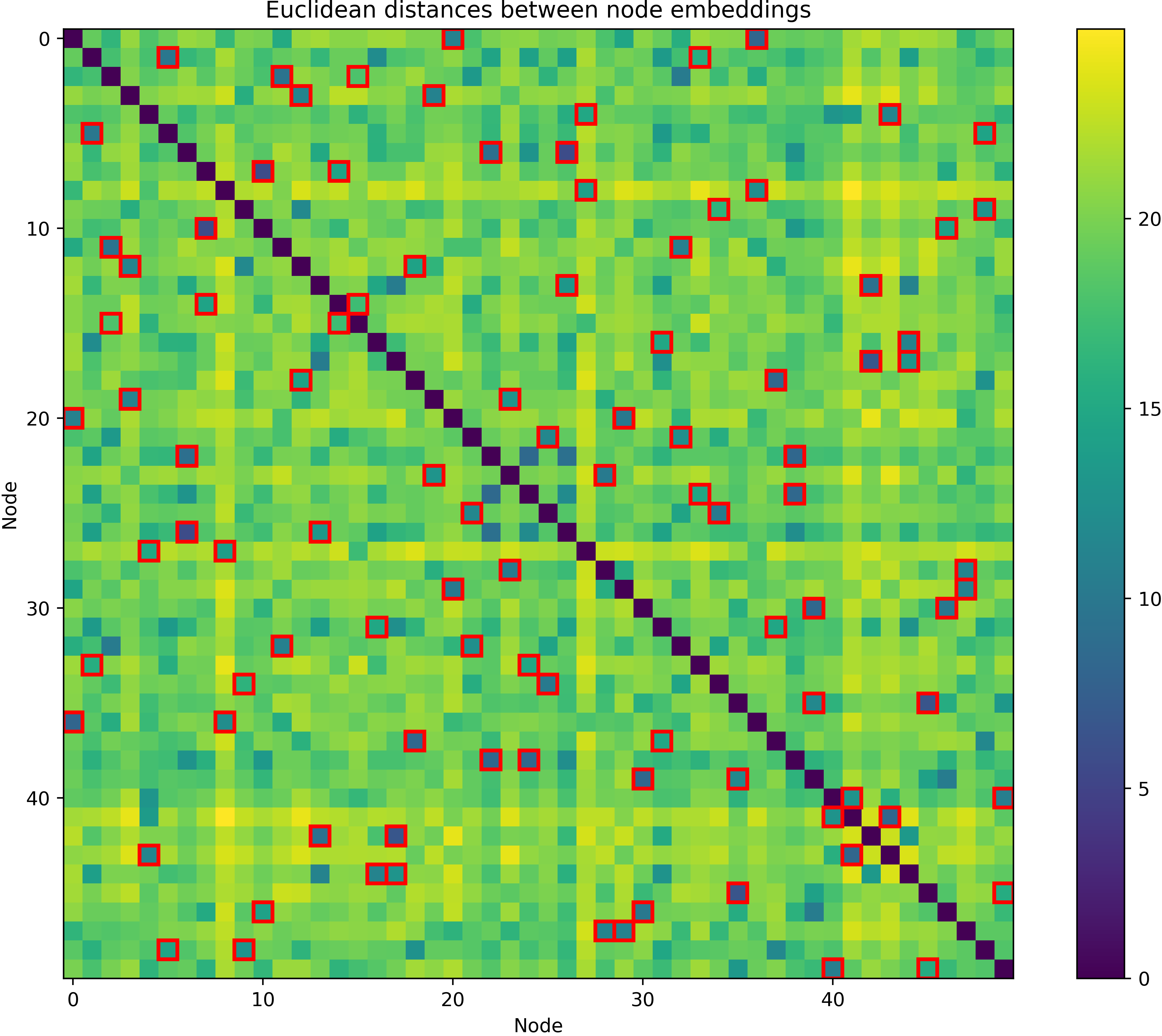}
        \caption{Euclidean distance, TSP50}
    \end{subfigure}
    \hfill
    \begin{subfigure}[b]{0.30\textwidth}
        \centering
        \includegraphics[width=\textwidth]{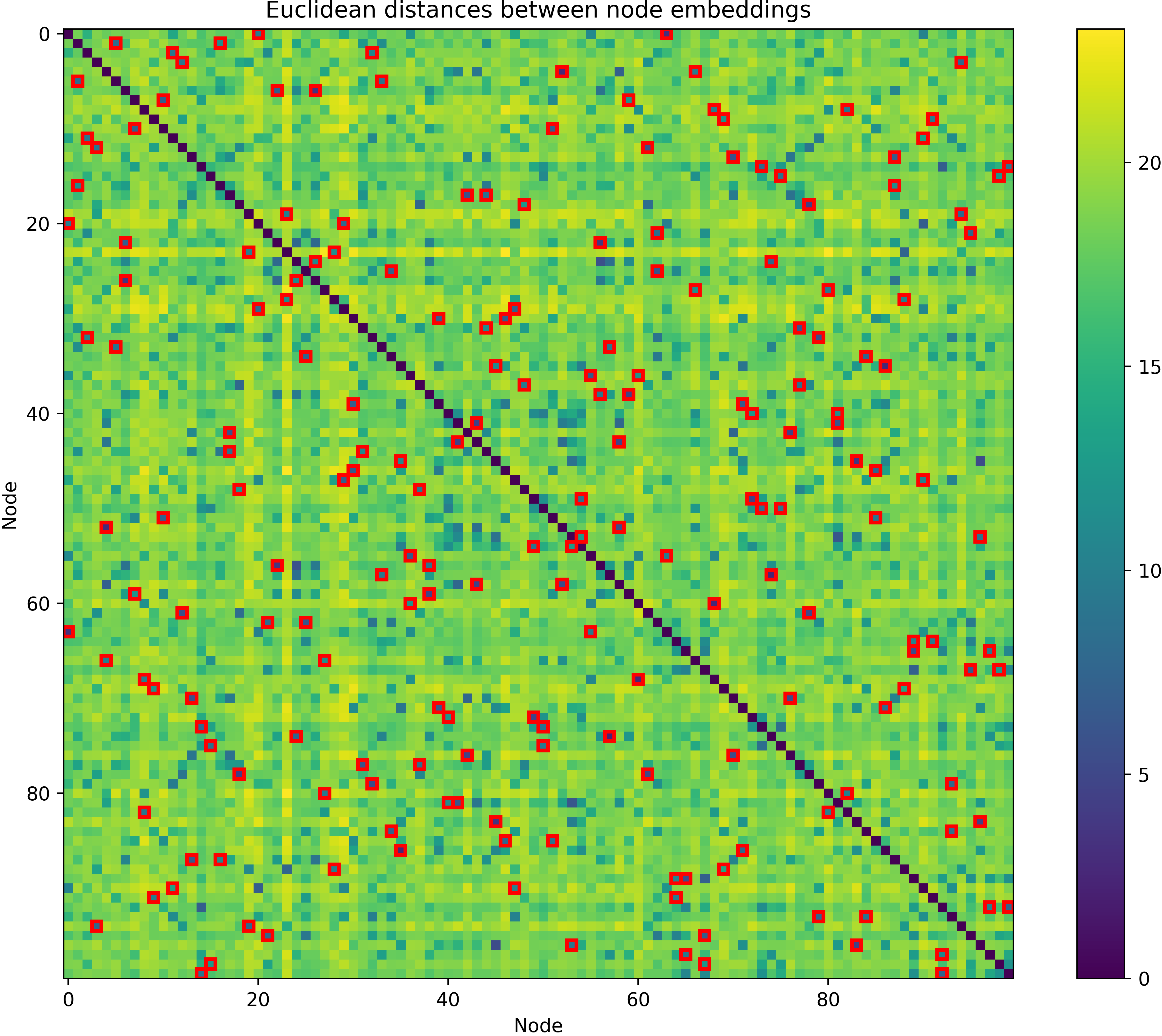}
        \caption{Euclidean distance, TSP100}
    \end{subfigure}

    \caption{Heatmaps of cosine similarities and Euclidean distances between GREAT node encodings. Red borders indicate edges present in the optimal TSP solution. High cosine similarity and low Euclidean distance between node encodings correlate strongly with solution edges.}
    \label{heatmaps_learning}
\end{figure*}

We visualize the framework in Figure \ref{fig:great_framework} and also show how GREAT potentially learns and passes on information.
For the sake of simplicity, we illustrate the idea of the framework for a Euclidean
TSP instance.
Non-Euclidean instances (and other routing problems) can be processed in the same way.
Given an input graph of edge features (e.g., the distances between TSP nodes), GREAT learns edge encodings that reflect the importance of individual edges (i.e., how promising they are to be part of the TSP solution).
In the visualization, we assign darker colors to the encodings of such promising edges.
Next, the edge encodings are transformed into node encodings by aggregation.
This is simply achieved by using a NB GREAT layer that returns the (otherwise temporary) intermediate internal node encodings (note that only the very last GREAT layer returns these node embeddings, otherwise, only edge encodings are passed on to the next layer). Since NF GREAT does not have temporary node embeddings, we use a final NB GREAT layer in our RL framework even when the other GREAT layers are node-free.
During this transformation, we assume that the edge information gets passed on and node encodings reflect which nodes are connected by important edges. 
Our assumption that nodes connected by important edges have similar encodings is backed up by the heatmap visualization in Figure \ref{heatmaps_learning}. 
The visualization shows Euclidean distances and cosine similarities between the vector encodings for the nodes in the TSP instance that are returned by the GREAT encoder to be passed on to the decoder.
Red frames around a tile in the heatmaps signal that there is an edge between the two nodes in the optimal TSP solution. 
We can see that for the cosine similarity, the red frames are mostly around tiles with high cosine similarity. 
Analogously, for the Euclidean distance heatmaps, we can see that the red frames are mostly around tiles with low distances.
For generating these heatmaps, we use a node-based GREAT encoder trained for TSP.
In Figure \ref{fig:great_framework}, we assign similar colors to nodes that are connected by edges deemed promising by GREAT to visualize the effect shown in the heatmaps.
In the last step of our encoder-decoder framework, we hypothesize that the decoder can construct solutions from these embeddings by iteratively selecting similar node encodings.

Although it would have been possible to create an edge-based decoder instead of transforming the edge-features produced by the GREAT encoder into node features, converting the edge encodings into node encodings allows us to freely plug in GREAT into existing training pipelines.
Further, it reduces the memory requirement from $\mathcal{O}(n^2)$ edges to $\mathcal{O}(n)$ nodes during decoding.

\subsection{Handling Node-Level Features in GREAT}

While GREAT operates on edge features as its input, it is still applicable to routing problems such as the EVRP that also involve node-level attributes such as customer demands.
This is because we can easily transform such node features into edge features.
Consider EVRP where a node $j$ has a demand $d_j$. We can simply associate demand $d_j$ to all edges $e_{i,j}$ in the graph. This is because we know that if we have an edge $e_{i, j}$ in the tour, we will visit node $j$ in the next step and therefore need a free capacity in our vehicle big enough to serve the demand of node $j$ which is $d_j$. An analogous extension works for other routing problems like CVRP or prizes for OP.

\subsection{Reinforcement Learning Setup}
\label{sec:great_encoding}
As explained, we use GREAT to encode our routing problem.
For this encoding, we pass a purely edge-based graph to GREAT. To capture all problem properties, for an edge $e_{i,j}$ we include the following edge features in the case of EVRP:
\begin{enumerate}
    \item Energy cost $\omega_{ij}$ in kWh when going from $i$ to $j$.
    \item Demand $d_j$ of node $j$ (if $j$ is the depot or a charger, the demand is set to 0)
    \item One-hot encoding whether edge $e_{i,j}$ leads to a customer, charger or depot.
    \item The energy cost from $i$ to $j$ compared to the cheapest edge cost from any node $k$ to $j$.
    \item The energy cost from $i$ to $j$ compared to the average edge cost from any node $k$ to $j$.
    \item The energy cost from $i$ to $j$ compared to the cheapest edge cost from node $i$ to any node $k$.
    \item The energy cost from $i$ to $j$ compared to the average edge cost from node $i$ to any node $k$.
\end{enumerate}
Note that TSP, CVRP and OP have different but analogous initial edge features to capture the properties of the corresponding problem.
\realrev{While storing each node feature in $n$ edges results in a certain redundancy, the real-world computational overhead is negligible since only the input layer is affected. To illustrate this, we consider the 9 dimensional edge input of EVRP consisting of 5 true edge features (feature 1., 4., 5., 6, and 7.) and 2 node level features (2. and 3. where 3. is itself a 3 dimensional encoding). All GREAT layers in our experiments (cf. Section \ref{sec:training_config}) have a hidden dimension of 128. We therefore map this 9 dimensional input to a hidden dimension of 128 by passing it through a simple linear layer. This linear layer has $9 \cdot 128 + 128 = 1{,}280$ parameters (including bias). If we didn't have any node features, the size of this hidden layer would be $5 \cdot 128 + 128 = 768$. Consequently, the additional node features result in a parameter overhead of $1,280-768 = 512$, a negligible number when compared to the overall architecture of approximately 1.42 million parameters (cf. Table \ref{table:training}). We conduct a small study that investigates the memory limits of our approach in Appendix~\ref{appendix_gpu_limits}. Furthermore, we acknowledge that it is also possible to design hybrid-like architectures that combine edge and node-level inputs. An example of such an architecture is MatNet \citep{kwon2021matrix}, which we compare against our approach in the experimental evaluation.}

After several GREAT layers, we obtain final encodings for all edges in the graph which are transformed by the attention-based aggregations of a GREAT NB layer resulting in node encodings.
These node encodings are passed to the decoder and used for autoregressive solution generation. Note that this decoding procedure corresponds to the execution of the MDP described in Section \ref{sec:mdp}. At every decoding step, a partial solution is extended by a further node to visit (customer, depot or charger). The decoding stops when a complete solution is generated.
The decoder provides probabilities for the next node to visit by comparing a query (that reflects the partial solution generated so far) with all node encodings.
We include the following information of a partial solution when building a query for the EVRP:
\begin{enumerate}
    \item The encoding of the node that represents the current vehicle location.
    \item The encoding of the depot node.
    \item The encoding of the whole EVRP graph (average over all node encodings).
    \item The encoding of all nodes visited so far (average over the corresponding node encodings).
    \item An encoding of the current vehicle load.
    \item An encoding of the remaining energy (in kWh) in the battery.
    \item An encoding of the battery size (in kWh).
\end{enumerate}
Again, similar queries can be built for TSP, CVRP and OP.
Solutions are constructed using POMO-based rollouts \cite{kwon2020pomo}, that is, after encoding, multiple solution trajectories are constructed simultaneously by selecting a random move in the first decoding step. During training, multiple rollouts allow us to obtain a robust baseline in the RL loss formulation. During inference, we use the best of all decoded rollouts as our final solution.
The reward for a solution is the negative total energy consumption (EVRP), negative total distance (TSP, CVRP), or total collected prize (OP).
We train the framework end-to-end using the REINFORCE algorithm \cite{williams1992simple}.

\subsection{Feasibility via Masking}
\label{sec:masking}

We ensure the validity of solutions generated by our framework through masking operations during decoding.
For example, for the EVRP, we mask customers that have already been visited. Moreover, we ensure that the vehicle capacity constraint is enforced by (temporarily) masking customer nodes that have demands that exceed the remaining vehicle capacity. We unmask such nodes after the next depot visit when enough capacity is available again. 
\realrev{Battery constraints are handled analogously: we temporarily mask nodes that are too ``far'' away and would violate the battery constraints. After visiting a charging node and recharging, nodes previously masked as unreachable become accessible again. This battery masking also contains a look-ahead that ensures it is still possible to visit a charger after visiting a customer. In other words, we do not allow moves to customers where we could get stuck with a battery capacity above the threshold but too low to reach a charger. Consequently, all solutions generated by our framework are feasible.}

Similar masking is done for the TSP, CVRP and OP settings to reflect the corresponding problem constraints.
A special case, however, arises for OP under the XASY distribution. As a result of the triangle inequality not holding, verifying if a node can still be visited within the allowed maximum travel distance would require solving a shortest-path problem at each decoding step (note that these shortest paths can change since it is not possible to visit nodes multiple times).
Therefore, in this setting, we relax the masking and instead set the reward to zero for infeasible solutions, which incentivizes the model to learn feasible actions.

\subsection{Training Configuration \realrev{and Hardware}}
\label{sec:training_config}

All models use a hidden dimension of 128, 5 GREAT layers, and 8 attention heads.
We train for 400 epochs with 25,000 instances per epoch, generating a fresh training dataset every 10 epochs.
We use the Adam optimizer \cite{DBLP:journals/corr/KingmaB14} with learning rate $10^{-4}$. 
\realrev{We set the batch size to 50 and use 100 POMO rollouts (cf. \cite{kwon2020pomo}) per instance unless stated otherwise. The node sets for the training, validation and test dataset for EVRP in San Francisco are furthermore disjoint.\footnote{\realrev{All random seeds that are used to generate datasets or initialize random number generators of software packages can be found in our code repository for reproducibility.}} Gothenburg and Vienna are only considered during testing (cf. Section \ref{sec:data_gen}). 
After every epoch of training, the model is evaluated on the validation dataset and the model with the best validation performance of any epoch is saved as our final model and used for testing. We show the validation curve developments of all trained models over the 400 training epochs in Appendix~\ref{appendix_val_curves}.}

Analogously to coordinate rotations done in \cite{kwon2020pomo} for Euclidean problems, \realrev{we apply a similar, scaling-based} data augmentation: during training, all pairwise distance costs in a batch are multiplied by a scalar drawn uniformly from $(0.5, 1.5)$.
For the EVRP, the battery capacity is scaled by the same factor to ensure an equivalent problem setting as in the original instance.
\realrev{While this augmentation does not have a significant effect on the model during training (cf. Appendix~\ref{appendix_aug}), it enables improved performance during inference} by solving multiple scaled (but equivalent) versions of a test instance. 

We note that we only train a single EVRP model (GREAT NB-based with $\times 9$ augmentation during inference) which is trained exclusively on San Francisco data with mixed battery sizes and demand distributions.
For TSP, CVRP and OP, we train a dedicated model for each data-distribution-problem combination, using GREAT NF and GREAT NB and test with different data augmentation levels ($\times 1$, $\times 9$ and $\times 33$) during inference. Further, for comparison, we train models on a mixture of all data-distributions.
As existing neural baselines, we also retrain a Pointerformer model \citep{jin2023pointerformer} for Euclidean problems and a MatNet model \citep{kwon2021matrix}. \rev{Crucially, all three models (GREAT, MatNet, Pointerformer) are trained within the identical RL pipeline and only the encoder component differs. These comparisons therefore constitute a direct ablation of the encoder: any performance difference is attributable solely to the choice of encoder architecture.} 
\realrev{
We acknowledge that GREAT receives four additional engineered edge features (cf. Section~\ref{sec:great_encoding}) compared to MatNet which operates on the raw edge costs. While this is an inherent limitation of the original MatNet architecture (since it can natively only process a single edge feature), we train an additional GREAT model that also only receives the raw edge costs in  Appendix~\ref{appendix_encoder_ablation} for a direct and fair comparison of those two encoders, confirming that GREAT achieves better performance in both scenarios.}

All training was performed on a single NVIDIA A40 (48\,GB VRAM).
Training times and parameter counts are reported in Table~\ref{table:training}. \realrev{Similarly, all validation, all testing and all ablations of neural models were done on the same NVIDIA A40 unless otherwise specified. Reported GPU runtimes correspond to wall-clock time and include the transfer time of the data to the processing device.}
All traditional baselines (such as HGS, \realrev{ALNS}, and LKH) and post-processing methods are run on an Apple M4 10-core CPU (at the time of the experiments among the fastest in consumer devices \cite{norem2024m4}). 
\realrev{Experiments where GREAT is explicitly reported to sequentially run on a CPU (as opposed to batched on a GPU; cf. Tables~\ref{tab:SF}--\ref{tab:Vienna}) are also executed on the same Apple M4.}

\begin{table}[!h]
\centering
\caption{Training times and model sizes (CVRP, 100 nodes)}
\label{table:training}
\begin{tabular}{c|c|c|c|c}
\textbf{Model} & Epoch & Val & Total & Params \\
\hline
 GREAT NB & $\simeq$ 7.5\,min & $\simeq$ 60\,s & $\simeq$ 60\,h & 1.42M \\
 GREAT NF & $\simeq$ 6.9\,min & $\simeq$ 50\,s & $\simeq$ 55\,h & 1.29M \\
 Pointerformer & $\simeq$ 2.1\,min & $\simeq$ 12\,s & $\simeq$ 19\,h & 1.09M \\
 MatNet & $\simeq$ 3.4\,min & $\simeq$ 22\,s & $\simeq$ 29\,h & 1.42M  \\
\end{tabular}
\end{table}

\subsection{Curriculum Learning for Large Instances}
\label{sec:cl}

The $\mathcal{O}(n^2)$ memory footprint of edge-based representations makes extensive training directly on large instances expensive. 
\realrev{We quantify this in Appendix~\ref{appendix_gpu_limits} where we show that on our NVIDIA A40 GREAT can only process a batch of size 10 during training for instances with 300 nodes. This makes extensive training on large datasets of large graphs infeasible.}
Therefore, to tackle larger instances, we employ a few-shot curriculum learning (CL) pipeline for TSP: starting from a model trained with 100 nodes, we fine-tune on progressively larger sizes
$[110, 121, 133, \ldots, 200, \ldots, 459, 500]$,
using only 2,000 instances per step and training for one epoch only (5 epochs for sizes 200 and 500).
This represents $0.02\%$ of the data used in original training, making the fine-tuning genuinely few-shot.
\realrev{We set the batch size to 16 for instances with less than 250 nodes, 8 for instances with more than 249 but less than 350 nodes and 4 for instances with more than 349 nodes. We keep the POMO size during training at 100 while setting it to the actual graph size during inference.}
Fine-tuned models for sizes 200 and 500 are denoted FT200 and FT500 in the result section, respectively. \rev{While the curriculum learning pipeline could in principle be applied to CVRP and OP as well, the TSP results already establish the general scalability of the approach. We leave CVRP/OP scaling to future work.}

\subsection{Post-Processing for EVRP Solutions}
\label{sec:postprocessing}

To further improve the solution quality of the EVRP tours generated by GREAT, we introduce an optional two-phase post-processing procedure.

\textbf{Phase 1 - Remove redundant stops.} For each depot or charger visit in the tour, we check whether it is feasible and also beneficial to remove it. Note that a stop is removed only if it is feasible to skip and doing so does not worsen the objective (due to the asymmetric energy costs without triangle inequality, some redundant stops can be energy-beneficial).

\textbf{Phase 2 - Local 2-opt on subtours.} We apply 2-opt moves to each customer subsequence between consecutive depot and charger visits until convergence. Since we modify the order of customers within a subtour only, we do not violate load or battery constraints (see Appendix~\ref{appendix_great_2opt}). Post-processing adds approximately $0.01$\,s of runtime per instance.

\section{Experiments}
\label{sec:experiments}

\subsection{EVRP Real-World Data Generation}
\label{sec:data_gen}

We generate EVRP instances using real-world road network data for three cities: San Francisco (United States), Vienna (Austria), and Gothenburg (Sweden).
Using OpenStreetMap \citep{openstreetmap}, we obtain road networks that are then enriched with elevation data from OpenTopography \citep{OpenTopography}.
For each city, we sample 106 nodes (1 depot, 5 chargers, 100 customers) with a probability that is proportional to the local population density, using the global human settlement population grid \citep{pesaresi2024advances}.
By this, we ensure that customer locations reflect realistic delivery patterns.
\realrev{We point out that all locations (depot, chargers and customers) are sampled per instance and are never fixed during neither training nor evaluation. 
Our model receives all information (pair-wise distances, node-type and demands) as its edge-input. This also means that the same set of customers can be solved with different charger locations. Solving an instance with different charger locations can be done efficiently in parallel in batches on GPUs and makes our model suitable for charging infrastructure planning. We conduct an ablation of such changing charger locations in Section~\ref{sec:chargers}.
}

Given the selected node locations, we compute the pair-wise fastest paths (w.r.t. travel time) between all nodes using Dijkstra's algorithm \citep{dijkstra2022note} on the road network. Using travel time as the metric to compute the shortest paths reflects real-world operations, as drivers would typically choose the fastest path. 
\realrev{Using energy-minimizing paths instead of fastest paths would result in an unrealistic setting. Unlike conventional internal combustion engine vehicles, electric vehicles are most efficient at low speeds \cite{HE2018907}. Therefore, an energy-minimizing path would prioritize roads with low speed limits, potentially through residential areas and de-prioritize highways. The resulting routes would consequently be unacceptable for the drivers and the operating company (due to much longer driving times) as well as for affected residents and traffic management. Consequently, we assume that drivers generally drive the fastest route between consecutive stops and optimize the order of those stops to minimize the energy consumption under this setting. This objective is stated in Section~\ref{sec:evrp_def}.} 
These fastest paths are translated into energy consumption values using the physics-based model of \cite{BASSO2019141}, resulting in a $106 \times 106$ energy matrix that forms the primary input to our model.

We note that these energy costs are asymmetric and can take negative values. Further, they do generally not respect the triangle inequality. This is because a fastest path from $i$ to $j$ might lead over a highway which results in higher energy consumption than a path from $i$ to $k$ to $j$ which potentially leads through a city center with lower speeds and lower energy consumption.

We consider four battery sizes: $B \in \{15, 20, 30, 40\}$\,kWh (note that we use 60\% of these nominal values only), three customer demand ranges: $d_v \sim \text{Uniform}\{1,\ldots,9\}$, $\{11,\ldots,19\}$, and $\{21,\ldots,29\}$, and a fixed vehicle load capacity of $Q = 50$.
Higher customer demands result in more depot returns and (due to the resulting increased energy demand) more frequent charging which increases problem complexity.

\begin{figure*}
\centering
    \begin{subfigure}[b]{0.30\textwidth}
        \centering
        \includegraphics[width=\textwidth]{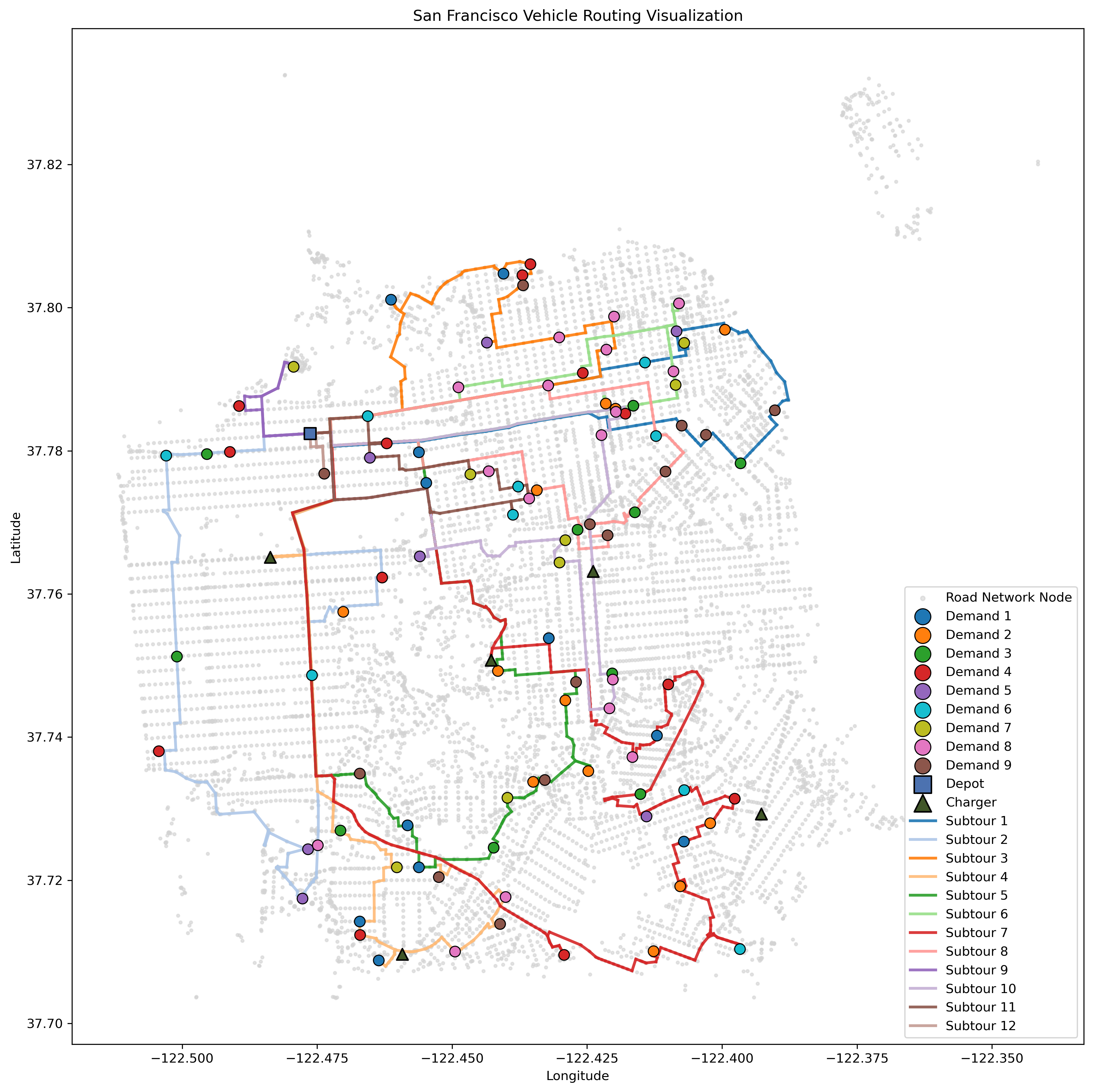}
        \caption{San Francisco}
        \label{sf_example_figure}
    \end{subfigure}
    \hfill
    \begin{subfigure}[b]{0.30\textwidth}
        \centering
        \includegraphics[width=\textwidth]{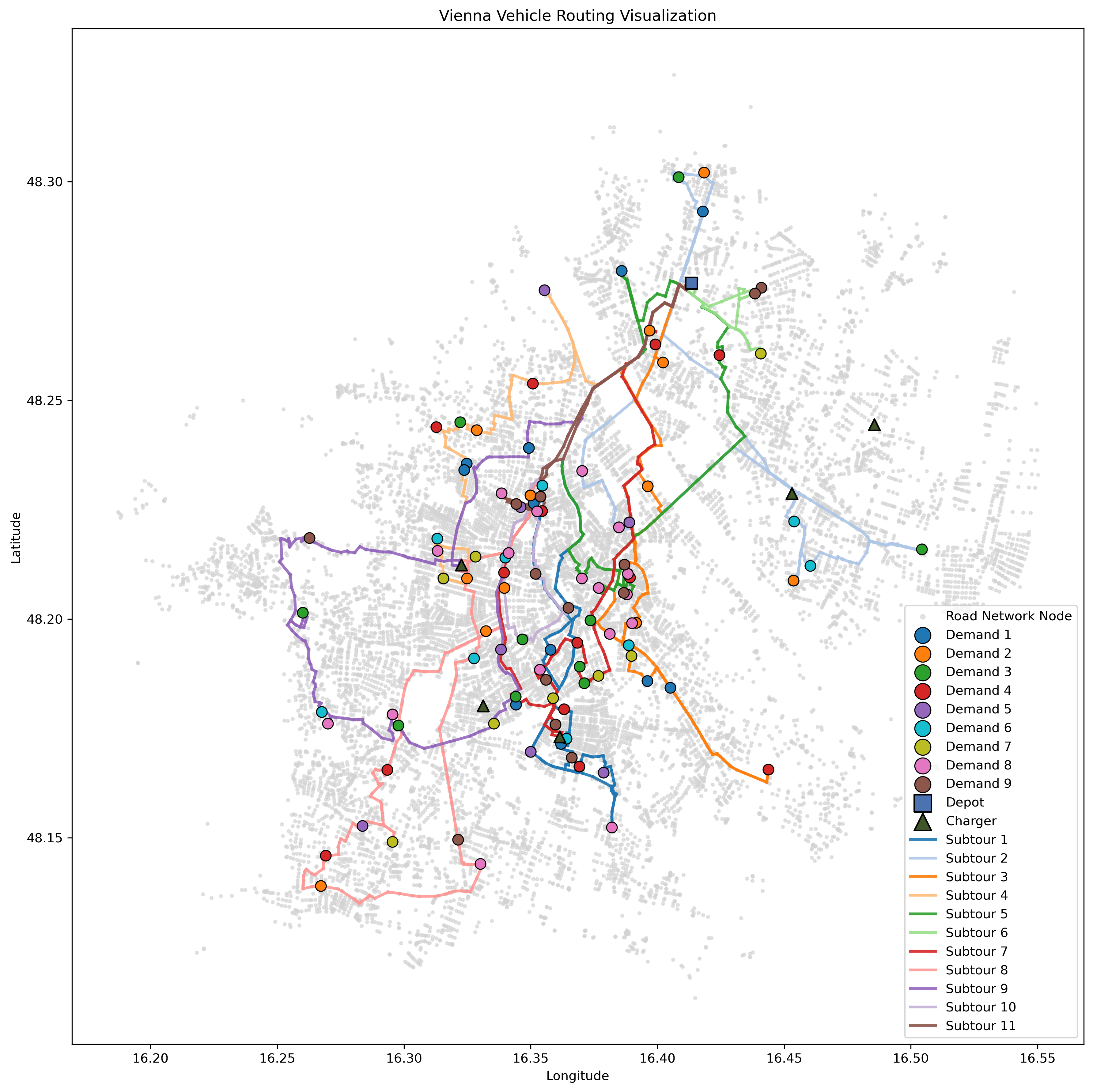}
        \caption{Vienna}
        \label{vienna_example_figure}
    \end{subfigure}
    \hfill
    \begin{subfigure}[b]{0.30\textwidth}
        \centering
        \includegraphics[width=\textwidth]{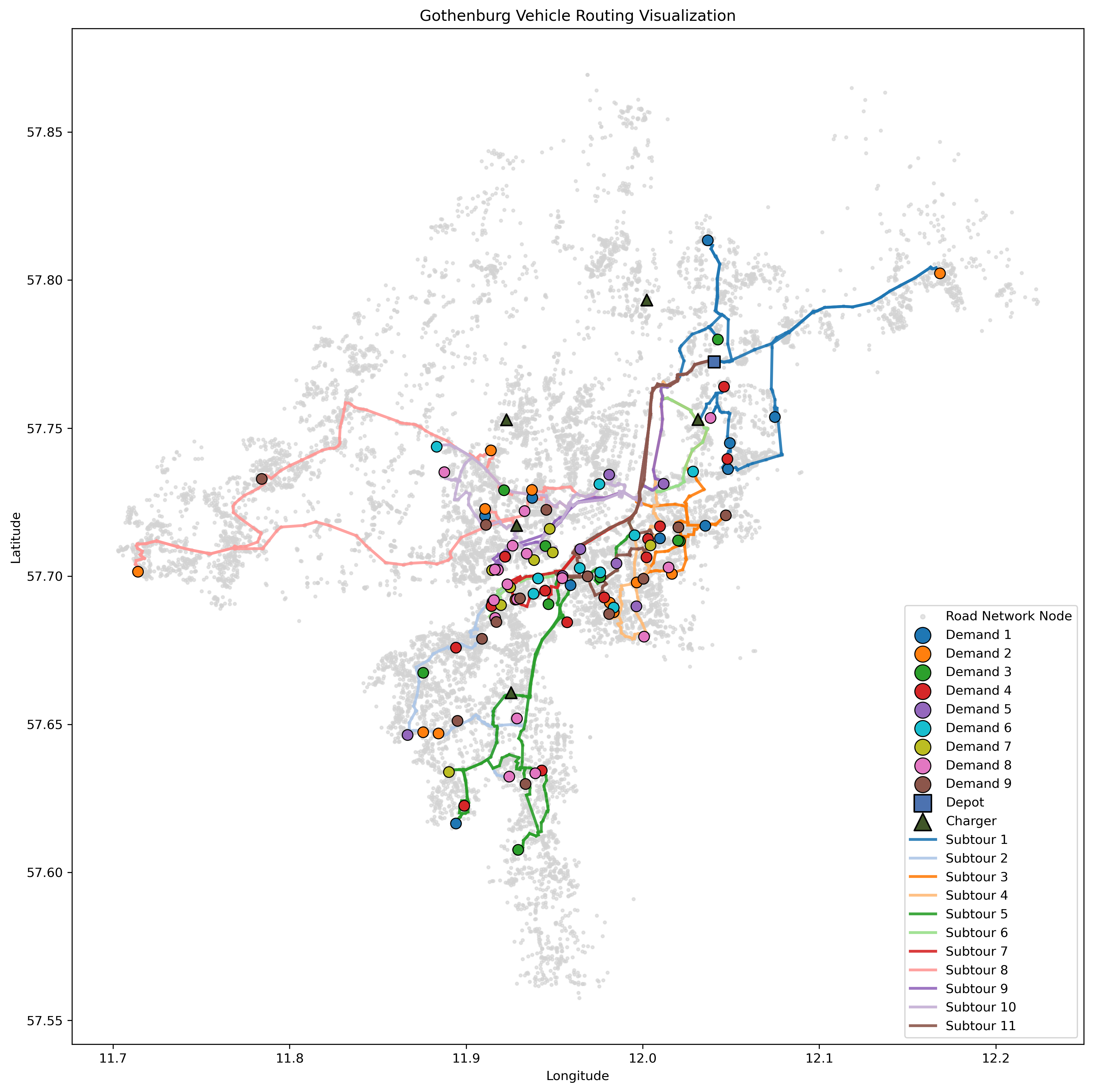}
        \caption{Gothenburg}
        \label{gothenburg_example_figure}
    \end{subfigure}
    \caption{EVRP instances with 100 customers, 5 chargers, and 1 depot across three cities. Node placement reflects population density.}
    \label{evrp_routing_problems}
\end{figure*}

While EVRP instances from San Francisco are used both for training and (with a disjoint node set) for testing, Vienna and Gothenburg instances are used exclusively for testing to investigate generalization capabilities.
Example EVRP instances for the three cities are visualized in Figure~\ref{evrp_routing_problems}.

\subsection{\realrev{EVRP Baselines}}
\label{sec:evrp_baselines}

\subsubsection{\realrev{Two-Stage} Baseline \realrev{(HGS)}}
As an EVRP baseline, we use HGS \citep{vidal2012hybrid}, which is a state-of-the-art CVRP solver, by extending it with a greedy charger insertion heuristic (see Appendix~\ref{appendix_hgs}).
\rev{HGS remains the strongest available general-purpose baseline for capacitated vehicle routing; it is used as the standard benchmark in virtually all recent ML-based routing papers \citep{drakulic2023bq, yi2026radar,son2026rrnco}. No out-of-the-box EVRP-specific exact solver is available that can handle the scale of our test instances (100 customers, hundreds of test instances per configuration); exact methods would require hours or days per single instance at this scale, making batch evaluation infeasible.} 

After solving the CVRP relaxation with HGS, we greedily insert charger stops wherever the limited battery capacity would lead to a depletion.
We note that this two-stage approach is simple and effective for settings with few charging stops. 
However, it is generally suboptimal when many charging stops are required, since routing and charging decisions are decoupled.

\rev{We do not apply the 2-opt post-processing step to HGS solutions, because HGS already incorporates 2-opt moves in the local search during its own optimization procedure. Therefore, the subtours it produces are already locally optimal with respect to such moves and applying an additional 2-opt pass would increase HGS runtime without improving solution quality. The combined GREAT + 2-opt runtime (at most \realrev{7.7}\,s per 100 instances) remains approximately $60\times$ faster than HGS alone \realrev{when using batched GPU processing}.}

\subsubsection{\realrev{ALNS Baseline with Joint Routing and Charging}}
\label{sec:alns}
\realrev{Our two-stage HGS baseline is at a disadvantage compared to GREAT since GREAT learns to jointly optimize charging and routing while the two-stage baseline is decoupled by construction. Therefore, we also introduce an additional adaptive large neighborhood search (ALNS) \citep{ropke2006adaptive, pisinger2007general} baseline that, just like GREAT, optimizes charging and routing holistically.
ALNS is a strong metaheuristic used to solve many different VRP variants and has successfully been used to solve EVRP in the past \citep{GOEKE201581, KESKIN2016111, HIERMANN2016995}. It works by iteratively applying destroy and repair operators to improve an existing solution until a convergence or time limit criterion is met.
}

\paragraph{\realrev{Representation and evaluation}}
\realrev{We represent a solution as a single giant tour $(0,\dots,0)$, that contains every customer exactly once, every charger zero or multiple times and as many intermediate depot visits as required. The vehicle's battery is reset every time it reaches a charger and its load is reset every time it reaches the depot. Therefore, there are two different types of non-aligned subtours which we will call ``battery legs'' and ``capacity segments''. Such battery leg can span several capacity segments and vice versa and, as a result, it is not possible to represent the overall tour as a simple ``list of lists''.
A candidate tour is evaluated following the MDP of Section~\ref{sec:mdp}. Going from node $i$ to $j$ changes the SoC $\omega_{ij}/B$ (with potential clipping in case the battery gets full on downhill roads), charger nodes reset the battery and the load of customer nodes gets accumulated until it is reset at the depot. 
The objective is to minimize the accumulated consumed energy, exactly as for GREAT.
The starting solution of ALNS is a simple nearest-neighbor solution construction w.r.t. the energy cost and returning to the depot when needed. Then, charging stops are inserted using a repair operator (described below). Optionally, our ALNS implementation also supports warm initialization with EVRP solutions generated by other methods. In our experiments, we will report \emph{ALNS HGS-init} which starts from our two-stage HGS baseline and \emph{ALNS GREAT-init} which starts from our post-processed GREAT solutions in addition to the ``pure ALNS'' experiments without warm initialization.
}

\paragraph{\realrev{Destroy and repair operators}}
\realrev{We consider two different operator pools for destruction for chargers and customers respectively, because customers are mandatory to visit and must be reinserted, whereas chargers are optional and do not have to be visited as long as the battery SoC is high enough.
For the customer operator pool, we consider random removal, worst removal (removes the customer $c$ between nodes $i,j$ with the highest gain $\omega_{ic}+\omega_{cj}-\omega_{ij}$), Shaw removal \citep{shaw1998using} (removes clusters of related customers based on energy consumption and demand weighted with 0.7 and 0.3 respectively) and segment removal (removes a whole sequence of customers \citep{christiaens2020slack}). To prevent getting stuck in the same set of removals, worst and Shaw removal use the randomized rank selection with determinism exponents $3$ and $6$ \citep{ropke2006adaptive}. 
For the chargers, it is possible to remove no charger, a random subset of existing chargers or all charging stops (resulting in a completely new charging plan).
To repair the customer constraints, one of the following repair operators is selected: greedy insertion, regret-$2$ insertion \citep{ropke2006adaptive}, or cheapest insertion subject to a fixed random customer order where the cost of an insertion of customer $c$ between nodes $i,j$ is $\omega_{ic}+\omega_{cj}-\omega_{ij}$. Candidate positions are filtered to respect vehicle capacities and, if no capacity-feasible position is available, a new subtour is added to the solution. Battery constraints are not enforced during the customer repair operations but restored afterwards.
}

\paragraph{\realrev{Charging as a searched decision}}
\realrev{The battery constraints are ensured by a repair operator that iterates over the tour and inserts charging stops where needed. When inserting a charger, the cheapest option is chosen in $85\%$ of cases while in the remaining $15\%$ a random charger is chosen from the best 3 candidates.
Through the combination of charger destroy operators and acceptance criterion it is ensured that alternative charging plans are generated. Without this, a given customer sequence would result in the same charging plan after every repair. 
After each iteration, we also apply a post-processing similar to the GREAT solution (cf. Section~\ref{sec:postprocessing}), where unnecessary charging and depot stops are removed if it improves the objective. Furthermore, we apply 2-opt and or-opt improvement operators on all subsequences between chargers and nodes. Similar to the GREAT post-processing, this ``local'' post-processing ensures that the feasibility of the overall tour is not violated.
}

\paragraph{\realrev{Search control and hyperparameters}}
\realrev{
A candidate solution is accepted using simulated annealing \citep{kirkpatrick1983optimization}. If the candidate solution is worse than the current solution by $\Delta$, we accept it with probability $\exp(-\Delta/T)$. The temperature is annealed as a function of the elapsed budget
\[
  T(t) \;=\; T_0 \Bigl(\tfrac{T_{\text{end}}}{T_0}\Bigr)^{t/t_{\max}}, \qquad T_{\text{end}}/T_0 = 10^{-3},
\]
where $T_0$ is chosen such that a solution that is 5\% worse has a $50\%$ chance of acceptance. In our experiments we run ALNS with three different (wall-clock) time budgets of 1\,s, 5\,s, and 60\,s per instance. 
To draw the operators, we use three separate wheels (customer destroy, charger destroy, repair) for roulette-wheel selection. We use scores $\sigma_1=33$, $\sigma_2=9$ and $\sigma_3=13$ for a new global best solution, an improvement on the current solution, and an accepted non-improving solution, respectively. We accumulate the scores over $100$ iterations and update the weights for the wheels using exponential smoothing $w \leftarrow (1-\rho)\,w + \rho\,(\text{score}/\text{usage})$ with $\rho=0.1$ and a floor of $10^{-3}$. In every iteration we remove a fraction $\mathcal{U}[0.10,0.30]$ of customers.
}

\subsection{EVRP Results}
\label{sec:evrp_results}

The results for the EVRP experiments are presented in Tables~\ref{tab:SF}, \ref{tab:Gothenburg}, and~\ref{tab:Vienna} for San Francisco, Gothenburg, and Vienna, respectively.
For each city, we consider several combinations of battery sizes and customer demands. For every combination we consider 100 test instances and report average energy consumption in kWh, average runtime w.r.t. the HGS baseline and total runtime for solving all instances.
\realrev{For ALNS, we report its standalone form as well as two warm-started variants with different time budgets.}
\realrev{We also provide Appendix~\ref{appendix_aug} where we show the performance of GREAT with and without augmentation during runtime and with and without post-processing. Compared to Tables~\ref{tab:SF}, \ref{tab:Gothenburg}, and~\ref{tab:Vienna} which only show average performance, we also provide paired per-instance confidence intervals, win rates and operational diagnostics in Appendix~\ref{appendix:stats}.}
Note that all results are achieved on data unseen during training. This is ensured by using disjoint node sets for San Francisco during training and testing to sample from. Moreover, data from Vienna and Gothenburg was not encountered at all during training to validate the generalization performance of the trained model.

A key observation is regarding the runtime of the different methods: while HGS requires 6-10 minutes for 100 instances, GREAT requires only 4.4-6.3 seconds, a speedup of approximately $60\times$ \realrev{using batched GPU processing}. Post-processing adds approximately 1 second in total.
\realrev{Neural models offer the advantage of GPU accelerated batch-processing over our traditional baselines (ALNS and HGS) that run on CPUs. Therefore, as an additional comparison, we also report the runtimes of GREAT when using sequential CPU processing instead on the exact same CPU that was used for the baselines. Those results can be found in Appendix~\ref{appendix_aug} and they show that even in this setting, GREAT solves the 100 instances in 25s (x1 augmentation) to 3.4min (x9 augmentation) on average compared to 5.7--10.9\,min for HGS.}

\realrev{Performance-wise,} GREAT generally performs best in settings with smaller batteries and higher customer demands, which correspond to settings with high energy demand that consequently require many charging stops.
In these operationally demanding settings, the holistic joint optimization of routing and charging of GREAT outperforms the two-stage HGS approach.
In settings with large batteries and low demands (resulting in fewer charging stops), HGS generally produces better solutions, as the decoupled approach is less penalized when charging decisions are less critical.

For San Francisco (training distribution), GREAT outperforms HGS in all configurations with demand ranges $\{11,\ldots,19\}$ and $\{21,\ldots,29\}$.
Post-processing extends GREAT's advantage, reaching up to $-8.78\%$ relative to HGS in the setting with 15\,kWh battery and demand $\{21,\ldots,29\}$.
\realrev{We observe that, due to reporting the average gap (the standard metric in learning-based routing literature) and average energy consumption, slight ``inconsistencies'' like in the San Francisco case with a 15kWh battery and demand range $\{1,\ldots,9\}$ can occur. In this setting, augmented GREAT with post-processing has a positive gap ($0.11\%$) while consuming slightly less energy than the HGS baseline ($26.64$kWh vs. $26.66$kWh). This can happen when one algorithm has a tendency to perform comparatively better or worse in settings where overall less or more energy is consumed.}
For Gothenburg and Vienna (unseen cities), GREAT with post-processing outperforms HGS in almost all configurations.
For Gothenburg, HGS is better than GREAT with post-processing in only a single configuration (40\,kWh battery and demand $\{1,\ldots,9\}$).
For Vienna, HGS leads only in three settings, i.e. with battery capacity of at least 20 kWh and demand $\{1,\ldots,9\}$.

\realrev{When evaluating the ALNS results (without warm initialization), we can see that ALNS performs best (w.r.t. HGS) on small battery sizes with high customer demand, underlining the importance of joint charging and routing optimizations. In large battery settings with low demands that require hardly any charging, HGS generally outperforms ALNS due to superior routing performance. ALNS typically requires 1min of runtime to roughly perform equally well or better than HGS, which, also corresponds to roughly the runtime of HGS itself (which is, however, unconstrained). In total, ALNS outperforms HGS 0 times with 1s budget, 6 times with 5s budget and 23 times with 60s budget (out of 26 combinations).}

\realrev{When comparing ALNS with 1s of runtime to GREAT (which, when running batched on GPUs, is still faster than ALNS), we can see that GREAT outperforms ALNS in all configurations. The same statement holds for ALNS with 5s of runtime. Even with 60s of time budget, GREAT outperforms ALNS on 29 out of 36 configurations (ALNS beats GREAT on some configurations with customer demand range $\{1,\ldots,9\}$).}

\realrev{We note, however, that ALNS is a very strong baseline when paired with warm initialization. Depending on the configuration, GREAT-based or HGS-based initialization performs better. 
On the San Francisco and Vienna data, GREAT-based initialization is best on all configurations apart from the ones with demand range $\{1,\ldots,9\}$. On Gothenburg data, HGS-based initialization is additionally better for demand $\{11,\ldots,19\}$-range if the battery capacity is at least 30kWh. We want to emphasize the strong performance of ALNS with only 1s of runtime when using GREAT-based warm initialization while still having an extremely fast runtime. 
}

The results confirm that the holistic, learning-based approach of GREAT to solve EVRP provides substantial practical advantages in many demanding and realistic operating conditions for EV-based VRPs, while also generalizing across geographically diverse cities.

\begin{table*}
  \centering
  \color{black}
  \caption{EVRP results for San Francisco (training city). \realrev{Gaps are reported relative to the HGS baseline; negative values indicate better performance than HGS. Results over 100 test instances.}}
  \label{tab:SF}
  \footnotesize
  \setlength{\tabcolsep}{3pt}
  \begin{tabular}{llrrrrrrrrr}
  \toprule
  \multicolumn{2}{c}{} & \multicolumn{3}{c}{Customer Demand [1, 9]} & \multicolumn{3}{c}{Customer Demand [11, 19]} & \multicolumn{3}{c}{Customer Demand [21, 29]} \\
  \cmidrule(lr){3-5} \cmidrule(lr){6-8} \cmidrule(lr){9-11}
  Battery (kWh) & Method & kWh & opt. gap & time & kWh & opt. gap & time & kWh & opt. gap & time \\
  \midrule
  \multirow{13}{*}{15} & Baseline (HGS) & 26.66 & -- & 8.6\,min & 65.32 & -- & 10.4\,min & 110.53 & -- & 6.2\,min \\
   & \realrev{ ALNS (1s)} & 31.92 & 20.10 & 11.7\,s & 72.79 & 11.61 & 7.8\,s & 112.43 & 1.82 & 8.2\,s \\
   & ALNS (5s) & 27.57 & 3.61 & 35.6\,s & 69.33 & 6.16 & 35.5\,s & 106.84 & -3.29 & 35.6\,s \\
   & ALNS (60s) & 26.53 & -0.39 & 7.0\,min & 63.65 & -2.57 & 7.0\,min & 102.66 & -7.06 & 7.0\,min \\
   & ALNS HGS-init (1s) & \underline{25.53} & \underline{-4.13} & 8.8\,min & 63.78 & -2.35 & 10.6\,min & 107.54 & -2.70 & 6.3\,min \\
   & ALNS HGS-init (5s) & 25.65 & -3.67 & 9.2\,min & 64.12 & -1.81 & 11.0\,min & 106.27 & -3.82 & 6.8\,min \\
   & ALNS HGS-init (60s) & \textbf{25.45} & \textbf{-4.45} & 15.7\,min & 63.01 & -3.53 & 17.4\,min & 102.63 & -7.11 & 13.2\,min \\
   & ALNS GREAT-init (1s) & 26.13 & -1.83 & 13.4\,s & \underline{61.47} & \underline{-5.81} & 13.7\,s & \underline{100.66} & \underline{-8.85} & 15.0\,s \\
   & ALNS GREAT-init (5s) & 26.12 & -1.86 & 41.4\,s & 61.48 & -5.80 & 41.7\,s & 100.66 & -8.84 & 43.0\,s \\
   & ALNS GREAT-init (60s) & 25.72 & -3.38 & 7.1\,min & \textbf{61.45} & \textbf{-5.85} & 7.1\,min & \textbf{100.64} & \textbf{-8.86} & 7.1\,min \\
   & GREAT NB x9 + 2opt & 26.64 & 0.11 & 5.8\,s & 61.58 & -5.64 & 6.2\,s & 100.73 & -8.78 & 7.6\,s \\
   & GREAT NB x9 & 27.17 & 2.13 & 4.6\,s & 62.40 & -4.35 & 5.4\,s & 101.52 & -8.07 & 6.3\,s \\
   & GREAT NB x1 & 27.50 & 3.37 & 0.8\,s & 62.79 & -3.73 & 1.0\,s & 101.80 & -7.81 & 1.2\,s \\
  \midrule
  \multirow{13}{*}{20} & Baseline (HGS) & 26.04 & -- & 9.7\,min & 64.46 & -- & 10.8\,min & 108.87 & -- & 6.5\,min \\
   & ALNS (1s) & 32.08 & 23.53 & 12.1\,s & 72.55 & 12.75 & 7.9\,s & 111.63 & 2.71 & 8.3\,s \\
   & ALNS (5s) & 28.11 & 8.08 & 36.0\,s & 69.21 & 7.41 & 35.6\,s & 106.82 & -1.81 & 35.6\,s \\
   & ALNS (60s) & 26.83 & 3.17 & 7.0\,min & 63.71 & -1.14 & 7.0\,min & 102.55 & -5.74 & 7.0\,min \\
   & ALNS HGS-init (1s) & \textbf{25.33} & \textbf{-2.63} & 9.8\,min & 63.15 & -1.96 & 10.9\,min & 107.01 & -1.65 & 6.6\,min \\
   & ALNS HGS-init (5s) & 25.45 & -2.21 & 10.3\,min & 63.52 & -1.41 & 11.4\,min & 106.05 & -2.52 & 7.1\,min \\
   & ALNS HGS-init (60s) & \underline{25.34} & \underline{-2.62} & 16.7\,min & 62.84 & -2.44 & 17.8\,min & 102.48 & -5.82 & 13.5\,min \\
   & ALNS GREAT-init (1s) & 25.92 & -0.33 & 13.4\,s & \underline{61.30} & \underline{-4.74} & 13.7\,s & \underline{100.48} & \underline{-7.60} & 15.0\,s \\
   & ALNS GREAT-init (5s) & 25.92 & -0.36 & 41.3\,s & 61.30 & -4.74 & 41.7\,s & 100.48 & -7.60 & 43.0\,s \\
   & ALNS GREAT-init (60s) & 25.62 & -1.52 & 7.1\,min & \textbf{61.28} & \textbf{-4.77} & 7.1\,min & \textbf{100.43} & \textbf{-7.65} & 7.1\,min \\
   & GREAT NB x9 + 2opt & 26.34 & 1.31 & 5.7\,s & 61.38 & -4.61 & 6.1\,s & 100.53 & -7.56 & 7.5\,s \\
   & GREAT NB x9 & 26.85 & 3.28 & 4.5\,s & 62.06 & -3.53 & 5.3\,s & 101.25 & -6.91 & 6.3\,s \\
   & GREAT NB x1 & 27.13 & 4.35 & 0.8\,s & 62.48 & -2.85 & 1.0\,s & 101.52 & -6.65 & 1.2\,s \\
  \midrule
  \multirow{13}{*}{30} & Baseline (HGS) & 25.66 & -- & 9.8\,min & 63.38 & -- & 10.8\,min & 107.51 & -- & 6.6\,min \\
   & ALNS (1s) & 31.57 & 23.22 & 12.5\,s & 72.40 & 14.29 & 8.0\,s & 112.80 & 5.12 & 8.4\,s \\
   & ALNS (5s) & 28.01 & 9.20 & 35.9\,s & 69.02 & 8.89 & 35.7\,s & 107.00 & -0.45 & 35.6\,s \\
   & ALNS (60s) & 26.42 & 2.89 & 7.0\,min & 64.04 & 0.96 & 7.0\,min & 102.74 & -4.40 & 7.0\,min \\
   & ALNS HGS-init (1s) & \textbf{25.14} & \textbf{-2.03} & 10.0\,min & 62.61 & -1.25 & 10.9\,min & 106.51 & -0.90 & 6.7\,min \\
   & ALNS HGS-init (5s) & 25.20 & -1.80 & 10.4\,min & 62.80 & -0.96 & 11.4\,min & 105.94 & -1.43 & 7.2\,min \\
   & ALNS HGS-init (60s) & \underline{25.18} & \underline{-1.88} & 16.9\,min & 62.61 & -1.24 & 17.8\,min & 102.56 & -4.57 & 13.6\,min \\
   & ALNS GREAT-init (1s) & 25.83 & 0.62 & 13.5\,s & \textbf{61.14} & \textbf{-3.48} & 13.6\,s & \underline{100.38} & \underline{-6.54} & 14.9\,s \\
   & ALNS GREAT-init (5s) & 25.85 & 0.68 & 41.4\,s & 61.15 & -3.47 & 41.6\,s & 100.38 & -6.54 & 42.9\,s \\
   & ALNS GREAT-init (60s) & 25.57 & -0.41 & 7.1\,min & \underline{61.14} & \underline{-3.48} & 7.1\,min & \textbf{100.37} & \textbf{-6.55} & 7.1\,min \\
   & GREAT NB x9 + 2opt & 26.17 & 1.96 & 5.7\,s & 61.23 & -3.33 & 6.0\,s & 100.42 & -6.51 & 7.4\,s \\
   & GREAT NB x9 & 26.58 & 3.58 & 4.5\,s & 61.91 & -2.25 & 5.3\,s & 101.13 & -5.85 & 6.2\,s \\
   & GREAT NB x1 & 26.89 & 4.79 & 0.8\,s & 62.28 & -1.64 & 1.0\,s & 101.36 & -5.64 & 1.2\,s \\
  \midrule
  \multirow{13}{*}{40} & Baseline (HGS) & 25.29 & -- & 10.0\,min & 62.99 & -- & 10.9\,min & 106.85 & -- & 6.6\,min \\
   & ALNS (1s) & 31.75 & 25.91 & 14.2\,s & 72.21 & 14.73 & 8.2\,s & 113.35 & 6.26 & 8.6\,s \\
   & ALNS (5s) & 28.45 & 12.73 & 36.5\,s & 69.37 & 10.11 & 35.8\,s & 107.38 & 0.52 & 35.8\,s \\
   & ALNS (60s) & 26.41 & 4.49 & 7.0\,min & 64.07 & 1.64 & 7.0\,min & 102.94 & -3.62 & 7.0\,min \\
   & ALNS HGS-init (1s) & \textbf{25.00} & \textbf{-1.02} & 10.1\,min & 62.41 & -0.94 & 11.0\,min & 106.19 & -0.60 & 6.7\,min \\
   & ALNS HGS-init (5s) & \underline{25.04} & \underline{-0.91} & 10.6\,min & 62.50 & -0.78 & 11.5\,min & 105.89 & -0.87 & 7.2\,min \\
   & ALNS HGS-init (60s) & 25.04 & -0.91 & 17.0\,min & 62.41 & -0.93 & 17.9\,min & 102.91 & -3.65 & 13.6\,min \\
   & ALNS GREAT-init (1s) & 25.85 & 2.37 & 13.8\,s & \underline{61.17} & \underline{-2.79} & 13.6\,s & \underline{100.36} & \underline{-5.99} & 14.8\,s \\
   & ALNS GREAT-init (5s) & 25.83 & 2.26 & 41.6\,s & 61.18 & -2.78 & 41.7\,s & 100.36 & -5.99 & 42.9\,s \\
   & ALNS GREAT-init (60s) & 25.55 & 1.12 & 7.1\,min & \textbf{61.16} & \textbf{-2.82} & 7.1\,min & \textbf{100.35} & \textbf{-6.00} & 7.1\,min \\
   & GREAT NB x9 + 2opt & 26.17 & 3.63 & 5.7\,s & 61.24 & -2.68 & 6.0\,s & 100.39 & -5.97 & 7.3\,s \\
   & GREAT NB x9 & 26.56 & 5.21 & 4.5\,s & 61.92 & -1.59 & 5.2\,s & 101.07 & -5.33 & 6.1\,s \\
   & GREAT NB x1 & 26.85 & 6.36 & 0.8\,s & 62.27 & -1.04 & 1.0\,s & 101.28 & -5.13 & 1.2\,s \\
  \bottomrule
  \end{tabular}
\end{table*}

\begin{table*}
  \centering
  \color{black}
  \caption{EVRP results for Gothenburg (unseen city, no retraining). \realrev{Gaps are reported relative to the HGS baseline; negative values indicate better performance than HGS. Results over 100 test instances.}}
  \label{tab:Gothenburg}
  \footnotesize
  \setlength{\tabcolsep}{3pt}
 \begin{tabular}{llrrrrrrrrr}
  \toprule
  \multicolumn{2}{c}{} & \multicolumn{3}{c}{Customer Demand [1, 9]} & \multicolumn{3}{c}{Customer Demand [11, 19]} & \multicolumn{3}{c}{Customer Demand [21, 29]} \\
  \cmidrule(lr){3-5} \cmidrule(lr){6-8} \cmidrule(lr){9-11}
  Battery (kWh) & Method & kWh & opt. gap & time & kWh & opt. gap & time & kWh & opt. gap & time \\
  \midrule
  \multirow{13}{*}{15} & Baseline (HGS) & 46.17 & -- & 7.0\,min & 90.25 & -- & 7.1\,min & 142.45 & -- & 5.7\,min \\
   & ALNS (1s) & 54.41 & 18.77 & 15.1\,s & 99.75 & 10.87 & 8.3\,s & 151.71 & 6.75 & 8.7\,s \\
   & ALNS (5s) & 47.79 & 3.99 & 36.7\,s & 95.73 & 6.27 & 35.9\,s & 142.19 & -0.08 & 35.7\,s \\
   & ALNS (60s) & 44.38 & -3.68 & 7.0\,min & 88.09 & -2.25 & 7.0\,min & 137.44 & -3.39 & 7.0\,min \\
   & ALNS HGS-init (1s) & 43.90 & -4.72 & 7.1\,min & 87.95 & -2.48 & 7.3\,min & 139.88 & -1.71 & 5.9\,min \\
   & ALNS HGS-init (5s) & 43.65 & -5.19 & 7.6\,min & 88.38 & -2.01 & 7.7\,min & 140.05 & -1.56 & 6.3\,min \\
   & ALNS HGS-init (60s) & \textbf{43.00} & \textbf{-6.60} & 14.0\,min & 87.43 & -3.04 & 14.1\,min & 137.50 & -3.35 & 12.7\,min \\
   & ALNS GREAT-init (1s) & 43.88 & -4.67 & 14.9\,s & \underline{87.31} & \underline{-3.11} & 13.8\,s & \underline{136.56} & \underline{-3.97} & 14.8\,s \\
   & ALNS GREAT-init (5s) & 43.67 & -5.14 & 41.8\,s & 87.32 & -3.10 & 41.8\,s & 136.58 & -3.96 & 42.8\,s \\
   & ALNS GREAT-init (60s) & \underline{43.02} & \underline{-6.54} & 7.1\,min & \textbf{87.02} & \textbf{-3.43} & 7.1\,min & \textbf{136.46} & \textbf{-4.05} & 7.1\,min \\
   & GREAT NB x9 + 2opt & 44.58 & -3.15 & 5.9\,s & 87.52 & -2.89 & 6.1\,s & 136.75 & -3.84 & 7.2\,s \\
   & GREAT NB x9 & 45.39 & -1.40 & 4.6\,s & 88.39 & -1.91 & 5.3\,s & 137.38 & -3.41 & 6.0\,s \\
   & GREAT NB x1 & 45.98 & -0.12 & 0.8\,s & 88.95 & -1.31 & 1.0\,s & 137.86 & -3.09 & 1.2\,s \\
  \midrule
  \multirow{13}{*}{20} & Baseline (HGS) & 44.55 & -- & 6.9\,min & 88.66 & -- & 7.1\,min & 140.28 & -- & 5.7\,min \\
   & ALNS (1s) & 52.80 & 18.93 & 14.9\,s & 98.20 & 11.00 & 8.5\,s & 150.56 & 7.64 & 9.0\,s \\
   & ALNS (5s) & 47.24 & 6.19 & 36.9\,s & 94.35 & 6.54 & 35.8\,s & 140.75 & 0.43 & 35.8\,s \\
   & ALNS (60s) & 43.55 & -2.15 & 7.0\,min & 87.36 & -1.42 & 7.0\,min & 136.63 & -2.46 & 7.0\,min \\
   & ALNS HGS-init (1s) & 42.94 & -3.57 & 7.1\,min & 86.88 & -1.99 & 7.2\,min & 138.57 & -1.14 & 5.9\,min \\
   & ALNS HGS-init (5s) & 42.98 & -3.46 & 7.5\,min & 87.15 & -1.67 & 7.7\,min & 138.54 & -1.16 & 6.3\,min \\
   & ALNS HGS-init (60s) & \underline{42.59} & \underline{-4.32} & 13.9\,min & 86.66 & -2.23 & 14.1\,min & 136.42 & -2.62 & 12.7\,min \\
   & ALNS GREAT-init (1s) & 43.30 & -2.68 & 14.7\,s & \underline{86.42} & \underline{-2.44} & 13.7\,s & \underline{135.50} & \underline{-3.22} & 14.5\,s \\
   & ALNS GREAT-init (5s) & 43.08 & -3.19 & 41.8\,s & 86.44 & -2.41 & 41.6\,s & 135.51 & -3.22 & 42.5\,s \\
   & ALNS GREAT-init (60s) & \textbf{42.58} & \textbf{-4.31} & 7.1\,min & \textbf{86.33} & \textbf{-2.54} & 7.1\,min & \textbf{135.41} & \textbf{-3.29} & 7.1\,min \\
   & GREAT NB x9 + 2opt & 43.87 & -1.39 & 5.8\,s & 86.60 & -2.24 & 6.0\,s & 135.61 & -3.15 & 6.9\,s \\
   & GREAT NB x9 & 44.54 & 0.12 & 4.5\,s & 87.29 & -1.43 & 5.2\,s & 136.04 & -2.84 & 5.8\,s \\
   & GREAT NB x1 & 45.00 & 1.16 & 0.8\,s & 87.78 & -0.87 & 1.0\,s & 136.40 & -2.59 & 1.1\,s \\
  \midrule
  \multirow{13}{*}{30} & Baseline (HGS) & 43.49 & -- & 7.0\,min & 86.70 & -- & 7.1\,min & 138.03 & -- & 5.7\,min \\
   & ALNS (1s) & 52.15 & 20.28 & 12.6\,s & 96.78 & 11.89 & 8.2\,s & 146.61 & 6.38 & 8.6\,s \\
   & ALNS (5s) & 46.28 & 6.47 & 36.4\,s & 93.37 & 7.82 & 35.9\,s & 139.12 & 0.79 & 35.6\,s \\
   & ALNS (60s) & 43.32 & -0.38 & 7.0\,min & 87.05 & 0.43 & 7.0\,min & 135.89 & -1.52 & 7.0\,min \\
   & ALNS HGS-init (1s) & \underline{42.29} & \underline{-2.74} & 7.1\,min & \underline{85.80} & \underline{-1.01} & 7.2\,min & 137.07 & -0.67 & 5.9\,min \\
   & ALNS HGS-init (5s) & 42.35 & -2.58 & 7.5\,min & 85.95 & -0.83 & 7.7\,min & 137.01 & -0.73 & 6.3\,min \\
   & ALNS HGS-init (60s) & \textbf{42.16} & \textbf{-3.03} & 14.0\,min & \textbf{85.76} & \textbf{-1.03} & 14.1\,min & 135.56 & -1.75 & 12.7\,min \\
   & ALNS GREAT-init (1s) & 42.93 & -1.24 & 14.2\,s & 85.94 & -0.76 & 13.3\,s & \underline{134.78} & \underline{-2.27} & 14.1\,s \\
   & ALNS GREAT-init (5s) & 42.78 & -1.59 & 41.6\,s & 85.96 & -0.74 & 41.3\,s & \underline{134.78} & \underline{-2.27} & 42.1\,s \\
   & ALNS GREAT-init (60s) & 42.35 & -2.57 & 7.1\,min & 85.89 & -0.83 & 7.1\,min & \textbf{134.75} & \textbf{-2.30} & 7.1\,min \\
   & GREAT NB x9 + 2opt & 43.41 & -0.10 & 5.7\,s & 86.08 & -0.60 & 5.8\,s & 134.86 & -2.21 & 6.7\,s \\
   & GREAT NB x9 & 43.99 & 1.21 & 4.5\,s & 86.70 & 0.14 & 5.1\,s & 135.17 & -1.98 & 5.7\,s \\
   & GREAT NB x1 & 44.42 & 2.23 & 0.8\,s & 87.14 & 0.67 & 0.9\,s & 135.47 & -1.76 & 1.1\,s \\
  \midrule
  \multirow{13}{*}{40} & Baseline (HGS) & 42.55 & -- & 6.9\,min & 86.18 & -- & 7.1\,min & 136.79 & -- & 5.7\,min \\
   & ALNS (1s) & 50.87 & 20.10 & 12.2\,s & 95.42 & 10.91 & 8.2\,s & 145.34 & 6.45 & 8.7\,s \\
   & ALNS (5s) & 45.10 & 6.14 & 36.2\,s & 92.13 & 7.00 & 35.7\,s & 138.52 & 1.28 & 35.7\,s \\
   & ALNS (60s) & 43.12 & 1.37 & 7.0\,min & 86.78 & 0.72 & 7.0\,min & 135.52 & -0.90 & 7.0\,min \\
   & ALNS HGS-init (1s) & \underline{41.93} & \underline{-1.37} & 7.1\,min & \textbf{85.33} & \textbf{-0.94} & 7.3\,min & 136.16 & -0.43 & 5.8\,min \\
   & ALNS HGS-init (5s) & 42.01 & -1.21 & 7.5\,min & 85.47 & -0.78 & 7.7\,min & 136.13 & -0.45 & 6.3\,min \\
   & ALNS HGS-init (60s) & \textbf{41.88} & \textbf{-1.47} & 14.0\,min & \underline{85.34} & \underline{-0.93} & 14.1\,min & 135.26 & -1.09 & 12.7\,min \\
   & ALNS GREAT-init (1s) & 42.73 & 0.55 & 13.7\,s & 85.74 & -0.40 & 13.2\,s & \underline{134.56} & \underline{-1.54} & 14.0\,s \\
   & ALNS GREAT-init (5s) & 42.67 & 0.40 & 41.5\,s & 85.76 & -0.39 & 41.1\,s & \underline{134.56} & \underline{-1.54} & 42.0\,s \\
   & ALNS GREAT-init (60s) & 42.19 & -0.72 & 7.1\,min & 85.69 & -0.48 & 7.1\,min & \textbf{134.49} & \textbf{-1.60} & 7.1\,min \\
   & GREAT NB x9 + 2opt & 43.22 & 1.72 & 5.7\,s & 85.86 & -0.27 & 5.6\,s & 134.63 & -1.49 & 6.5\,s \\
   & GREAT NB x9 & 43.78 & 3.06 & 4.4\,s & 86.44 & 0.43 & 5.0\,s & 134.90 & -1.29 & 5.6\,s \\
   & GREAT NB x1 & 44.19 & 4.04 & 0.8\,s & 86.92 & 1.00 & 0.9\,s & 135.20 & -1.07 & 1.1\,s \\
  \bottomrule
  \end{tabular}
\end{table*}

\begin{table*}
  \centering
  \color{black}
  \caption{EVRP results for Vienna (unseen city, no retraining). \realrev{Gaps are reported relative to the HGS baseline; negative values indicate better performance than HGS. Results over 100 test instances.}}
  \label{tab:Vienna}
  \footnotesize
  \setlength{\tabcolsep}{3pt}
  \begin{tabular}{llrrrrrrrrr}
  \toprule
  \multicolumn{2}{c}{} & \multicolumn{3}{c}{Customer Demand [1, 9]} & \multicolumn{3}{c}{Customer Demand [11, 19]} & \multicolumn{3}{c}{Customer Demand [21, 29]} \\
  \cmidrule(lr){3-5} \cmidrule(lr){6-8} \cmidrule(lr){9-11}
  Battery (kWh) & Method & kWh & opt. gap & time & kWh & opt. gap & time & kWh & opt. gap & time \\
  \midrule
  \multirow{13}{*}{15} & Baseline (HGS) & 37.73 & -- & 8.0\,min & 79.72 & -- & 7.5\,min & 127.24 & -- & 5.8\,min \\
   & ALNS (1s) & 44.84 & 19.33 & 14.1\,s & 87.40 & 10.19 & 8.4\,s & 134.71 & 6.16 & 8.9\,s \\
   & ALNS (5s) & 39.84 & 6.00 & 36.4\,s & 84.15 & 5.74 & 35.7\,s & 126.24 & -0.64 & 36.1\,s \\
   & ALNS (60s) & 37.13 & -1.58 & 7.0\,min & 77.59 & -2.60 & 7.0\,min & 121.91 & -4.04 & 7.0\,min \\
   & ALNS HGS-init (1s) & 36.23 & -3.86 & 8.1\,min & 77.78 & -2.39 & 7.6\,min & 124.77 & -1.87 & 5.9\,min \\
   & ALNS HGS-init (5s) & 36.37 & -3.49 & 8.6\,min & 78.07 & -2.01 & 8.1\,min & 124.68 & -1.90 & 6.4\,min \\
   & ALNS HGS-init (60s) & \textbf{35.96} & \textbf{-4.56} & 15.0\,min & 77.24 & -3.07 & 14.5\,min & 121.88 & -4.07 & 12.8\,min \\
   & ALNS GREAT-init (1s) & 36.80 & -2.29 & 14.4\,s & \underline{76.66} & \underline{-3.64} & 13.5\,s & \underline{120.60} & \underline{-4.99} & 14.5\,s \\
   & ALNS GREAT-init (5s) & 36.64 & -2.74 & 41.7\,s & 76.68 & -3.62 & 41.6\,s & 120.61 & -4.98 & 42.5\,s \\
   & ALNS GREAT-init (60s) & \underline{36.16} & \underline{-4.02} & 7.1\,min & \textbf{76.56} & \textbf{-3.78} & 7.1\,min & \textbf{120.51} & \textbf{-5.06} & 7.1\,min \\
   & GREAT NB x9 + 2opt & 37.42 & -0.61 & 5.7\,s & 76.83 & -3.42 & 6.0\,s & 120.70 & -4.91 & 7.0\,s \\
   & GREAT NB x9 & 38.06 & 1.07 & 4.6\,s & 77.67 & -2.36 & 5.2\,s & 121.26 & -4.47 & 5.8\,s \\
   & GREAT NB x1 & 38.53 & 2.33 & 0.8\,s & 78.09 & -1.81 & 1.0\,s & 121.61 & -4.20 & 1.1\,s \\
  \midrule
  \multirow{13}{*}{20} & Baseline (HGS) & 37.01 & -- & 7.9\,min & 78.37 & -- & 7.7\,min & 125.39 & -- & 5.8\,min \\
   & ALNS (1s) & 43.91 & 19.17 & 14.0\,s & 86.77 & 11.20 & 8.3\,s & 133.64 & 6.79 & 8.7\,s \\
   & ALNS (5s) & 39.64 & 7.40 & 36.4\,s & 83.80 & 7.00 & 35.7\,s & 125.14 & -0.03 & 35.7\,s \\
   & ALNS (60s) & 37.23 & 0.49 & 7.0\,min & 77.32 & -1.33 & 7.0\,min & 121.18 & -3.15 & 7.0\,min \\
   & ALNS HGS-init (1s) & 35.88 & -2.97 & 8.1\,min & 76.92 & -1.81 & 7.8\,min & 123.60 & -1.34 & 5.9\,min \\
   & ALNS HGS-init (5s) & 35.94 & -2.75 & 8.5\,min & 77.24 & -1.43 & 8.3\,min & 123.59 & -1.33 & 6.4\,min \\
   & ALNS HGS-init (60s) & \textbf{35.70} & \textbf{-3.42} & 14.9\,min & 76.75 & -2.03 & 14.7\,min & 121.26 & -3.10 & 12.8\,min \\
   & ALNS GREAT-init (1s) & 36.50 & -1.19 & 14.2\,s & \underline{76.18} & \underline{-2.60} & 13.5\,s & \underline{119.99} & \underline{-4.03} & 14.2\,s \\
   & ALNS GREAT-init (5s) & 36.35 & -1.63 & 41.6\,s & 76.20 & -2.59 & 41.4\,s & 120.00 & -4.02 & 42.2\,s \\
   & ALNS GREAT-init (60s) & \underline{35.86} & \underline{-2.96} & 7.1\,min & \textbf{76.12} & \textbf{-2.70} & 7.1\,min & \textbf{119.96} & \textbf{-4.05} & 7.1\,min \\
   & GREAT NB x9 + 2opt & 37.01 & 0.23 & 5.7\,s & 76.33 & -2.42 & 5.8\,s & 120.07 & -3.97 & 6.7\,s \\
   & GREAT NB x9 & 37.52 & 1.60 & 4.5\,s & 77.00 & -1.54 & 5.1\,s & 120.45 & -3.66 & 5.6\,s \\
   & GREAT NB x1 & 37.89 & 2.66 & 0.8\,s & 77.40 & -1.03 & 0.9\,s & 120.75 & -3.41 & 1.1\,s \\
  \midrule
  \multirow{13}{*}{30} & Baseline (HGS) & 36.14 & -- & 7.9\,min & 77.01 & -- & 7.6\,min & 123.90 & -- & 5.8\,min \\
   & ALNS (1s) & 43.14 & 19.93 & 13.4\,s & 85.53 & 11.39 & 8.1\,s & 131.64 & 6.55 & 8.6\,s \\
   & ALNS (5s) & 38.45 & 6.71 & 36.1\,s & 83.12 & 7.96 & 35.7\,s & 124.35 & 0.48 & 35.6\,s \\
   & ALNS (60s) & 36.77 & 1.84 & 7.0\,min & 77.12 & 0.10 & 7.0\,min & 120.77 & -2.34 & 7.0\,min \\
   & ALNS HGS-init (1s) & \underline{35.41} & \underline{-1.88} & 8.1\,min & 76.11 & -1.16 & 7.7\,min & 122.76 & -0.88 & 5.9\,min \\
   & ALNS HGS-init (5s) & 35.46 & -1.77 & 8.5\,min & 76.26 & -0.96 & 8.2\,min & 122.71 & -0.91 & 6.4\,min \\
   & ALNS HGS-init (60s) & \textbf{35.37} & \textbf{-1.99} & 14.9\,min & 76.07 & -1.22 & 14.6\,min & 120.68 & -2.41 & 12.8\,min \\
   & ALNS GREAT-init (1s) & 36.22 & 0.40 & 13.6\,s & \underline{75.88} & \underline{-1.32} & 13.4\,s & \underline{119.54} & \underline{-3.26} & 14.0\,s \\
   & ALNS GREAT-init (5s) & 36.13 & 0.14 & 41.4\,s & 75.88 & -1.31 & 41.4\,s & 119.54 & -3.26 & 42.0\,s \\
   & ALNS GREAT-init (60s) & 35.68 & -1.12 & 7.1\,min & \textbf{75.83} & \textbf{-1.39} & 7.1\,min & \textbf{119.51} & \textbf{-3.28} & 7.1\,min \\
   & GREAT NB x9 + 2opt & 36.74 & 1.90 & 5.6\,s & 76.01 & -1.14 & 5.7\,s & 119.60 & -3.21 & 6.5\,s \\
   & GREAT NB x9 & 37.17 & 3.09 & 4.4\,s & 76.58 & -0.38 & 5.0\,s & 119.89 & -2.97 & 5.5\,s \\
   & GREAT NB x1 & 37.51 & 4.08 & 0.8\,s & 76.98 & 0.16 & 0.9\,s & 120.17 & -2.74 & 1.1\,s \\
  \midrule
  \multirow{13}{*}{40} & Baseline (HGS) & 35.73 & -- & 7.9\,min & 76.38 & -- & 7.6\,min & 122.99 & -- & 5.8\,min \\
   & ALNS (1s) & 42.82 & 20.16 & 12.0\,s & 85.34 & 11.99 & 8.2\,s & 131.88 & 7.36 & 8.8\,s \\
   & ALNS (5s) & 38.27 & 7.20 & 36.1\,s & 82.70 & 8.33 & 35.7\,s & 124.07 & 0.98 & 35.8\,s \\
   & ALNS (60s) & 36.65 & 2.35 & 7.0\,min & 77.15 & 0.91 & 7.0\,min & 120.69 & -1.69 & 7.0\,min \\
   & ALNS HGS-init (1s) & \textbf{35.21} & \textbf{-1.41} & 8.0\,min & 75.89 & -0.63 & 7.7\,min & 122.36 & -0.47 & 5.9\,min \\
   & ALNS HGS-init (5s) & 35.28 & -1.22 & 8.5\,min & 75.98 & -0.51 & 8.2\,min & 122.21 & -0.58 & 6.4\,min \\
   & ALNS HGS-init (60s) & \underline{35.25} & \underline{-1.29} & 14.9\,min & 75.86 & -0.68 & 14.6\,min & 120.57 & -1.80 & 12.8\,min \\
   & ALNS GREAT-init (1s) & 36.12 & 1.12 & 13.6\,s & \underline{75.80} & \underline{-0.60} & 13.4\,s & \underline{119.41} & \underline{-2.65} & 13.9\,s \\
   & ALNS GREAT-init (5s) & 36.05 & 0.93 & 41.5\,s & 75.80 & -0.59 & 41.3\,s & 119.41 & -2.64 & 42.0\,s \\
   & ALNS GREAT-init (60s) & 35.63 & -0.24 & 7.1\,min & \textbf{75.74} & \textbf{-0.70} & 7.1\,min & \textbf{119.37} & \textbf{-2.68} & 7.1\,min \\
   & GREAT NB x9 + 2opt & 36.58 & 2.45 & 5.6\,s & 75.91 & -0.44 & 5.7\,s & 119.47 & -2.59 & 6.4\,s \\
   & GREAT NB x9 & 37.06 & 3.85 & 4.4\,s & 76.50 & 0.37 & 5.0\,s & 119.74 & -2.37 & 5.5\,s \\
   & GREAT NB x1 & 37.38 & 4.74 & 0.8\,s & 76.85 & 0.82 & 0.9\,s & 120.00 & -2.15 & 1.1\,s \\
  \bottomrule
  \end{tabular}
\end{table*}

\FloatBarrier

\subsection{\realrev{Robustness to Charging Station Locations}}
\label{sec:chargers}

\realrev{When generating EVRP data in Section~\ref{sec:data_gen}, we assumed that both customers as well as chargers are placed according to population density. Although it would generally be desirable when chargers were placed close to people for increased convenience, this might not reflect reality.
Therefore, in this section, we investigate different charger locations.
In particular, we now choose chargers based on real charger location in the San Francisco OpenStreetMap data. We map those chargers to the nearest locations in our road network graph.
As a further ablation, we also vary the number of chargers in our instances (while GREAT has been trained on instances with 5 chargers only, it obtains the information whether an edge leads to a charger or not as a one-hot encoding and we can test GREAT's zero-shot generalization to settings with different numbers of chargers). We consider instances with 3, 5, and 10 random chargers from OpenStreetMap. Further, we also include the setting with 5 population density based chargers from our main result section again for comparison. For all four settings, we consider the set of the same 100 customers and the same depot, only the charger locations and numbers change, making the results between the different settings comparable. 
We conduct this ablation on the four extreme cases with very low and very high customer demands (demands $\{1, \dots, 9\} \text{ and } \{21, \dots, 29\}$) and very small and very large battery capacities (15kWh and 40kWh).
}

\realrev{The results can be observed in Table~\ref{tab:charger-sensitivity}. We can see that the best performance for a battery size - demand configuration is generally achieved by the same method independent of the charger placement and consistent with the results of Section~\ref{sec:evrp_results}. The only exception is the setting with 40kWh battery and low demands (the setting where charging is the least important) where sometimes shorter or longer ALNS runtime with an HGS-based initialization leads to the best results. We note however, that the energy consumptions in this setting are very close to one another.
We reiterate that this experiment also successfully shows that GREAT can generalize to unseen charger distributions and charger numbers. Further, we highlight GREAT's capability of solving many batched EVRP instances in parallel, allowing to test different charger locations which can potentially be useful to evaluate candidate charging-infrastructure layouts.}

\begin{table*}
\centering
\color{black}
\footnotesize
  \setlength{\tabcolsep}{3pt}
\caption{\realrev{Robustness of the EVRP results to the location and the number of charging stations, on San Francisco data. OSM$k$ chooses $k$ charging stations drawn from OpenStreetMap data. POP5 is the population-density sampling used in the main experiments. Average energy consumption in kWh over 100 instances.}}
\label{tab:charger-sensitivity}
\begin{tabular}{lrrrr}
\toprule
Method & OSM3 & OSM5 & OSM10 & POP5 \\
\midrule
\multicolumn{5}{l}{\emph{Battery 15\,kWh, demand $\{1,\ldots,9\}$}} \\
  Baseline (HGS) & 28.43 & 27.61 & 27.19 & 27.47 \\
  ALNS (1\,s) & 33.67 & 33.26 & 32.63 & 32.91 \\
  ALNS (5\,s) & 29.40 & 29.17 & 28.73 & 29.02 \\
  ALNS (60\,s) & 26.58 & 26.47 & 26.37 & 26.43 \\
  ALNS HGS-init (1\,s) & 26.71 & 26.32 & 26.17 & 26.25 \\
  ALNS HGS-init (5\,s) & 26.83 & 26.46 & 26.35 & 26.45 \\
  ALNS HGS-init (60\,s) & \textbf{26.36} & \textbf{26.19} & \textbf{26.11} & \textbf{26.18} \\
  ALNS GREAT-init (1\,s) & 27.38 & 26.90 & 26.81 & 26.86 \\
  ALNS GREAT-init (5\,s) & 27.30 & 26.89 & 26.78 & 26.82 \\
  ALNS GREAT-init (60\,s) & 26.60 & 26.33 & 26.35 & 26.40 \\
  GREAT NB $\times 9$ & 28.79 & 27.85 & 27.74 & 27.93 \\
  GREAT NB $\times 9$ + 2opt & 27.95 & 27.32 & 27.25 & 27.33 \\
\midrule
\multicolumn{5}{l}{\emph{Battery 15\,kWh, demand $\{21,\ldots,29\}$}} \\
  Baseline (HGS) & 111.93 & 110.23 & 109.31 & 109.55 \\
  ALNS (1\,s) & 113.99 & 112.64 & 109.42 & 112.61 \\
  ALNS (5\,s) & 108.62 & 106.98 & 104.83 & 106.86 \\
  ALNS (60\,s) & 104.53 & 102.66 & 100.57 & 102.36 \\
  ALNS HGS-init (1\,s) & 108.33 & 106.79 & 105.85 & 106.59 \\
  ALNS HGS-init (5\,s) & 107.84 & 106.27 & 104.84 & 105.93 \\
  ALNS HGS-init (60\,s) & 104.32 & 102.57 & 100.52 & 102.28 \\
  ALNS GREAT-init (1\,s) & 103.27 & 100.54 & 98.64 & 100.07 \\
  ALNS GREAT-init (5\,s) & 103.31 & 100.54 & 98.65 & 100.07 \\
  ALNS GREAT-init (60\,s) & \textbf{103.13} & \textbf{100.52} & \textbf{98.62} & \textbf{100.03} \\
  GREAT NB $\times 9$ & 105.08 & 100.97 & 98.96 & 100.94 \\
  GREAT NB $\times 9$ + 2opt & 103.46 & 100.61 & 98.71 & 100.14 \\
\midrule
\multicolumn{5}{l}{\emph{Battery 40\,kWh, demand $\{1,\ldots,9\}$}} \\
  Baseline (HGS) & 25.62 & 25.50 & 25.43 & 25.39 \\
  ALNS (1\,s) & 31.58 & 31.65 & 31.30 & 31.50 \\
  ALNS (5\,s) & 28.17 & 28.21 & 27.83 & 27.88 \\
  ALNS (60\,s) & 25.79 & 25.80 & 25.76 & 25.78 \\
  ALNS HGS-init (1\,s) & 25.26 & \textbf{25.19} & \textbf{25.15} & 25.20 \\
  ALNS HGS-init (5\,s) & 25.28 & 25.24 & 25.20 & 25.23 \\
  ALNS HGS-init (60\,s) & \textbf{25.25} & 25.24 & 25.16 & \textbf{25.18} \\
  ALNS GREAT-init (1\,s) & 26.06 & 26.04 & 26.00 & 25.97 \\
  ALNS GREAT-init (5\,s) & 26.06 & 26.01 & 25.99 & 26.03 \\
  ALNS GREAT-init (60\,s) & 25.75 & 25.70 & 25.69 & 25.71 \\
  GREAT NB $\times 9$ & 26.89 & 26.77 & 26.86 & 26.76 \\
  GREAT NB $\times 9$ + 2opt & 26.44 & 26.34 & 26.37 & 26.32 \\
\midrule
\multicolumn{5}{l}{\emph{Battery 40\,kWh, demand $\{21,\ldots,29\}$}} \\
  Baseline (HGS) & 110.34 & 109.84 & 109.68 & 109.49 \\
  ALNS (1\,s) & 114.52 & 113.87 & 113.76 & 113.73 \\
  ALNS (5\,s) & 110.41 & 110.11 & 109.34 & 109.65 \\
  ALNS (60\,s) & 107.11 & 106.49 & 105.60 & 105.43 \\
  ALNS HGS-init (1\,s) & 109.62 & 109.12 & 109.01 & 108.80 \\
  ALNS HGS-init (5\,s) & 109.21 & 108.77 & 108.45 & 108.39 \\
  ALNS HGS-init (60\,s) & 106.96 & 106.32 & 105.57 & 105.19 \\
  ALNS GREAT-init (1\,s) & 105.13 & 104.09 & 103.24 & 102.54 \\
  ALNS GREAT-init (5\,s) & 105.13 & 104.09 & 103.24 & 102.53 \\
  ALNS GREAT-init (60\,s) & \textbf{105.10} & \textbf{104.08} & \textbf{103.18} & \textbf{102.45} \\
  GREAT NB $\times 9$ & 105.47 & 104.37 & 103.60 & 103.10 \\
  GREAT NB $\times 9$ + 2opt & 105.18 & 104.15 & 103.31 & 102.59 \\
\bottomrule
\end{tabular}
\end{table*}
\FloatBarrier

\subsection{\realrev{Auxiliary Routing Benchmarks (TSP, CVRP, OP)}}
\realrev{
Additionally to our extensive EVRP experiments, we also evaluate GREAT on several standard routing problems (TSP, OP and CVRP) in several distance distributions. The purpose of this evaluation is to compare GREAT against other, learning-based methods that have been developed and fine-tuned with those problem settings in mind. Further, simply applying and generalize those other learning-based methods to our EVRP setting can be difficult since they are often not applicable to asymmetric settings or have very different training pipelines (that require, e.g., ground truth labels for supervised learning).
The detailed results can be found in Appendix~\ref{sec:tspcvrpop_results}.
To summarize, we show that GREAT achieves competitive performance compared to existing neural methods, especially on asymmetric distance distributions, where it is often among the first learning-based method to be used to solve the problem altogether (e.g., asymmetric CVRP and OP variants).
GREAT generally outperforms MatNet which indicates that our purely edge-based approach is advantageous over a hybrid architecture such as MatNet.
We acknowledge, however, that GREAT cannot compete with purely node-focused Pointerformer on Euclidean TSP, an observation also reported by \cite{kwon2021matrix} for MatNet.
In addition to this in-distribution benchmark, we further show GREAT performs well out of distribution on standard ``real-world'' benchmarking dataset such as TSPLIB. Moreover, we investigate GREAT's generalization performance to large instances with finetuning using a curriculum learning pipeline but also without fine-tuning in a zero-shot generalization evaluation. We can show GREAT performs well in the zero-shot setting when applied to instances of with 200 nodes. However, it can additionally profit from the curriculum learning, especially on instances of size 500.
}

\section{Conclusion}
\label{sec:conclusion}
Due to the ongoing electrification of the transportation sector, energy-based routing problems are gaining increasing relevance, resulting in a need for fast and effective solution methods. 
While machine learning-based methods have successfully been applied to standard routing problems, the underlying GNN and transformer architectures are inherently limited to Euclidean coordinate-based settings. As a result, real-world asymmetric problems like the energy-based EVRP remain largely underexplored.
We introduce GREAT, a novel GNN-based architecture that operates directly on edge features through an attention-based mechanism between adjacent edges.
GREAT is natively suited to non-Euclidean or even asymmetric cost structures that are often found in real-world transportation problems. Thereby, it can overcome a fundamental limitation of existing GNN and transformer architectures. 

We develop a GREAT-based RL framework for the EVRP grounded in real-world conditions.
By using an established physics-based energy model applied to real road networks augmented with elevation data, we derive realistic energy consumption estimates that naturally give rise to asymmetric and negative energy costs.
We consider a range of battery capacities (15–40 kWh) and customer demand distributions that influence the charging decisions.
While we train the framework exclusively on data from San Francisco, it generalizes to other cities without retraining.
GREAT outperforms a strong HGS-based baseline in the majority of tested configurations, particularly those with small batteries and high demands, while being approximately $60\times$ faster \realrev{when using batched GPU processing}. 
\realrev{GREAT also outperforms an ALNS that optimizes routing and charging jointly in all considered settings when the ALNS time budget is 5s or less and in the majority of settings with a 60s time budget. 
Further, we point out that ALNS can profit considerably when using a GREAT-based warm initialization, even under small time budgets, showing that neural methods and metaheuristics can be complementary rather than competing.}

To further benchmark GREAT against standard learning-based methods, we evaluate it on classical routing problems (TSP, CVRP, OP). We show that GREAT is well-suited for such asymmetric routing settings, achieving state-of-the-art performance among neural methods for non-Euclidean instances and establishing, to our knowledge, the first neural results for asymmetric CVRP and OP.

Future directions include integrating GREAT into divide-and-conquer frameworks for fleet-scale EVRP and extending the setting to more involved problems, for example by including multiple vehicles, time windows and stochasticity. \rev{The memory footprint of GREAT scales as $\mathcal{O}(n^2)$ due to the dense edge representation, which limits direct training on very large instances. The introduced curriculum learning mitigates this for TSP and could be applied to CVRP/OP in future work. }

\rev{The primary advantage of GREAT over classical heuristics lies in its runtime: while state-of-the-art heuristics such as HGS can achieve better solution quality in unconstrained settings, they are orders of magnitude slower, making GREAT particularly attractive for real-time fleet management applications. Several concurrent and extending works have already picked up the edge-input paradigm introduced here \citep{meng2026uniteformer, ijcai2025p954, rydin2026beyond, rydin2026twostagelearneddecompositionscalable, jacobsen2026learningdialaridedeepgraph, lischka2026greateergraphedgeattention}, demonstrating its broad applicability.}

\section{Acknowledgments}

This work was performed as a part of the research project ``LEAR: Robust LEArning methods for electric vehicle Route selection'' funded by the Swedish Electromobility Centre (SEC). The computations were enabled by resources provided by the National Academic Infrastructure for Supercomputing in Sweden (NAISS) at Chalmers e-Commons partially funded by the Swedish Research Council through grant agreement no. 2022-06725.

\section{Declaration of generative AI and AI-assisted technologies in the manuscript preparation process}
During the preparation of this work, the authors used Claude for manuscript rewording and coding tasks. The authors reviewed and edited the output as needed and take full responsibility for the content of the published article.

\bibliographystyle{cas-model2-names}

\bibliography{cas-refs}

\appendix
\section{HGS Adaptation for EVRP}
\label{appendix_hgs}

As a baseline for the EVRP experiments, we use an adaptation of HGS \cite{vidal2012hybrid}. 
HGS itself is not applicable to EVRP since it cannot handle battery constraints. Therefore, in order to solve an EVRP instance, we apply HGS to the CVRP instance that results from the EVRP instance by ignoring battery constraints and removing the charger nodes (since those do not have to be visited in the solution tour). 
Afterwards, we transform the CVRP solution into an EVRP solution by inserting charger stops where necessary. 
We do this greedily, i.e. we insert a charging stop between two nodes $i, j$ in a tour when visiting the charger after node $j$ is no longer possible or it is not possible to reach node $j$ from node $i$ anymore. In such a case, we insert a node $k$ corresponding to a charger between $i$ and $j$ resulting in a subsequence $(i,k,j)$ where k is the nearest charger to node $i$.
We acknowledge that this way of post-processing CVRP solutions to obtain EVRP solutions is fast and simple but generally suboptimal because routing and charging decisions are optimized independently.

\section{Post-Processing GREAT EVRP Solutions}
\label{appendix_great_2opt}
GREAT is, due to masking operations, able to directly construct feasible solutions for the EVRP (and other routing problems).
However, as can be seen by the performance comparison to different baselines, solutions are not necessarily optimal.
Therefore, we introduce a simple, and very fast, post-processing method to improve solution quality.
The post-processing method has two phases:

\textbf{Phase 1: removing redundant stops.} We try to remove any unnecessary charging stops and depot visits. This means that given a tour subsequence $0, i_1, \dots, i_n, 0, j_1, \dots, j_k, 0$, where $0$ denotes the depot and $0 \notin \{i_1, \dots, i_n\} \cup \{j_1, \dots, j_k\}$, we check if the depot visit between node $i_n$ and $j_k$ is necessary or if the subtour $0, i_1, \dots, i_n, j_1, \dots, j_k, 0$ would also satisfy the vehicle capacity. An analogous check is performed for charger visits. Due to the triangle inequality not being satisfied in the energy-based distribution, it is possible that additional depot and charger stops result in better solution objectives. Therefore, in case a stop is unnecessary but beneficial, we still keep it.
  
\textbf{Phase 2: Local 2-opt on subtours.} We try to optimize the subsequences between charger and depot visits. I.e., we consider every subsequence $j, i_1, \dots, i_n, k$ where $j, k \in \{0\} \cup V_s$ and $i_1, \dots, i_n \notin \{0\} \cup V_s$. To optimize the subsequences, we apply 2-opt moves until convergence. Since we only apply 2-opt moves to such local subsequences, the procedure is generally fast (due to the subsequences being relatively short) and does not violate any capacity or charging constraints. This is because changing the order of customer visits between depot visits does not change the vehicle load. Similarly, further minimizing the energy-consumption of a subsequence that already respects battery constraints cannot lead to a violation of the same \footnote{Note that when minimizing the energy consumption of a subsequence, we have to consider the current SoC when entering this subsequence to ensure valid solutions due to the possibility of energy costs being negative. Consider the following example: $(0, i, j, c)$ where $i,j$ are customers, 0 is the depot and $c$ is a charger. Assume we have 60\% SoC at the depot, traveling to i increases our SoC to $90\%$, going to $j$ decreases it to $10\%$ and we arrive at the charger with slightly more than $0\%$ (approximately 10\% consumption). Therefore, we consume $30 - 80 -10 = -60 \%$ of our battery SoC. Assume that we switch the order to $(0, j, i, c)$ and traveling from 0 to $j$ costs 65\% SoC, from $j$ to $i$ regenerates $35\%$ and going from $i$ to the charger costs $10\%$ SoC. Clearly this results in less SoC consumption, however, we will run out of battery when going from 0 to $j$ when we start with less than 65\% SoC. Therefore, it is important to consider the starting SoC to prevent generating invalid solutions during the 2-opt improvements.}. Applying improvement operations more globally, on the other hand, would result in a much longer runtime and can easily violate constraints when applied without precaution.
The procedure adds approximately 0.01\,s per instance.

\section{\realrev{GPU Memory Limits of GREAT}}
\label{appendix_gpu_limits}

\realrev{In this appendix, we empirically investigate the GPU memory consumption and limits of GREAT during training and inference and compare it to existing node-based transformer models. GREAT computes edge features for $n^2$ many edges in a routing problem instance with $n$ nodes, while a node-based architecture computes only $n$ many node features. However, the attention mechanism of a transformer still has a complexity of $n^2$ when running the model. This study also motivates the few-shot curriculum learning pipeline of Section~\ref{sec:cl}.}

\realrev{In particular, we try to train a Pointerformer \citep{jin2023pointerformer} and a GREAT NB model for Euclidean TSP of sizes $n \in \{100, 200, \ldots, 1000\}$ for three batch sizes $B \in \{1, 10, 50\}$. 
Additionally to the training (where a step includes forward pass, backward pass, and optimizer step), we also test the models for inference (tour construction step) for the same batch and instance size combinations.  We use the same hyperparameters as in Section~\ref{sec:training_config} which, in particular means a hidden dimension of 128. As a consequence, we cannot include MatNet \citep{kwon2021matrix} in this experiment, as a hidden dimension of 128 limits it to instances of at most 128 nodes due to its one-hot initialization of the node features.
We use the same NVIDIA A40 GPU with 48GB of VRAM as in our other experiments.
}

\realrev{The results of the experiment can be found in Table~\ref{tab:gpu-limits-grid}. We can observe that for training, Pointerformer crashes for instances of 1000 nodes when the batch size is 1, whereas GREAT crashes for 900 nodes. Similarly, for batches of 10 instances, Pointerformer can handle 400 nodes whereas GREAT can only handle 300. When we set the batch size to 50, GREAT can only be trained for instances of 100 nodes whereas Pointerformer can handle 200 nodes. During inference, Pointerformer can handle all settings but instances with 900 nodes or more with a batch size of 50. GREAT, on the other hand crashes for batch size 10 already for instances of size 1000 and for instances of size 500 or larger when the batch size is 50. 
Overall, we can summarize that GREAT does indeed have an increased memory consumption compared to Pointerformer. However, we can clearly see that the overhead is far from being quadratic (e.g., training with a batch size of 10 fails for 400 vs. 500 nodes).}

\begin{table*}[tbp]
\centering
\color{black}
\footnotesize
\setlength{\tabcolsep}{4pt}
\caption{\realrev{Result of each test across models, batch sizes $B$ and instance sizes $n$, during training and inference. All models use hidden dim.\ 128 and run on a single GPU (NVIDIA A40) with 48GB of memory.}}
\label{tab:gpu-limits-grid}
\begin{tabular}{llrcccccccccc}
\toprule
 &  &  & \multicolumn{10}{c}{Instance size $n$} \\
\cmidrule(lr){4-13}
Model & Mode & $B$ & 100 & 200 & 300 & 400 & 500 & 600 & 700 & 800 & 900 & 1000 \\
\midrule
\multirow{6}{*}{Pointerformer} & \multirow{3}{*}{Train} & 1 & \rOK & \rOK & \rOK & \rOK & \rOK & \rOK & \rOK & \rOK & \rOK & \rOOM \\
 &  & 10 & \rOK & \rOK & \rOK & \rOK & \rOOM & \rSKIP & \rSKIP & \rSKIP & \rSKIP & \rSKIP \\
 &  & 50 & \rOK & \rOK & \rOOM & \rSKIP & \rSKIP & \rSKIP & \rSKIP & \rSKIP & \rSKIP & \rSKIP \\
\cmidrule(l){2-13}
 & \multirow{3}{*}{Infer} & 1 & \rOK & \rOK & \rOK & \rOK & \rOK & \rOK & \rOK & \rOK & \rOK & \rOK \\
 &  & 10 & \rOK & \rOK & \rOK & \rOK & \rOK & \rOK & \rOK & \rOK & \rOK & \rOK \\
 &  & 50 & \rOK & \rOK & \rOK & \rOK & \rOK & \rOK & \rOK & \rOK & \rOOM & \rSKIP \\
\midrule
\multirow{6}{*}{GREAT NB} & \multirow{3}{*}{Train} & 1 & \rOK & \rOK & \rOK & \rOK & \rOK & \rOK & \rOK & \rOK & \rOOM & \rSKIP \\
 &  & 10 & \rOK & \rOK & \rOK & \rOOM & \rSKIP & \rSKIP & \rSKIP & \rSKIP & \rSKIP & \rSKIP \\
 &  & 50 & \rOK & \rOOM & \rSKIP & \rSKIP & \rSKIP & \rSKIP & \rSKIP & \rSKIP & \rSKIP & \rSKIP \\
\cmidrule(l){2-13}
 & \multirow{3}{*}{Infer} & 1 & \rOK & \rOK & \rOK & \rOK & \rOK & \rOK & \rOK & \rOK & \rOK & \rOK \\
 &  & 10 & \rOK & \rOK & \rOK & \rOK & \rOK & \rOK & \rOK & \rOK & \rOK & \rOOM \\
 &  & 50 & \rOK & \rOK & \rOK & \rOK & \rOOM & \rSKIP & \rSKIP & \rSKIP & \rSKIP & \rSKIP \\
\bottomrule
\end{tabular}

\vspace{2pt}
{\scriptsize \rOK~ran to completion \quad \rOOM~CUDA out of memory \quad \rSKIP~not attempted (the sweep halts at the first failure).}
\end{table*}

\section{\realrev{Effects of Scaling-Based Augmentation and Post-processing}} %
\label{appendix_aug}
\realrev{As described in Section~\ref{sec:training_config}, we apply a scaling-based data augmentation method when using GREAT. 
Similarly to \cite{kwon2020pomo} where Euclidean data in the unit square is augmented by coordinate rotations, we scale our data by multiplying all pairwise distances of a VRP with factors in the range $(0.5, 1.5)$. Since those ``distances'' correspond to the energy consumption when traversing an edge, we additionally scale the battery size with the same factor to obtain an equivalent EVRP. This augmentation allows us to solve several equivalent EVRP instances simultaneously and, by selecting the best solution achieved on any of those augmented instances as our ``final solution'', improve the performance of our model. 
}

\realrev{
Additionally to the augmentation method,  we also apply a simple post-processing to the solutions of the neural model as described in Section~\ref{sec:postprocessing} and Appendix~\ref{appendix_great_2opt}.
}

\realrev{In the following, we report how much the performance of GREAT changes when augmentation and post-processing are applied. The results can be found in Table~\ref{tab:inference-ablation}. This table shows the average performance of the augmentation factors (x1 with 5 and 100 POMO rollouts and x9 with 100 POMO rollouts) with and without 2-opt post-processing over all 36 test configurations of Tables~\ref{tab:SF}, \ref{tab:Gothenburg}, and~\ref{tab:Vienna} in Section~\ref{sec:evrp_results} where every configuration contains 100 instances. Note that the GREAT results in those main result tables were achieved with x9 augmentation and post-processing with 100 POMO rollouts. Additionally to running the experiments on a GPU in batches, we also report the runtime when running them sequentially on a CPU instead. We can see that x9 augmentation (with 100 POMO rollouts) results in approximately five times the runtime (over 100 POMO rollouts with no augmentation) on both GPU and CPU (without 2-opt post-processing). Post-processing itself adds approximately 1s of runtime no matter the augmentation factor or hardware. It is worth pointing out that x1 augmentation with post-processing results in better performance than x9 augmentation without post-processing. We further note that while using CPUs results in a significantly increased runtime, GREAT is still faster than the HGS baseline which generally took 5.7 - 10.9 min.
Regarding the number of POMO rollouts, we can see that when using batched GPU processing, there is no runtime advantage at all using only 5 rollouts compared to 100. When running on a CPU the runtime decreases from 25s to 14.1s, a $43.6\%$ improvement even though only $5\%$ of rollouts are processed. Performance-wise, 5 POMO rollouts with post-processing are better than 100 POMO rollouts without post-processing (no augmentation). 
We further note that POMO is an integral part of our framework since we always apply a POMO step as the initial step and the model therefore is not trained to choose the initial step itself. In our ablation for Table~\ref{tab:inference-ablation}, we therefore manually choose the 5 nearest customers to the depot as the initial step.
}

\begin{table}
  \centering
  \color{black}
  \caption{\realrev{Average performance of GREAT NB $\times 9$ and GREAT NB $\times 1$, with and without 2-opt post-processing against HGS baseline, averaged over all 36 city/battery/demand configurations (100 instances per configuration). GPU solves batches of EVRP instances; CPU runs the same model sequentially, one instance at a time. CPU and GPU achieve the same gaps (same model, same resulting tours), while runtime depends on the hardware. Runtime is the average wall-clock time for solving the 100 instances of one configuration.}}
  \label{tab:inference-ablation}
\begin{tabular}{lrrr}
  \toprule
  Method & opt. gap (\%) & time GPU & time CPU (seq.) \\
  \midrule
  GREAT NB $\times 9$, POMO 100 & -1.13 & 5.2\,s & 3.4\,min \\
  GREAT NB $\times 9$, POMO 100 + 2opt & \textbf{-2.10} & 6.2\,s & 3.5\,min \\
  GREAT $\times 1$, POMO 100 & -0.50 & \underline{1.0\,s} & 25.0\,s \\
  GREAT $\times 1$, POMO 100 + 2opt & \underline{-1.61} & 2.0\,s & 26.0\,s \\
  GREAT $\times 1$, POMO 5 & 0.74 & \textbf{0.9\,s} & \textbf{14.1\,s} \\
  GREAT $\times 1$, POMO 5 + 2opt & -0.59 & 2.0\,s & \underline{15.2\,s} \\
  \bottomrule
  \end{tabular}
\end{table}

\subsection*{\realrev{Augmentation During Training}}
\realrev{To familiarize our model to scaled EVRP instances, we apply scaling not only during inference but also during training.
In the following, we investigate whether this is necessary or whether it suffices to apply scaling during inference only.
To do so, we train a second EVRP model that is completely identical to the one used in the main results of this paper. This means, we use the exact same architecture, hyperparameters, training budget and training data. The only difference is whether we apply scaling to the training instances or not. 
We then evaluate both trained models on the same test dataset.
The results of this comparison can be found in Table~\ref{tab:aug-effect}. We present the results according to the configuration and group them based on battery capacity, city and customer demand range. We present the performance with (x9) and without augmentation (x1). In this setting the results in the table do not include any post-processing.
According to the table, augmentation during training does not have a positive effect on the result. In fact, contrary to augmentation during inference, it seems to have a slightly negative influence as all results are slightly negative. We note that all results in the paper stem, nevertheless, unless otherwise specified, from the model that was exposed to augmentation during training.
} 

\begin{table}[ht]
\centering
\color{black}
\footnotesize
\caption{\realrev{Effect of augmentation during training on EVRP performance. The entries show the mean per-instance relative difference (\%) between the model trained \emph{without} augmentation and the model trained \emph{with} augmentation. \emph{Positive} values would indicate that training-time augmentation helped. We consider 100 test instance per configuration and each block pools over the remaining two factors.}}
\label{tab:aug-effect}
  \begin{tabular}{lrr}
  \toprule
   & \multicolumn{2}{c}{Inference augmentation} \\
  \cmidrule(lr){2-3}
  Grouping & $\times 1$ & $\times 9$ \\
  \midrule
  \multicolumn{3}{l}{\emph{By battery capacity (kWh)}} \\
  15 & \text{-0.49} & \text{-0.44} \\
  20 & \text{-0.26} & \text{-0.20} \\
  30 & \text{-0.13} & \text{-0.07} \\
  40 & -0.06 & -0.02 \\
  \midrule
  \multicolumn{3}{l}{\emph{By city}} \\
  San Francisco  & \text{-0.11} & \text{-0.10} \\
  Gothenburg  & \text{-0.33} & \text{-0.25} \\
  Vienna  & \text{-0.26} & \text{-0.20} \\
  \midrule
  \multicolumn{3}{l}{\emph{By customer demand range}} \\
  $\{1,\ldots,9\}$ & \text{-0.61} & \text{-0.47} \\
  $\{11,\ldots,19\}$ & \text{-0.07} & \text{-0.06} \\
  $\{21,\ldots,29\}$ & -0.02 & -0.02 \\
  \bottomrule
  \end{tabular}
\end{table}

\section{\realrev{Encoder Ablation: GREAT on the Raw Edge Costs}}
\label{appendix_encoder_ablation}

\realrev{
In Section~\ref{sec:training_config} we acknowledge that the comparison between the GREAT and MatNet encoder is not a complete ablation on the encoder itself due to GREAT obtaining 4 additional edge-based inputs compared to MatNet. 
This is due to the original MatNet architecture being incompatible with multidimensional edge inputs. 
To demonstrate that the advantage of GREAT stems not only from the design of the input data, we train an additional GREAT model that also only obtains the raw 1D edge costs as its input, just like MatNet.
In particular, we train a model for TSP on the TMAT distribution, since TSP is the purest ``edge-only-input'' problem and MatNet was originally introduced for TMAT.
The training pipeline for the 1D edge input GREAT is exactly the same as for the default GREAT model and the MatNet model. All hyperparameters (such as hidden dimension, batch size, number of training epochs and amount of training data) are also equivalent. 
}

\realrev{We evaluate all models on the same test dataset of $10,000$ instances. The results can be observed in Table~\ref{tab:encoder-ablation}. 
We can see that GREAT does profit from the additional edge input features, however, only slightly. The GREAT encoder operating on the raw edge costs achieves a performance much closer to the GREAT model with engineered edge features than to MatNet. This confirms that GREAT generates better encodings than MatNet even when operating on raw edge costs.
}

\begin{table}[ht]
\centering
\color{black}
\footnotesize
\setlength{\tabcolsep}{4pt}
\caption{\realrev{Encoder ablation on TSP with TMAT distribution. Optimality gaps (\%) are w.r.t. to Gurobi over 10{,}000 test instances. ``$c_{ij}$ only'' denotes the raw edge costs, and $d_{\mathrm{in}}$ is input dimension per edge.}}
\label{tab:encoder-ablation}
\begin{tabular}{llcrrr}
\toprule
Model & Edge input & $d_{\mathrm{in}}$ & $\times 1$ & $\times 9$ & $\times 33$ \\
\midrule
MatNet & $c_{ij}$ only & 1 & 7.54 & 5.28 & 4.40 \\
GREAT NB & $c_{ij}$ + 4 engineered & 5 & \textbf{1.52} & \textbf{1.07} & \textbf{1.03} \\
GREAT NB (ablation) & $c_{ij}$ only & 1 & 1.78 & 1.19 & 1.12 \\
\bottomrule
\end{tabular}
\end{table}

\section{\realrev{Statistical Uncertainties and Operational Diagnostics}}
\label{appendix:stats}

\realrev{The main results in Section~\ref{sec:evrp_results} in Tables~\ref{tab:SF}--\ref{tab:Vienna} report the averages over 100 test instances per configuration. In this appendix, we additionally report per-instance uncertainties and operational statistics for more detailed analysis on the same test instances. We reiterate here that the 100 test instances of a configuration (battery-demand-city) are the same for all algorithms (HGS, GREAT and other methods), making the results comparable between them.}

\realrev{Additionally to the mean gap of GREAT (x9 with post-processing)  w.r.t. the HGS-baseline, Table~\ref{tab:paired-stats} reports the 95\% confidence interval of the mean gap and the ``win rate'', i.e. the number of instances (out of 100) where GREAT achieves a better solution than the baseline. We can see that for the $\{21, \dots, 29\}$ demand case, our method outperforms the baseline in virtually all cases. In the setting with demands in $\{21, \dots, 29\}$, the 95\% confidence interval is generally still mostly negative and the win rate is mostly more than 50\% (the Gothenburg setting with 40kWh battery is the only exception). For low customer demands, the situation is more mixed, with our method generally being competitive when small batteries are used and the baseline leading when larger batteries are used.
An observation worth noticing is that all algorithms (GREAT, ALNS and the two-stage HGS-baseline) only produce feasible tours by design, so the failure rate when constructing solutions is 0\% for all methods.
}

\realrev{The operational diagnostics are shown in Table~\ref{tab:diagnostics}.
We show the four edge cases for San Francisco with the largest and smallest battery sizes as well as the lowest and highest customer demand distribution. We show how often the depot and chargers are visited on average for each configuration. Further, we show the mean SoC that the vehicle has left when going to a charger and the average minimum SoC (per tour) recorded when going to a charger.
While we can see there is hardly any difference in the number of depot visits between the methods, we can see that GREAT-based solutions generally charge more often and, as a result, have higher SoCs when going to the charger. Similarly, the ALNS which also jointly optimizes charging and routing (compared to the HGS-based method which inserts charging into a valid route in a two-stage approach) also visits the charger much more often than HGS, even though not as often as the learning-based solution.
This underlines that GREAT learns to take advantage of inserting cheap charging stops (which can be very cheap or even beneficial due to the triangle inequality not necessarily holding on our data) into the solution.}

\begin{table*}[ht]
\centering
\color{black}
\footnotesize
\setlength{\tabcolsep}{3.5pt}
\caption{\realrev{Comparison of GREAT NB $\times 9$ + 2opt against the HGS baseline. Table shows mean \emph{gap}(negative means GREAT consumes less energy), \emph{95\% CI} of the mean gap, and \emph{win} the number of instances (out of 100) on which GREAT strictly beats HGS.}}
\label{tab:paired-stats}
  \begin{tabular}{llrcrrcrrcr}
  \toprule
   &  & \multicolumn{3}{c}{Customer Demand [1, 9]} & \multicolumn{3}{c}{Customer Demand [11, 19]} & \multicolumn{3}{c}{Customer Demand [21, 29]} \\
  \cmidrule(lr){3-5} \cmidrule(lr){6-8} \cmidrule(lr){9-11}
  City & Battery (kWh) & gap & 95\% CI & win & gap & 95\% CI & win & gap & 95\% CI & win \\
  \midrule
  \multirow{4}{*}{San Francisco} & 15 & 0.11 & [-0.57,\,0.80] & 52 & -5.64 & [-6.34,\,-4.94] & 96 & -8.78 & [-9.54,\,-8.02] & 100 \\
   & 20 & 1.31 & [0.61,\,2.02] & 30 & -4.61 & [-5.32,\,-3.89] & 90 & -7.56 & [-8.33,\,-6.78] & 100 \\
   & 30 & 1.96 & [1.46,\,2.45] & 20 & -3.33 & [-3.94,\,-2.73] & 86 & -6.51 & [-7.25,\,-5.77] & 99 \\
   & 40 & 3.63 & [3.14,\,4.11] & 10 & -2.68 & [-3.24,\,-2.12] & 80 & -5.97 & [-6.70,\,-5.23] & 99 \\
  \midrule
  \multirow{4}{*}{Gothenburg} & 15 & -3.15 & [-3.93,\,-2.37] & 83 & -2.89 & [-3.40,\,-2.38] & 86 & -3.84 & [-4.26,\,-3.42] & 98 \\
   & 20 & -1.39 & [-2.03,\,-0.74] & 64 & -2.24 & [-2.70,\,-1.77] & 80 & -3.15 & [-3.48,\,-2.81] & 100 \\
   & 30 & -0.10 & [-0.68,\,0.48] & 45 & -0.60 & [-0.91,\,-0.29] & 58 & -2.21 & [-2.47,\,-1.95] & 95 \\
   & 40 & 1.72 & [1.29,\,2.15] & 22 & -0.27 & [-0.61,\,0.08] & 47 & -1.49 & [-1.72,\,-1.26] & 92 \\
  \midrule
  \multirow{4}{*}{Vienna} & 15 & -0.61 & [-1.28,\,0.05] & 55 & -3.42 & [-3.91,\,-2.93] & 93 & -4.91 & [-5.36,\,-4.46] & 100 \\
   & 20 & 0.23 & [-0.34,\,0.80] & 43 & -2.42 & [-2.85,\,-1.98] & 86 & -3.97 & [-4.40,\,-3.54] & 100 \\
   & 30 & 1.90 & [1.33,\,2.46] & 26 & -1.14 & [-1.53,\,-0.74] & 68 & -3.21 & [-3.61,\,-2.82] & 100 \\
   & 40 & 2.45 & [2.07,\,2.83] & 10 & -0.44 & [-0.76,\,-0.12] & 60 & -2.59 & [-2.97,\,-2.21] & 98 \\
\bottomrule
\end{tabular}
\end{table*}

\begin{table*}[ht]
\centering
\color{black}
\footnotesize
\setlength{\tabcolsep}{4pt}
\caption{\realrev{Operational diagnostics for San Francisco. \emph{depot} is the number of depot visits, \emph{charger} the number of charging stops, and the two SoC columns are the mean and the minimum state of charge on arrival at a charger.}}
\label{tab:diagnostics}
  \begin{tabular}{llrrrrrrrr}
  \toprule
   &  & \multicolumn{4}{c}{Customer Demand [1, 9]} & \multicolumn{4}{c}{Customer Demand [21, 29]} \\
  \cmidrule(lr){3-6} \cmidrule(lr){7-10}
  Battery (kWh) & Method & depot & charger & mean SoC & min SoC & depot & charger & mean SoC & min SoC \\
  \midrule
  \multirow{3}{*}{15} & Baseline (HGS) & 10.52 & 2.49 & 0.50 & 0.01 & 54.53 & 12.76 & 0.45 & 0.01 \\
   & ALNS (60\,s) & 11.06 & 2.91 & 0.56 & 0.04 & 55.20 & 20.70 & 0.60 & 0.03 \\
   & GREAT NB $\times 9$ + 2opt & 10.81 & 3.68 & 0.62 & 0.13 & 54.73 & 27.10 & 0.68 & 0.12 \\
  \midrule
  \multirow{3}{*}{40} & Baseline (HGS) & 10.52 & 0.55 & 0.54 & 0.04 & 54.53 & 4.05 & 0.50 & 0.01 \\
   & ALNS (60\,s) & 11.12 & 0.87 & 0.57 & 0.06 & 55.10 & 11.34 & 0.67 & 0.06 \\
   & GREAT NB $\times 9$ + 2opt & 10.85 & 1.01 & 0.64 & 0.25 & 54.82 & 20.38 & 0.78 & 0.35 \\
\bottomrule
\end{tabular}
\end{table*}

\section{\realrev{Validation Curves During Training}}
\label{appendix_val_curves}

\begin{figure}
    \centering
    \includegraphics[width=0.5\linewidth]{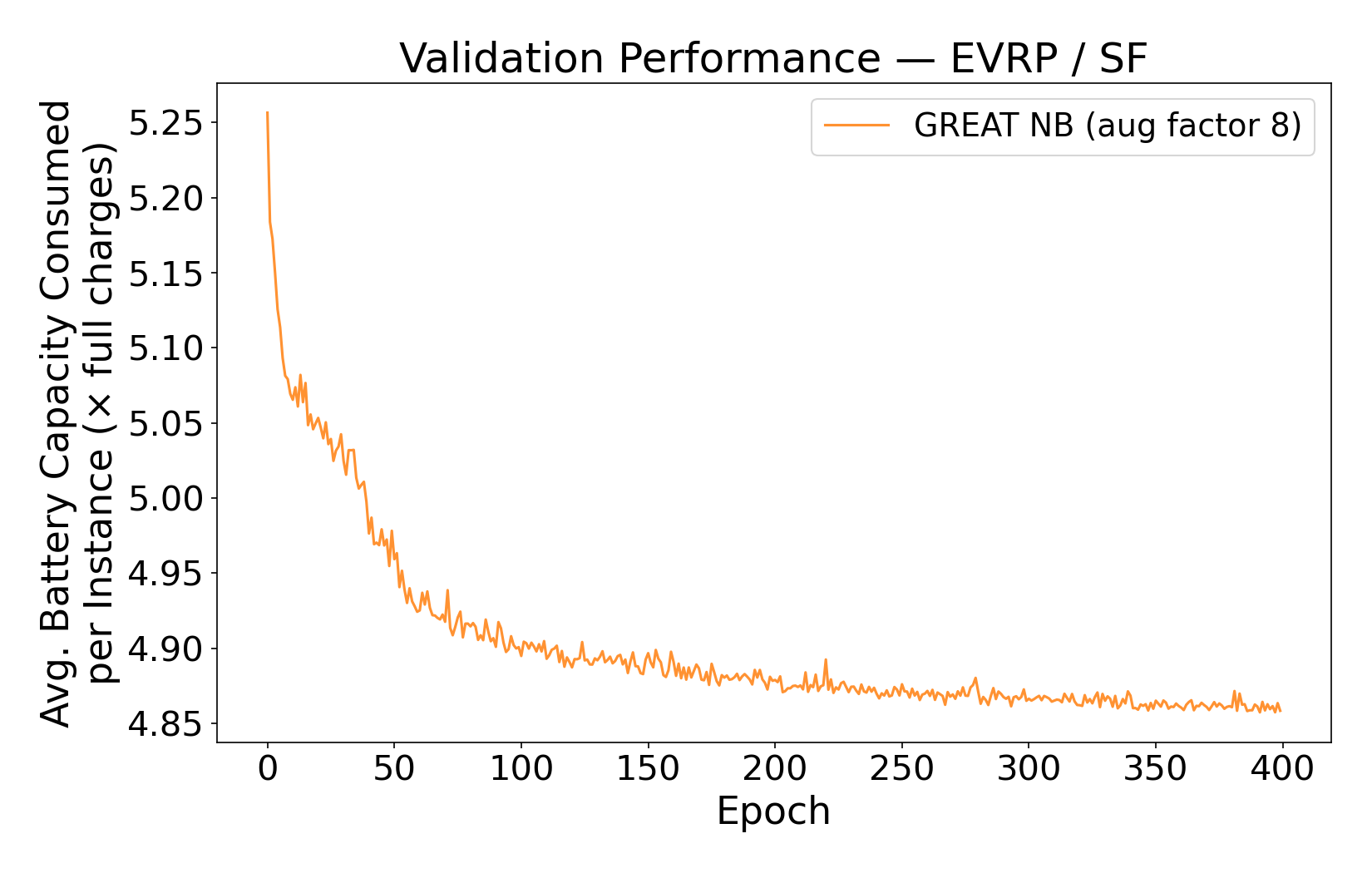}
    \caption{\realrev{Validation performance over 400 training epochs. The y-axis shows the average amount of battery charge that was consumed on the 1,000 validation instances.}}
    \label{fig:val_evrp}
\end{figure}

\begin{figure*}
\centering
\begin{subfigure}[b]{0.32\textwidth}
    \centering
    \includegraphics[width=\textwidth]{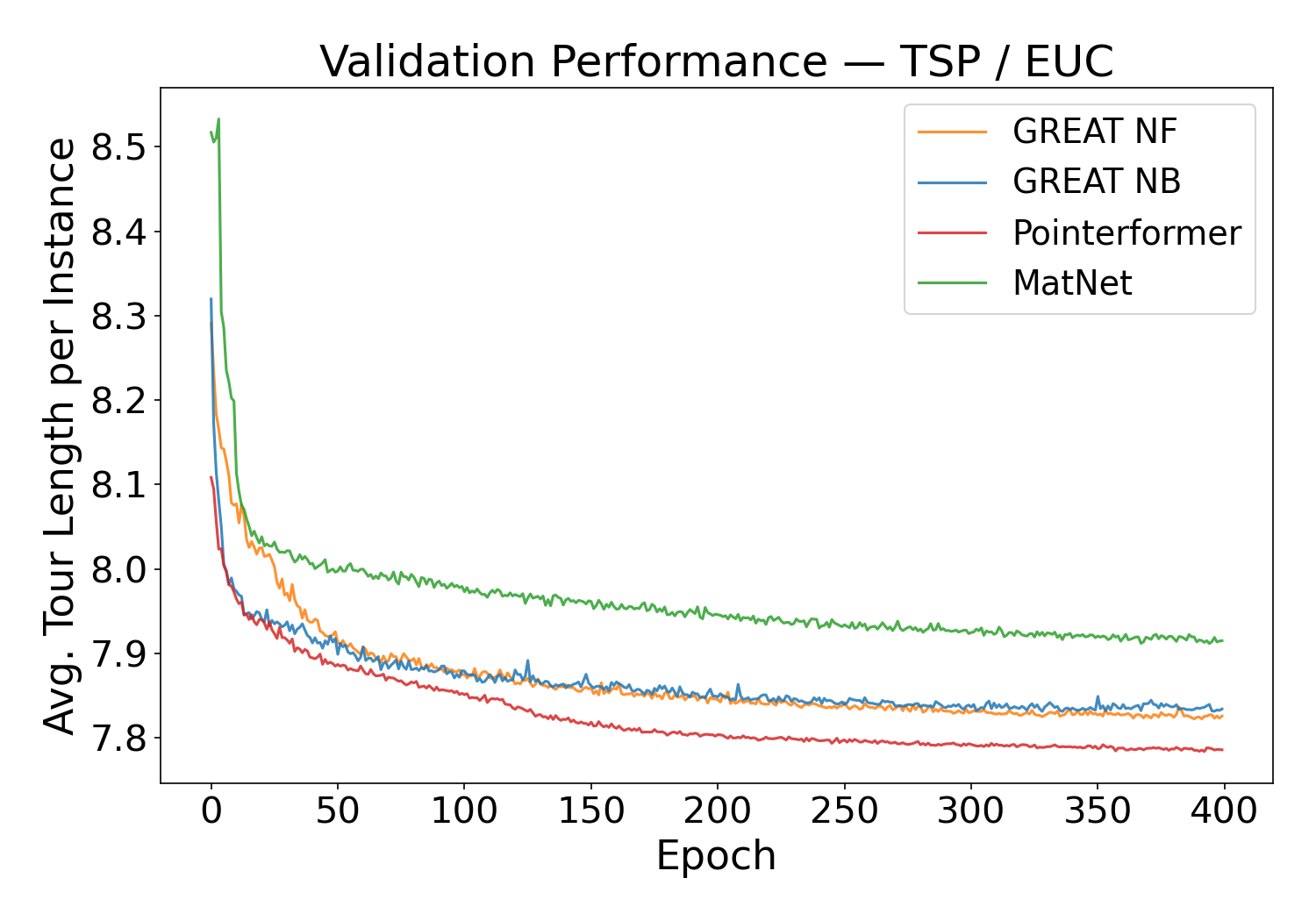}
\end{subfigure}
\hfill
\begin{subfigure}[b]{0.32\textwidth}
    \centering
    \includegraphics[width=\textwidth]{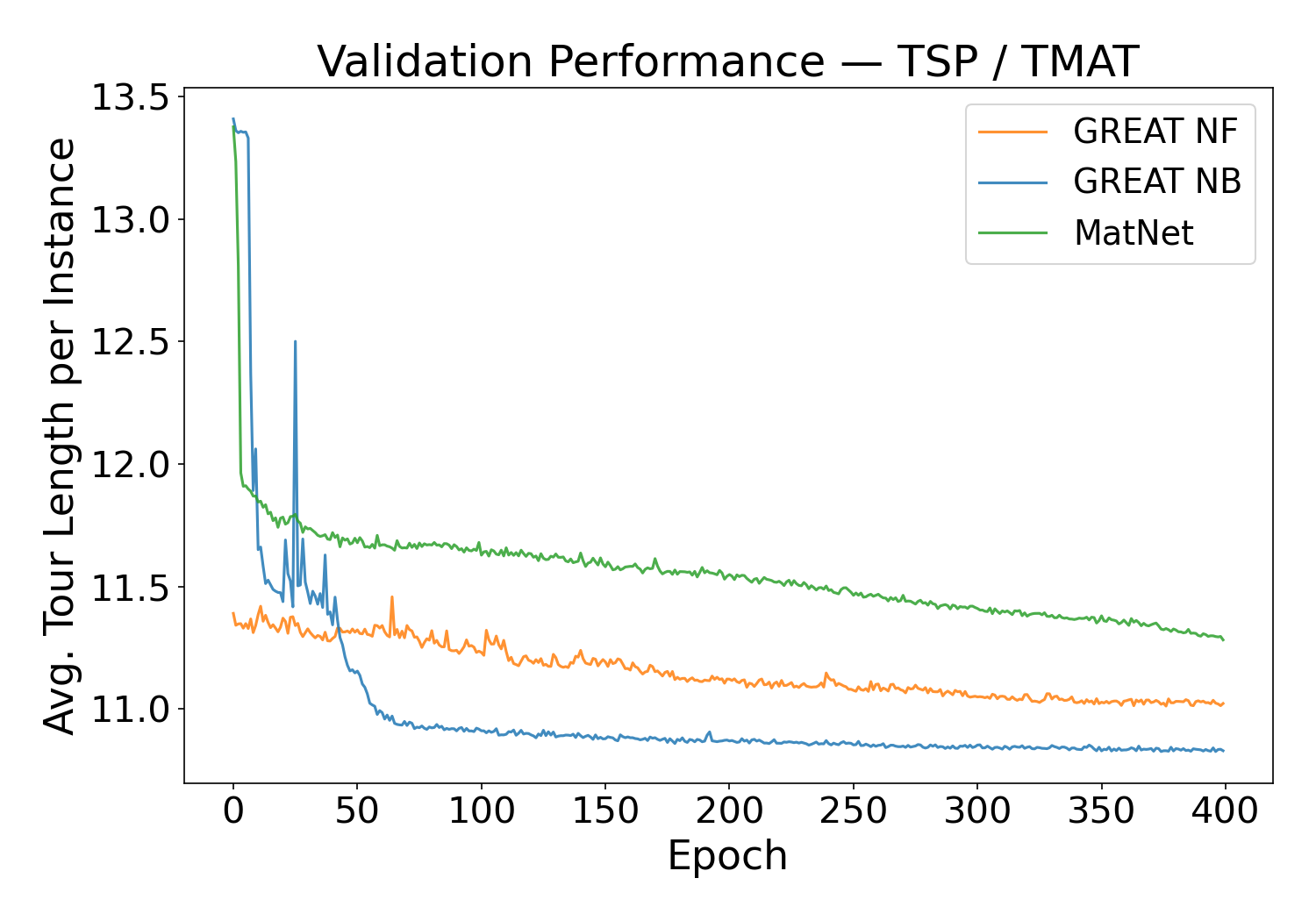}
\end{subfigure}
\hfill
\begin{subfigure}[b]{0.32\textwidth}
    \centering
    \includegraphics[width=\textwidth]{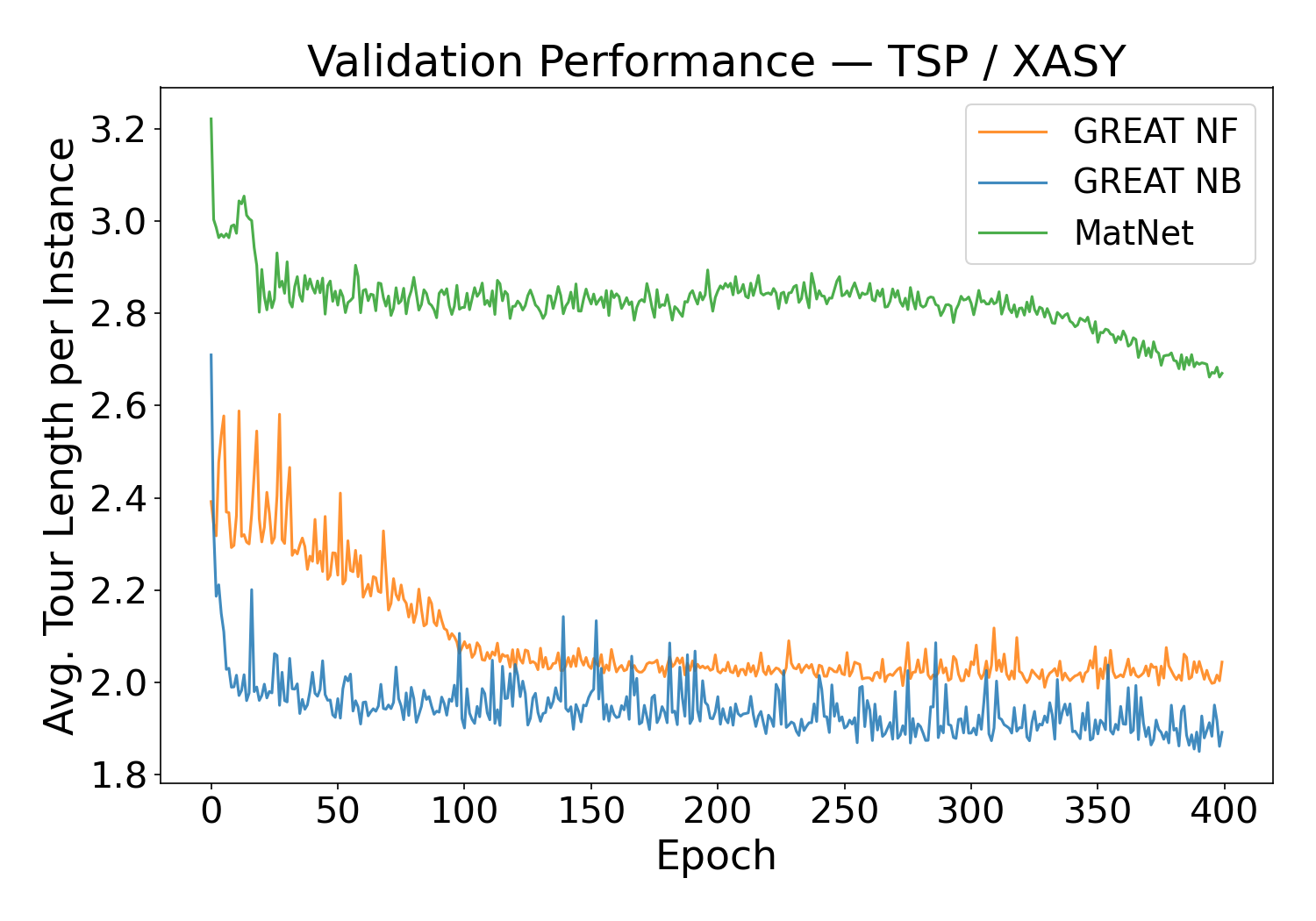}
\end{subfigure}

\vspace{0.6em}

\begin{subfigure}[b]{0.32\textwidth}
    \centering
    \includegraphics[width=\textwidth]{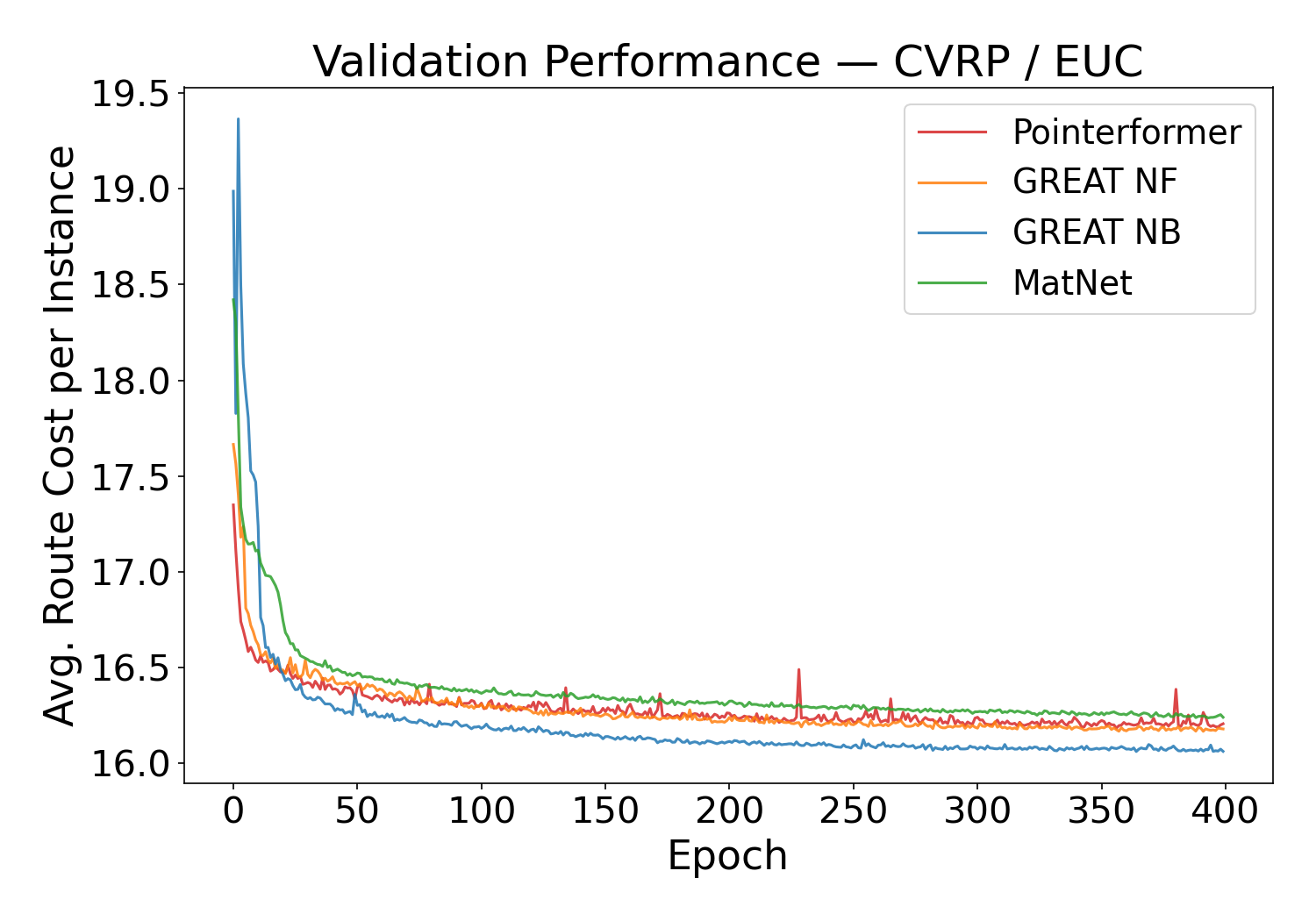}
\end{subfigure}
\hfill
\begin{subfigure}[b]{0.32\textwidth}
    \centering
    \includegraphics[width=\textwidth]{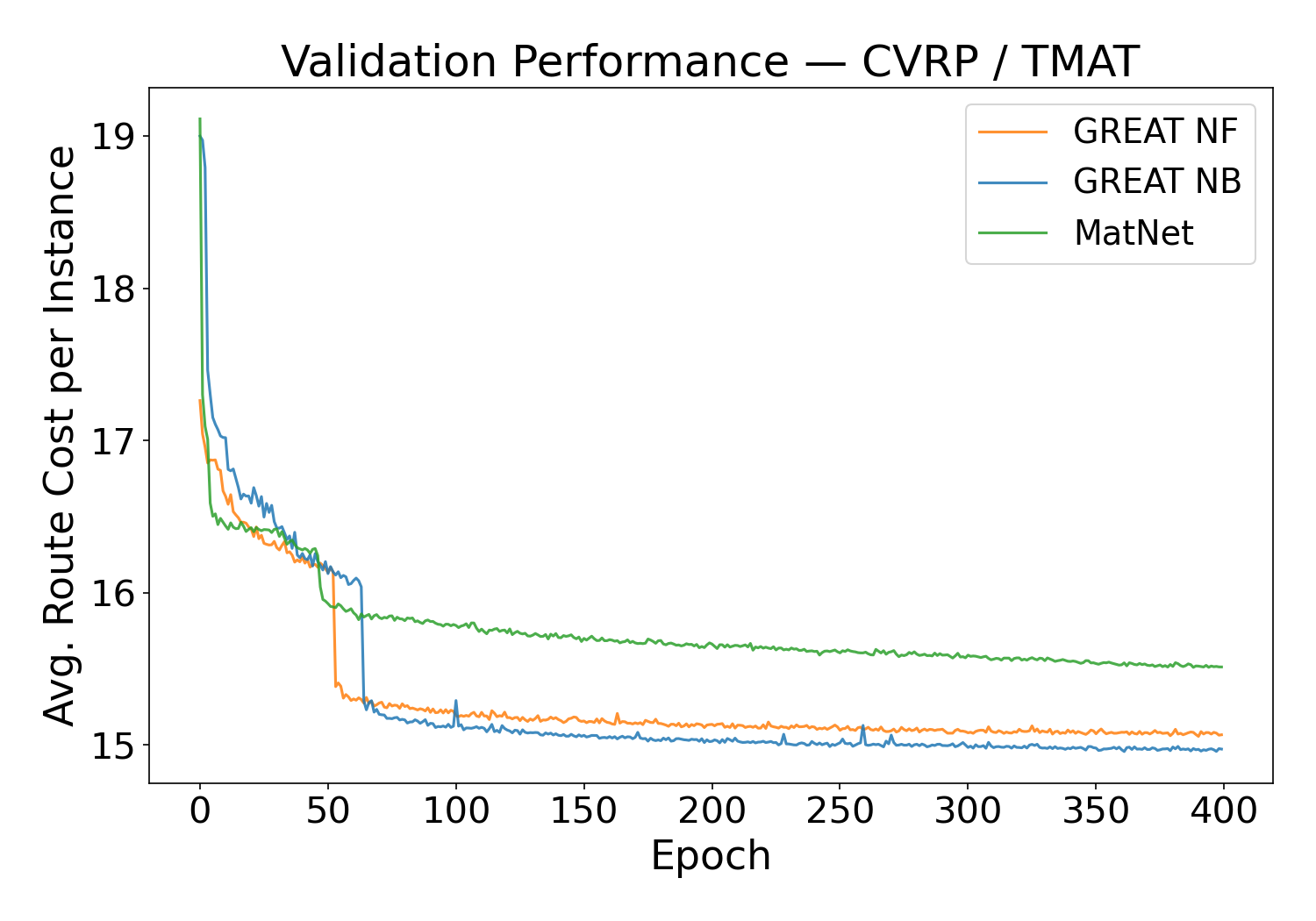}
\end{subfigure}
\hfill
\begin{subfigure}[b]{0.32\textwidth}
    \centering
    \includegraphics[width=\textwidth]{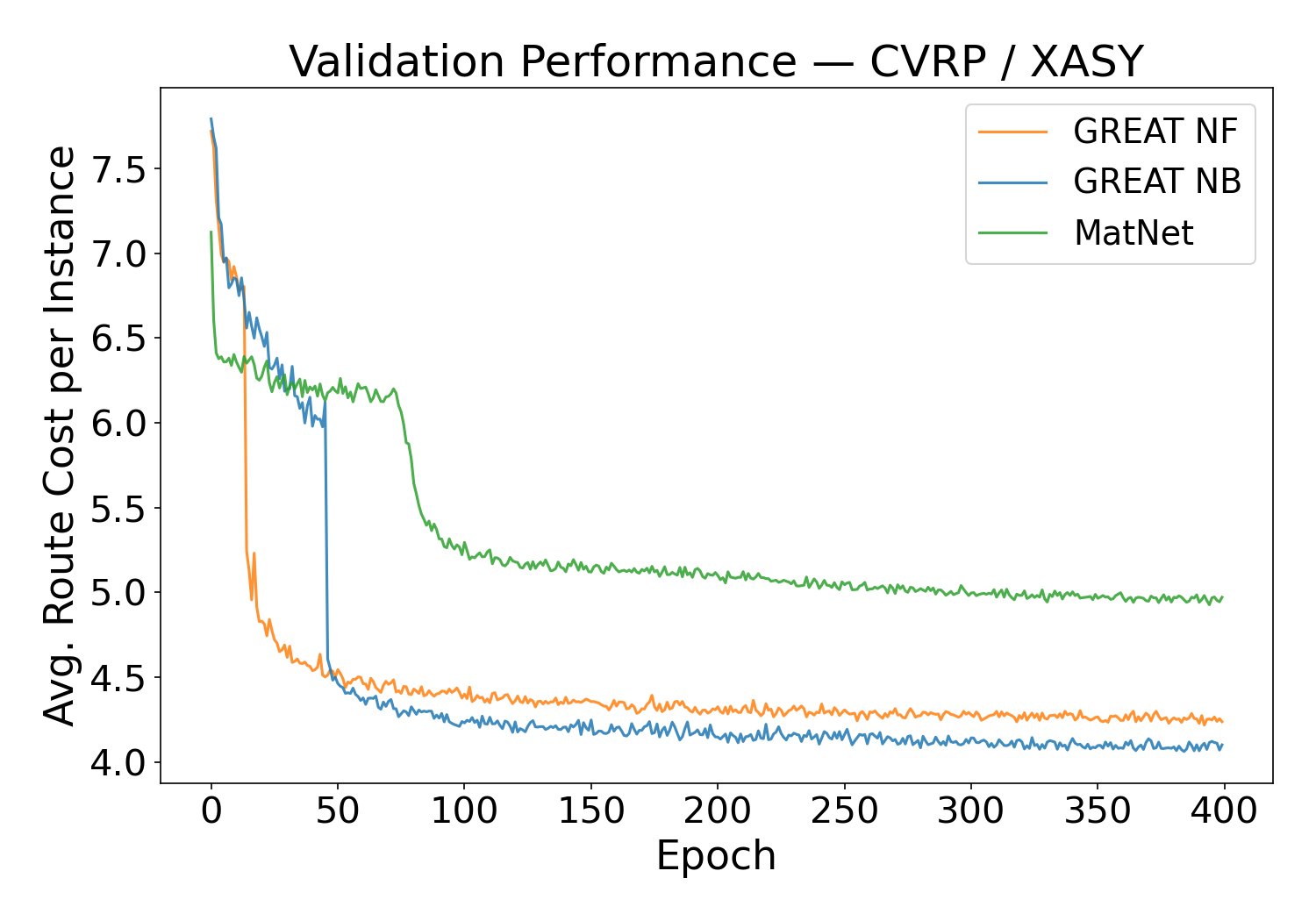}
\end{subfigure}

\vspace{0.6em}

\begin{subfigure}[b]{0.32\textwidth}
    \centering
    \includegraphics[width=\textwidth]{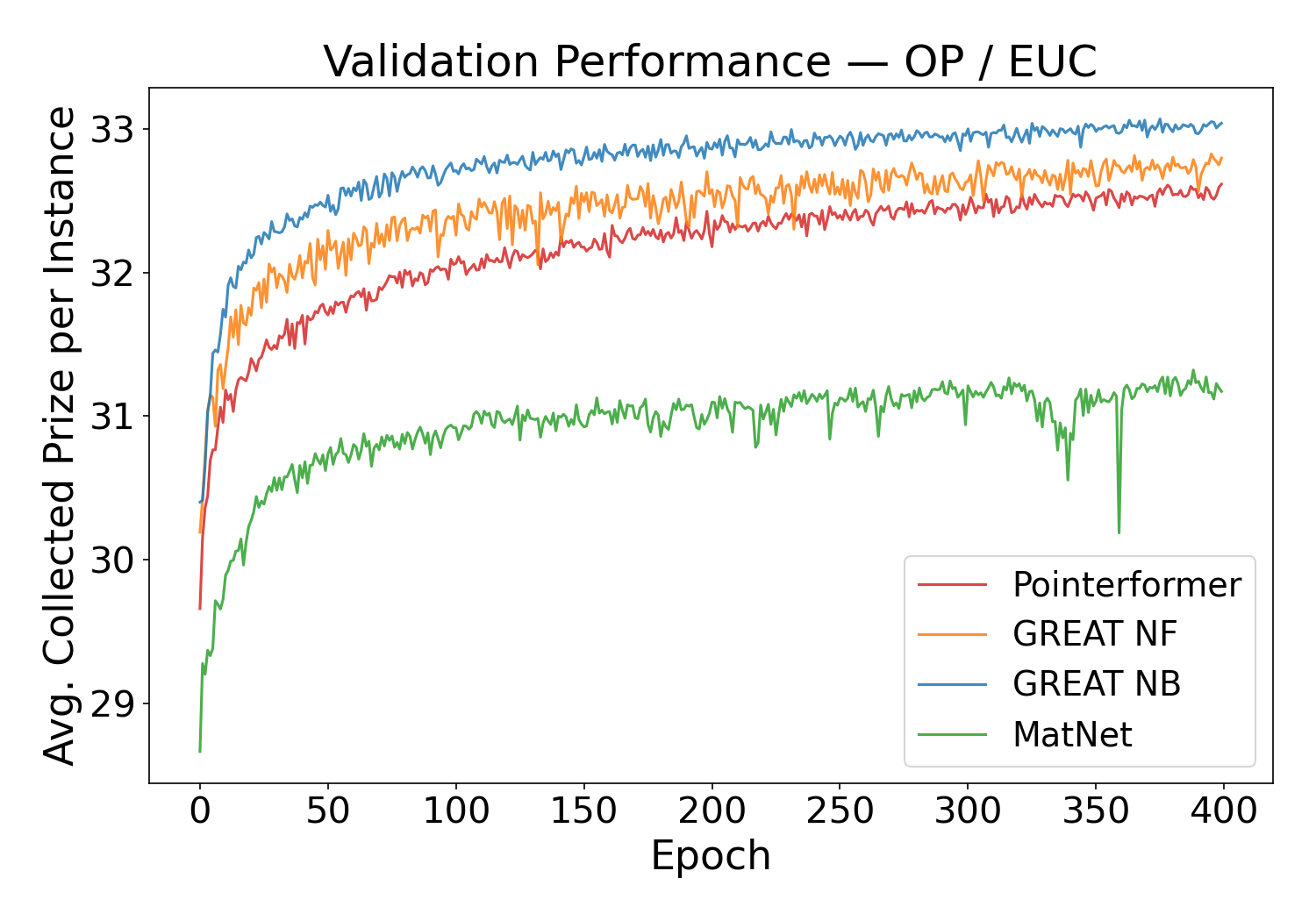}
\end{subfigure}
\hfill
\begin{subfigure}[b]{0.32\textwidth}
    \centering
    \includegraphics[width=\textwidth]{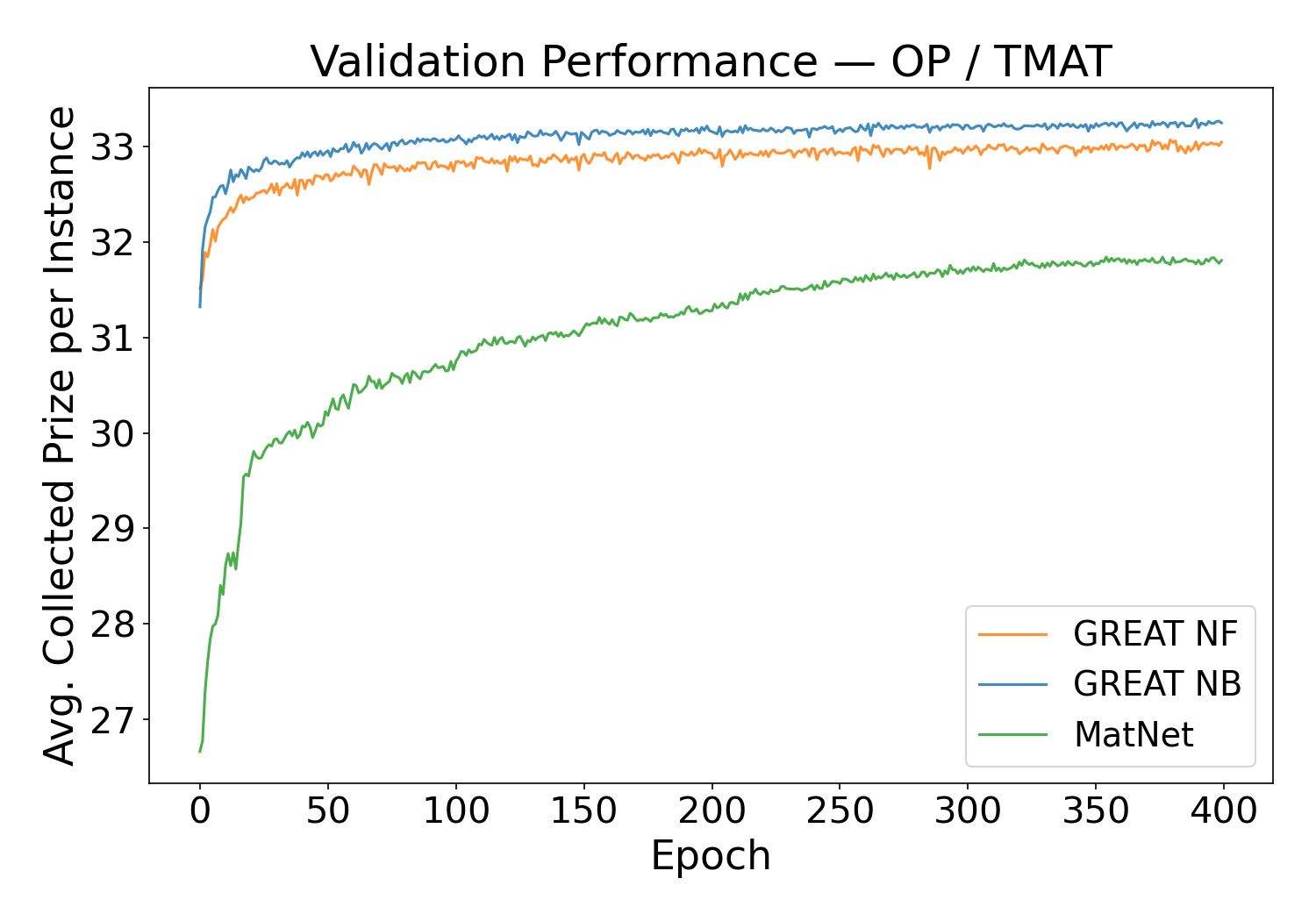}
\end{subfigure}
\hfill
\begin{subfigure}[b]{0.32\textwidth}
    \centering
    \includegraphics[width=\textwidth]{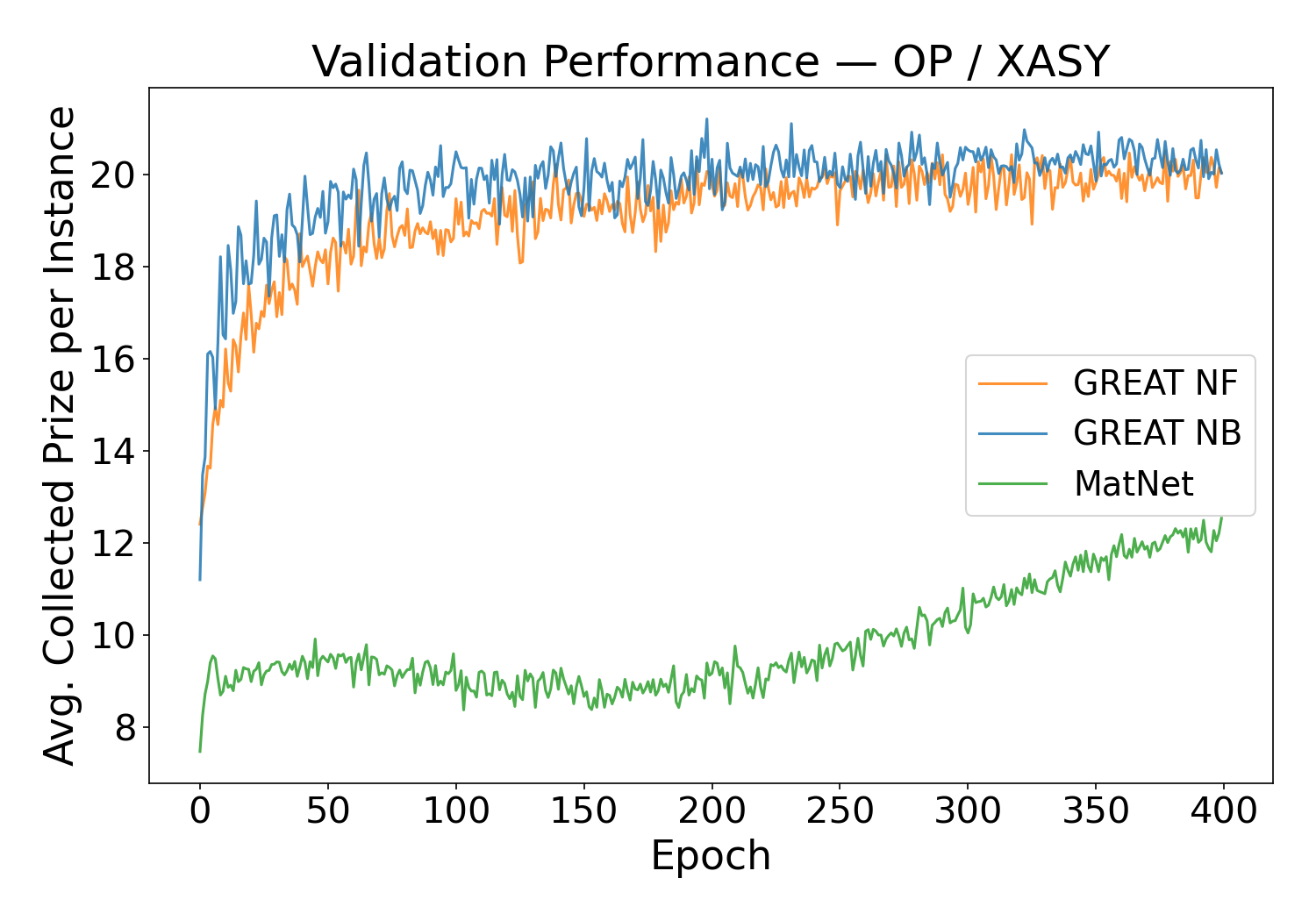}
\end{subfigure}
\caption{\realrev{Validation performance over the 400 training epochs, for TSP (top row), CVRP (middle row) and OP (bottom row) under the Euclidean (left), TMAT (middle) and XASY (right) distributions. TSP and CVRP show average tour length (lower is better) and OP show average collected prize (higher is better). Pointerformer is shown for the Euclidean distributions only, as it requires node coordinates.}}
\label{fig:val_curves}
\end{figure*}

\realrev{
In the following, we show the validation ``loss'' curves of all models while training for 400 epochs. This means, that after every epoch of training, we report the average performance of the model on the fixed validation dataset of $1,000$ instances that we described in Section~\ref{sec:training_config}.
For EVRP, the resulting plot can be found in Figure~\ref{fig:val_evrp}. The curve shows how much of the battery capacity was consumed on average over the $1,000$ validation instances. A value of $5$ means that on average $500\%$ of the SoC was consumed which implies several recharging visits.
For TSP, CVRP, and OP, the plots can be found in Figure~\ref{fig:val_curves} and show the average tour length (TSP and CVRP) and average collected prize (OP).
The curve for EVRP in Figure~\ref{fig:val_evrp} seems to have converged. Similarly, most curves for the auxiliary routing problems seem to have converged well, with the exception of the MatNet model in the TSP and OP case on the XASY distribution. However, since training has been performed for 400 epochs with $25,000$ instances per epoch already, further training seems excessive and would make the comparison between models unfair.
}


\section{Full Auxiliary Routing Benchmarks (TSP, CVRP, OP) Results}
\label{sec:tspcvrpop_results}

\subsection{Standard In-Distribution Benchmark}

We report results for instances of 100 nodes on 10,000 test instances in Table~\ref{table:GREAT} and Table~\ref{table:GREAT2}. We include different baselines depending on the problem: exact Gurobi \citep{gurobi} formulation and LKH \citep{lkh3} for TSP, HGS for CVRP and EA4OP \citep{kobeaga2018efficient} and a custom greedy baseline (cf. \ref{appendix_greedy}) for OP.

For asymmetric TSP with triangle inequality (TMAT), GREAT NB achieves 1.03\% optimality gap ($\times 33$ augmentation), substantially outperforming MatNet (4.4\%) and only being outperformed by BQ-NCO \citep{drakulic2023bq}.
For extremely asymmetric TSP (XASY), GREAT NB achieves 10.72\%, far outperforming MatNet (54.55\%).
For Euclidean TSP, GREAT (0.94\% gap) is competitive though Pointerformer (0.41\%) and BQ-NCO (0.01\%) perform slightly better, underlining the strong optimization of node-based architectures for Euclidean settings.

For CVRP, GREAT NB is the best-performing neural model across all three distributions.
On XASY CVRP, GREAT NB achieves a remarkable 0.76\% gap with $\times 33$ augmentation.
We note that these results are consistent with our findings for EVRP: while HGS achieves better performance (as seen in EVRP settings with few charging stops), it is considerably slower than GREAT.

For OP, GREAT achieves the best results among all neural models across all distributions, with GREAT NB achieving gaps of $-22.53\%$, $-28.87\%$, and $-24.46\%$ relative to the greedy baseline for EUC, TMAT, and XASY.
To the best of our knowledge, GREAT is the first neural approach to be applied to CVRP and OP on asymmetric distributions.

We note that GREAT NB generally outperforms GREAT NF in this experiment setting. We further point out that GREAT MIX (one model jointly trained for all distributions) also achieves convincing performance, often being close to the models trained for a dedicated, single distribution.

\subsection{Generalization to Large Instances and Real-World Libraries}

Table~\ref{table2} shows that GREAT NB exhibits strong zero-shot generalization (ZS in the table) to larger TSP instances.
GREAT NB ZS ($\times 9$) achieves 1.77\% gap on TMAT TSP200 and 37.47\% on XASY for TSP200 without fine-tuning.
After CL fine-tuning, GREAT NB FT200 achieves 2.86\% on EUC TSP200. 
For TSP with 500 nodes, the GREAT NF FT500 achieves convincing performance together with GREAT NB FT200. We note that our models often generalize well to bigger instances but not to smaller instances. We point out that BQ-NCO (which is designed to be trained on small instances and generalize well to large instances) generally performs better than GREAT in these generalization tasks.

Table~\ref{table:greatglop} shows that GREAT outperforms GLOP \citep{ye2024glop} which is a divide-and-conquer method designed for large instances on large TMAT benchmarks (150, 250, 1000 nodes) in both zero-shot and fine-tuned settings.
\realrev{We point out that the results for BQ-NCO and GLOP are taken from the original papers as reported and were not reproduced by us. This is because, unlike MatNet and Pointerformer, those approaches differ not only in the encoder architecture used but also represent completely different training pipelines. GLOP is a divide-and-conquer approach and BQ-NCO is trained using supervised learning. Therefore, it is not possible to retrain them in our exact setting. We include those results to acknowledge the existence of other, alternative strong neural approaches and the comparison to them should be interpreted as indicative rather than exact ablations.
We reiterate that GREAT can be seen as an orthogonal contribution to both GLOP and BQ-NCO and it is possible to use a GREAT-based encoder in the pipelines of those studies. However, this is outside of the scope of this work.}

We further provide Tables~\ref{TSPLIBtablesmall} and~\ref{TSPLIBtablebig} that show real-world generalization results on TSPLIB \citep{reinhelt2014tsplib}, CVRPLIB \citep{uchoa2017new} and OPLIB\footnote{https://github.com/bcamath-ds/OPLib/}.
GREAT NB trained on TMAT achieves 3.81\% mean gap on small, asymmetric TSPLIB instances, substantially outperforming MatNet (16.93\%). GREAT also achieves convincing performance on small Euclidean CVRP and OP instances. However, we note that Pointerformer achieves better performance on Euclidean TSP.
A similar result holds for larger instances with 101-200 nodes.

\begin{table*}[!t]
\centering
\caption{Benchmarking GREAT against neural methods and traditional baselines on TSP and CVRP with 100 nodes. Optimality gaps relative to Gurobi (TSP) and HGS (CVRP). \textbf{Bold} = best, \underline{underline} = second-best.}
\label{table:GREAT}
\begin{tabular}{c l | r r | r r | r r}
& & \multicolumn{2}{c|}{EUC100} & \multicolumn{2}{c|}{TMAT100} & \multicolumn{2}{c}{XASY100} \\ 
&  & opt. gap & time & opt. gap & time & opt. gap & time  \\ \hline\hline
\parbox[t]{2mm}{\multirow{22}{*}{\rotatebox[origin=c]{90}{TSP}}}
& Baseline - Gurobi & - & - & - & - & - & -   \\  \cline{2-8}
& LKH & 0.0\% & 442s & 0.0\% & 936s & 0.01\% & 1006s  \\
& Nearest Neighbor & 24.86\% & 1.3s & 36.04\% & 1.2s & 185.26\% & 1.2s \\ 
& Nearest Insertion & 21.8\% & 27s & 31.8\% & 28s & 301.65\% &  28s \\
& Farthest Insertion & 7.66\% & 32s  & 23.92\% & 33s & 310.98\% & 33s\\
\cline{2-8}
& BQ-NCO & $\textbf{0.01\%}^*$ & $\text{32m}^*$ & $\textbf{0.96\%}^*$ & $\text{19m}^*$ & UNK & UNK \\ 
& Pointerformer x1  &  0.96\% & 18s  & NA &  NA & NA &  NA \\
& Pointerformer x8  & \underline{0.41\%} & 131s  & NA &  NA & NA &  NA \\
& MatNet x1 & 3.28\% & 29s & 7.54\% & 27s & 82.26\% & 27s \\
& MatNet x9 & 2.09\% & 233s & 5.28\% & 229s & 62.57\% & 231s \\
& MatNet x33 & 1.69\% & 814s & 4.4\% & 812s& 54.55\% & 808s \\ 
\cline{2-8}
& GREAT NF x1 & 1.32\% & 61s & 3.73\% & 65s & 28.54\% & 64s \\
& GREAT NF x9 & 0.9\% & 525s & 2.75\% & 579s & 21.15\% & 568s\\
& GREAT NF x33 &  0.85\% & 1952s & 2.59\% & 2071s & 20.06\% & 2075s \\
& GREAT NB x1 & 1.43\% & 47s  & 1.53\% & 69s  &  21.93\% & 70s  \\
& GREAT NB x9 & 1.0\% & 404s & 1.07\% & 614s &  \underline{12.6\%} & 610s \\
& GREAT NB x33 & 0.94\% & 1446s & \underline{1.03\%} & 2235s & \textbf{10.72\%} & 2206s  \\
\cline{2-8}
& GREAT MIX NF x1  &  4.45\% & 64s   &  5.77\% & 64s  & 45.78\% & 64s   \\
& GREAT MIX NF x9  &  3.5\% &  563s & 4.63\% &  554s &  33.32\% &  552s  \\
& GREAT MIX NF x33  &  3.2\% & 2017s &  4.39\% &  2014s &  29.8\% & 2015s  \\
& GREAT MIX NB x1  & 2.53\% & 70s   &  4.99\% & 70s   &  45.97\% & 70s  \\
& GREAT MIX NB x9  &  1.9\% &  612s &  3.85\% & 608s &    33.19\% & 609s \\
& GREAT MIX NB x33  &   1.84\% &  2212s &   3.7\% & 2211s &    30.74\% &  2207s  \\
\hline\hline
& Baseline - HGS & - & 544.2m& - & 1316.6m & - & 2163.4m   \\ \cline{2-8}
\parbox[t]{2mm}{\multirow{16}{*}{\rotatebox[origin=c]{90}{CVRP}}}
& Pointerformer x1  &  4.6\% & 17s  & NA &  NA & NA &  NA \\
& Pointerformer x8  & 3.29\% & 115s  & NA &  NA & NA &  NA \\
& MatNet x1 & 4.79\% & 27s & 9.41\% & 27s & 32.54\% & 27s \\
& MatNet x9 & 3.6\% & 205s & 7.55\% & 208s & 24.23\% & 209s \\
& MatNet x33 & 3.21\% & 725s & 6.72\% & 727s & 20.37\% & 738s  \\
\cline{2-8}
& GREAT NF x1 & 3.88\% & 64s & 5.49\% & 65s & 12.1\% & 64s\\
& GREAT NF x9 & 3.13\% & 540s & 4.4\% & 551s & 6.29\% & 541s \\
& GREAT NF x33 & 3.01\% & 1958s & 4.2\% & 1983s & 5.31\% & 1963s \\
& GREAT NB x1 & 3.06\% & 71s  & 4.66\% & 70s & 8.19\% & 70s \\
& GREAT NB x9 & \underline{2.47\%}  & 601s   &  \underline{3.72\%} & 594s & \underline{2.23\%} & 595s \\
& GREAT NB x33 & \textbf{2.39\%} & 2184s & \textbf{3.6\%} & 2172s & \textbf{0.76\% } & 2171s \\
\cline{2-8}
& GREAT MIX NF x1  & 4.12\% & 64s  & 6.37\% & 64s  & 14.83\% & 64s   \\
& GREAT MIX NF x9  & 3.38\% & 543s  & 5.11\% & 536s & 8.04\% &  542s \\
& GREAT MIX NF x33  & 3.15\% & 1961s & 4.8\% & 1956s  & 5.42\% &  1962s \\
& GREAT MIX NB x1  & 3.98\% & 70s  & 5.49\% & 71s &  11.88\% & 71s  \\
& GREAT MIX NB x9  & 3.18\% & 594s  & 4.35\% & 598s &  4.91\% & 604s  \\
& GREAT MIX NB x33  & 3.03\% & 2148s  & 4.15\% & 2183s &  3.18\% & 2199s   \\
\multicolumn{8}{c}{* values from original paper, model was not reevaluated by us}
\end{tabular}
\end{table*}

\begin{table*}[!t]
\centering
\caption{Benchmarking GREAT against neural methods and traditional baselines on OP with 100 nodes. Optimality gaps relative to greedy heuristic. \textbf{Bold} = best, \underline{underline} = second-best.}
\label{table:GREAT2}
\begin{tabular}{c l | r r | r r | r r}
& & \multicolumn{2}{c|}{EUC100} & \multicolumn{2}{c|}{TMAT100} & \multicolumn{2}{c}{XASY100} \\ 
&  & opt. gap & time & opt. gap & time & opt. gap & time  \\ \hline\hline
& Baseline - Greedy & - & 7.9s  & - & 6.9s  & - & 340s \\ \cline{2-8}
\parbox[t]{2mm}{\multirow{16}{*}{\rotatebox[origin=c]{90}{OP}}}
& EA4OP & -23.64\% & 5609s & NA  & NA & NA  & NA \\
\cline{2-8}
& Pointerformer x1  &  -19.61\% & 12s  & NA &  NA  & NA &  NA \\
& Pointerformer x8  & -21.31\% & 74s  & NA &  NA   & NA &  NA \\
& MatNet x1 & -15.54\% & 21s & -24.13\% & 21s & 54.98\% & 17s \\
& MatNet x9 & -18.08\% & 161s & -25.51\% & 166s &  28.76\% & 137s \\
& MatNet x33 & -18.99\% & 551s &  -26.01\% & 583s & 20.15\% & 471s  \\
\cline{2-8}
& GREAT NF x1 & -20.92\% & 59s & -27.45\% & 60s &  -18.1\% & 56s \\
& GREAT NF x9 & -21.82\% & 498s & -28.24\% & 510s &  -21.34\% & 479s  \\
& GREAT NF x33 & -22.03\% & 1806s & -28.39\% & 1840s &  -21.5\%  & 1738s\\
& GREAT NB x1 &  -21.65\% & 65s   & -28.08\% & 66s  &  -20.69\% & 62s \\
& GREAT NB x9 &   \underline{-22.38}\% & 552s    &  \underline{-28.71\%} & 559s &  \underline{-24.18\%} &  533s \\
& GREAT NB x33 & \textbf{ -22.53\%} & 1999s    &  \textbf{-28.87\%} & 2011s &  \textbf{-24.46\%} & 1948s \\
\cline{2-8}
& GREAT MIX NF x1  & -16.97\% & 58s   &  -24.43\% & 60s  & -15.84\% & 55s   \\
& GREAT MIX NF x9  & -18.59\% &  493s & -26.07\% &  498s & -21.06\% &   478s \\
& GREAT MIX NF x33  &  -19.09\% & 1774s &  -26.38\% & 1806s  &  -21.78\% &  1742s \\
& GREAT MIX NB x1  & -18.22\% & 65s  &  -25.5\% & 66s  &  -19.34\% &  62s \\
& GREAT MIX NB x9  &  -19.56\% &  548s & -27.02\% & 556s &   -23.7\% & 537s \\
& GREAT MIX NB x33  &  -19.96\% &  1995s &  -27.35\% & 2032s &  -24.08\% & 1928s   \\
\end{tabular}
\end{table*}

\begin{table*}[t]
\centering
\caption{Generalization on 128 synthetic test instances of larger TSP sizes}
\label{table2}
\begin{tabular}{c l | r r | r r | r r}
& & \multicolumn{2}{c|}{EUC} & \multicolumn{2}{c|}{TMAT} & \multicolumn{2}{c}{XASY} \\ 
&  & opt. gap & time & opt. gap & time & opt. gap & time  \\ \hline\hline
\parbox[t]{2mm}{\multirow{16}{*}{\rotatebox[origin=c]{90}{TSP200}}}
& Baseline - LKH & - & 23s & - & 40s & - & 36s   \\  \cline{2-8}
& Nearest Neighbor & 25.5\% & 0.1s & 36.71 \% & 0.1s & 224.93 \% & 0.1s \\ 
& Nearest Insertion & 23.39\% & 3.8s & 37.95 \% & 3.9s &  465.76\% &  4.1s \\
& Farthest Insertion & 9.04\% &  2.9s & 29.41 \% & 2.9s &  481.4\% & 3.0s \\
 \cline{2-8}
& BQ-NCO bs16*  & \textbf{0.09\%*} & 3min* & \textbf{1.41\%*} & 1min* & UNK & UNK \\ 
\cline{2-8}
& GREAT NF ZS x1 & 15.78\% & 12s & 6.97\% & 12s & 209.3 \% & 12s  \\
& GREAT NF ZS x9 & 13.69\% & 41s & 5.42\% & 44s &  148.18\% & 43s \\
& GREAT NB ZS x1 &  5.27\% &  11s &  2.27\% &  13s &  53.28 \% & 13s  \\
& GREAT NB ZS x9 &  4.54\% & 37s &  \underline{1.77\%} & 55s &  \textbf{37.47 \%} &  53s\\
 \cline{2-8}
& GREAT NF FT200 x1 & 3.29\% & 13s  & 5.08\% & 12s & 56.0 \% & 12s \\
& GREAT NF FT200 x9 & 2.66\% & 37s & 4.18\% & 38s & 45.22 \% & 38s\\
& GREAT NB FT200 x1 &  3.47\% & 11s  & 2.48\% &  13s &  51.43 \% & 13s  \\
& GREAT NB FT200 x9 &  \underline{2.86\%} & 29s &  1.92\% & 42s &   \underline{38.38\%} &  42s\\
 \cline{2-8}
& GREAT NF FT500 x1 & 105.86\% & 13s & 7.28\% & 12s &  142.91\% & 12s \\
& GREAT NF FT500 x9 & 92.58\% & 37s & 6.46\% & 38s &  103.47\% & 38s\\
& GREAT NB FT500 x1 & 5.94 \% &  10s &  29.08\% & 13s  &  134.77 \% & 13s \\
& GREAT NB FT500 x9 &  4.67 \% & 29s &  28.54\% & 42s &  127.68 \% & 42s \\
\hline\hline
\parbox[t]{2mm}{\multirow{16}{*}{\rotatebox[origin=c]{90}{TSP500}}}
& Baseline - LKH & - & 272s & - & 373s & - &  367s  \\  \cline{2-8}
& Nearest Neighbor & 25.78\% & 0.4s & 37.27 \% & 0.4s & 280.26 \% & 0.5s \\ 
& Nearest Insertion & 25.12\% & 41s & 46.77\% & 51s &  801.73\% &  49s \\
& Farthest Insertion & 10.66\% &  50s &  36.61\% & 65s & 815.74 \% & 77s \\
 \cline{2-8}
 & BQ-NCO bs16* & \textbf{0.55\%*} & 15min* & \textbf{2.43\%*} & 3min* & UNK & UNK \\ 
\cline{2-8}
& GREAT NF ZS x1 & 49.81\% & 49s & 56.46\% & 46s & 369.13 \% & 46s \\
& GREAT NF ZS x9 & 39.51\% & 288s & 31.42\% & 301s & 309.12 \% & 297s \\
& GREAT NB ZS x1 &  16.75 \% &  40s &  3.66\% & 49s  &  307.52 \% &  49s \\
& GREAT NB ZS x9 &  15.89\% &  247s&  3.14\% & 315s &  246.44 \% & 318s \\
 \cline{2-8}
& GREAT NF FT200 x1 & 36.13\% & 46s  & 7.72 \% & 45s & 470.31 \% & 45s \\
& GREAT NF FT200 x9 & 33.13\% & 286s & 6.81\% & 291s &  367.82 \% & 293s\\
& GREAT NB FT200 x1 &  10.96 \% & 39s  & 3.42\% & 49s  &  136.48 \% & 46s  \\
& GREAT NB FT200 x9 & 10.16 \% & 245s &  \underline{3.01\%} & 317s &   \underline{118.01\%} & 311s \\
 \cline{2-8}
& GREAT NF FT500 x1 & 7.52\% & 44s & 7.06\% & 45s &  128.75\% & 46s \\
& GREAT NF FT500 x9 & \underline{6.96\%} & 283s & 6.5\% & 291s &  \textbf{105.23\%} & 314s\\
& GREAT NB FT500 x1 & 8.04 \% & 42s  &  32.99\% & 46s  &  179.01 \% & 47s  \\
& GREAT NB FT500 x9 & 7.11 \% & 259s &  31.98\% & 310s &   167.71 \% & 310s\\
\multicolumn{8}{c}{$^*$ values from original paper tested on a different dataset of 128 instances} \\

\end{tabular} \vspace{0.1cm}
\end{table*}

\begin{table*}[t]
\centering
\caption{Generalization results on large TMAT instances (30 instances, from \cite{ye2024glop})}
\label{table:greatglop}
\begin{tabular}{c|cc|cc|cc}
  & \multicolumn{2}{c|}{TMAT150} &  \multicolumn{2}{c|}{TMAT250} & \multicolumn{2}{c}{TMAT1000} \\
  &  opt. gap & time & opt. gap & time & opt. gap & time  \\ \hline\hline
LKH & - & 6s & - &  15s & - & 231s \\ \hline
BQ-NCO bs16 & UNK & UNK & UNK & UNK & $8.26 \%^*$ & $7\text{m}^*$\\
GLOP & $19.28\%^{\text{\textdagger}}_{\diamond}$ & $8.2s^{\text{\textdagger}}$ & $29.67\%^{\text{\textdagger}}_{\diamond}$ & $9.3s^{\text{\textdagger}}$  &  $44.27 \%^{\text{\textdagger}}_{\diamond}$ & $15s^{\text{\textdagger}}$ \\ \hline
GREAT NF ZS x1& 4.47\% & 3s & 11.98\%  & 6s & 85.54\% & 54s \\
GREAT NF ZS x9& 3.42\% & 19s &  7.81\% & 17s & 52.07\% & 455s \\
GREAT NB ZS x1& 1.64\% & 3s & 2.23\%  & 6s & 4.17\% & 56s \\
GREAT NB ZS x9& \textbf{1.28\%} & 11s &  \textbf{1.8\%} & 17s & 2.76\% & 470s \\
\hline
GREAT NF FT200 x1& 4.3\% & 3s & 4.91 \% & 6s & 19.05\% & 54s\\
GREAT NF FT200 x9& 3.28\% & 8s & 4.18\% & 17s & 15.42\% & 457s\\
GREAT NB FT200 x1& 2.12\% & 3s &  2.6\% & 6s & \underline{1.8\%} & 56s\\
GREAT NB FT200 x9& \underline{1.54\%} & 8s & \underline{1.9\%} & 18s & \textbf{1.45\%} & 472s \\
\multicolumn{7}{c}{$^*$ values from the original paper (different dataset of 128 instances)} \\
\multicolumn{7}{c}{$^\text{\textdagger}$ values from the original paper (same dataset as GREAT)} \\
\multicolumn{7}{c}{$\diamond$ original paper reports absolute values which we transformed into opt. gaps}
\end{tabular}
\end{table*}

\begin{table*}[t]
\centering
\caption{Generalization to real-world routing libraries ($\leq 100$ nodes)}
\label{TSPLIBtablesmall}
\begin{tabular}{c@{\hspace{6mm}}l c | c  c c c c c}
 &  & Dist. &  Mean & Q10 & Q25 & Median & Q75 & Q90 \\ \hline\hline
\parbox[t]{2mm}{\multirow{7}{*}{\rotatebox[origin=c]{90}{\shortstack{TSPLIB\\$n = 12$}}}}
& Baseline - LKH & - & - & - & - & - & - & -  \\
 \cline{2-9}
& GREAT NB x9 & EUC & 2.38\% & \underline{0.66\%} & \underline{1.32\%} & \underline{2.16\%} & 3.44\% & 3.65\% \\
& GREAT NF x9 & EUC & 20.45\% & 0.72\% & 1.9\% & 3.56\% & 7.17\% & 57.38\% \\
& GREAT NB x9 & MIX & \underline{2.25\%} & 1.45\% & 1.85\% & 2.28\% & \underline{2.59\%} & \textbf{2.82\%} \\
& GREAT NF x9 & MIX & 5.6\% & 1.61\% & 2.69\% & 4.64\% & 7.76\% & 10.68\% \\
& MatNet x9 & EUC & 2.82\% & 1.08\% & 1.43\% & 3.08\% & 3.55\% & 3.97\% \\
& Pointerformer x8 & EUC & \textbf{1.36\%} & \textbf{0.04\%} & \textbf{0.18\%} & \textbf{1.19\%} & \textbf{2.33\%} & \underline{2.87\%} \\
\hline\hline
\parbox[t]{2mm}{\multirow{9}{*}{\rotatebox[origin=c]{90}{\shortstack{ATSPLIB\\$n = 14$}}}}
& Baseline - LKH & - &  - & - & - & - & - & -  \\
 \cline{2-9}
& GREAT NB x9 & TMAT & \textbf{3.81\%} & \textbf{0.46\%} & \textbf{0.97\%} & \textbf{2.65\%} & \textbf{6.0\%} & \textbf{8.85\%} \\
& GREAT NF x9 & TMAT & \underline{6.03\%} & \underline{2.1\%} & \underline{3.47\%} & \underline{6.06\%} & \underline{8.56\%} & \underline{10.4\%} \\
& GREAT NB x9 & XASY & 45.88\% & 9.33\% & 14.74\% & 18.27\% & 41.47\% & 80.41\% \\
& GREAT NF x9 & XASY & 27.12\% & 6.26\% & 18.48\% & 27.32\% & 37.51\% & 39.37\% \\
& GREAT NB x9 & MIX & 17.65\% & 5.22\% & 8.46\% & 15.79\% & 24.29\% & 29.97\% \\
& GREAT NF x9 & MIX & 19.58\% & 4.65\% & 5.53\% & 7.44\% & 9.06\% & 16.09\% \\
& MatNet x9 & TMAT & 16.93\% & 8.89\% & 12.29\% & 18.94\% & 20.1\% & 25.32\% \\
& MatNet x9 & XASY & 14.46\% & 5.09\% & 9.62\% & 15.04\% & 20.43\% & 22.06\% \\
\hline\hline
\parbox[t]{2mm}{\multirow{7}{*}{\rotatebox[origin=c]{90}{\shortstack{CVRPLIB\\$n = 92$}}}}
& Baseline - HGS & - & - & - & - & - & - & -  \\
 \cline{2-9}
& GREAT NB x9 & EUC & \textbf{3.82\%} & \underline{1.44\%} & \textbf{2.05\%} & \textbf{2.93\%} & 4.63\% & 7.52\% \\
& GREAT NF x9 & EUC & 50.49\% & 4.62\% & 7.37\% & 30.24\% & 92.81\% & 120.92\% \\
& GREAT NB x9 & MIX & \underline{3.92\%} & 1.46\% & 2.23\% & 3.24\% & \textbf{4.46\%} & \underline{7.49\%} \\
& GREAT NF x9 & MIX & 7.93\% & 2.55\% & 3.32\% & 4.83\% & 8.02\% & 15.21\% \\
& MatNet x9 & EUC & 7.62\% & 2.04\% & 3.03\% & 4.01\% & 8.09\% & 16.57\% \\
& Pointerformer x8 & EUC & 4.09\% & \textbf{1.33\%} & \underline{2.2\%} & \underline{3.21\%} & \underline{4.47\%} & \textbf{7.01\%} \\
\hline\hline
\parbox[t]{2mm}{\multirow{7}{*}{\rotatebox[origin=c]{90}{\shortstack{OPLIB\\$n = 48$}}}}
& Baseline - EA4OP & - &  - & - & - & - & - & -  \\
 \cline{2-9}
& GREAT NB x9 & EUC & \underline{8.68\%} & \textbf{1.83\%} & \textbf{2.84\%} & \textbf{5.18\%} & \textbf{6.46\%} & \textbf{8.29\%} \\
& GREAT NF x9 & EUC & 11.66\% & 3.59\% & \underline{4.75\%} & 8.37\% & 10.58\% & 28.11\% \\
& GREAT NB x9 & MIX & 9.12\% & 4.19\% & 5.64\% & 8.92\% & 11.14\% & \underline{13.0\%} \\
& GREAT NF x9 & MIX & 12.1\% & 6.31\% & 8.91\% & 11.04\% & 13.42\% & 18.64\% \\
& MatNet x9 & EUC & 14.9\% & 7.21\% & 10.76\% & 13.26\% & 17.37\% & 24.36\% \\
& Pointerformer x8 & EUC & \textbf{8.19\%} & \underline{2.0\%} & 4.81\% & \underline{7.5\%} & \underline{9.87\%} & 13.49\% \\
\end{tabular} \vspace{0.1cm}
\end{table*}

\begin{table*}[t]
\centering
\caption{Generalization to real-world routing libraries (101--200 nodes)}
\label{TSPLIBtablebig}
\begin{tabular}{c@{\hspace{6mm}}l c | c  c c c c c}
 &  & Dist. &  Mean & Q10 & Q25 & Median & Q75 & Q90 \\ \hline\hline
\parbox[t]{2mm}{\multirow{6}{*}{\rotatebox[origin=c]{90}{\shortstack{TSPLIB\\$n = 17$}}}}
& Baseline - LKH & - & - & - & - & - & - & -  \\
 \cline{2-9}
& GREAT NB x9 & EUC & \underline{5.97\%} & 1.68\% & \underline{2.2\%} & \underline{3.94\%} & \underline{6.89\%} & 10.76\% \\
& GREAT NF x9 & EUC & 9.27\% & \underline{1.55\%} & 3.44\% & 7.07\% & 14.26\% & 22.3\% \\
& GREAT NB x9 & MIX & 6.28\% & 2.42\% & 3.11\% & 4.35\% & 7.03\% & \underline{8.56\%} \\
& GREAT NF x9 & MIX & 14.26\% & 3.81\% & 5.81\% & 13.87\% & 20.91\% & 23.64\% \\
& Pointerformer x8 & EUC & \textbf{3.9\%} & \textbf{0.93\%} & \textbf{1.63\%} & \textbf{2.18\%} & \textbf{5.3\%} & \textbf{6.88\%} \\
\hline\hline
\parbox[t]{2mm}{\multirow{6}{*}{\rotatebox[origin=c]{90}{\shortstack{CVRPLIB\\$n = 49$}}}}
& Baseline - HGS & - & - & - & - & - & - & -  \\
 \cline{2-9}
& GREAT NB x9 & EUC & \textbf{7.97\%} & \textbf{2.38\%} & \textbf{2.81\%} & \textbf{5.23\%} & \textbf{6.72\%} & \textbf{10.49\%} \\
& GREAT NF x9 & EUC & 9.38\% & 3.64\% & 5.87\% & 8.43\% & 11.9\% & 15.87\% \\
& GREAT NB x9 & MIX & 14.94\% & 3.57\% & \underline{4.68\%} & \underline{6.23\%} & \underline{9.8\%} & \underline{12.01\%} \\
& GREAT NF x9 & MIX & 9.36\% & \underline{3.55\%} & 4.9\% & 7.41\% & 12.92\% & 17.5\% \\
& Pointerformer x8 & EUC & \underline{8.61\%} & 3.77\% & 5.14\% & 6.25\% & 12.4\% & 15.46\% \\
\hline\hline
\parbox[t]{2mm}{\multirow{6}{*}{\rotatebox[origin=c]{90}{\shortstack{OPLIB\\$n = 60$}}}}
& Baseline - EA4OP & - &  - & - & - & - & - & -  \\
 \cline{2-9}
& GREAT NB x9 & EUC & 18.54\% & \textbf{2.64\%} & \textbf{4.81\%} & \textbf{6.11\%} & \textbf{9.91\%} & \textbf{17.94\%} \\
& GREAT NF x9 & EUC & 16.16\% & \underline{5.61\%} & 7.57\% & 11.81\% & 17.94\% & 24.86\% \\
& GREAT NB x9 & MIX & \underline{14.39\%} & 7.17\% & 8.44\% & 11.16\% & 15.34\% & 30.06\% \\
& GREAT NF x9 & MIX & 19.36\% & 7.46\% & 11.59\% & 16.84\% & 24.07\% & 30.99\% \\
& Pointerformer x8 & EUC & \textbf{12.93\%} & 5.95\% & \underline{7.49\%} & \underline{10.52\%} & \underline{13.02\%} & \underline{19.1\%} \\
\end{tabular} \vspace{0.1cm}
\end{table*}

\section{Greedy Heuristic for the Orienteering Problem}
\label{appendix_greedy}
In our experiments, we use a greedy heuristic as a baseline for solving the OP. 
This method prioritizes visiting customers that offer a favorable trade-off between prize and distance, while respecting the specified maximum travel distance.
At each step, starting from the current node $i$, the algorithm selects the next unvisited node $j$ that maximizes the ratio
\[
\frac{\text{prize}(j)}{\text{distance}(i, j)},
\]
subject to the constraint that traveling to $j$ and then returning to the depot does not exceed the remaining travel budget. This means that we mask all moves $j$ where
\[
\text{distance}(i, j) + \text{distance}(j, \text{depot})
\]
exceeds the remaining travel distance.

While this approach is straightforward on Euclidean instances, applying it to instances of the asymmetric XASY distribution is significantly more challenging.
Since the triangle inequality does not hold for XASY, we cannot rely on simple additive checks to ensure feasibility. 
That is, even if the direct distances from $i$ to $j$ and from $j$ to the depot appear infeasible, the actual shortest return path might respect the travel budget. Consequently, we must compute the true shortest return path to the depot from each candidate node $j$ in order to check feasibility. 
This requires solving a series of shortest path problems during the greedy selection process (since after each selection step the shortest path might change as we cannot visit nodes multiple times), which results in an increased runtime on XASY instances.

\end{document}